\documentclass[10pt,twocolumn,letterpaper]{article}

\usepackage[pagenumbers]{cvpr}

\usepackage{xcolor}

\newcount\paraboxline
\newcommand{\parabox}[3]{%
  \par
  \begingroup
  \setlength{\parskip}{0pt}%
  \noindent{\footnotesize\textbf{#2}
    \textcolor{gray}{(#1 lines $\approx$ \the\numexpr#1*10\relax{} words)}:
    #3}\par
  \paraboxline=1
  \loop\ifnum\paraboxline<#1
    \noindent\textcolor{gray!22}{\rule[-0.25ex]{\linewidth}{1.7ex}}\par
    \advance\paraboxline by 1
  \repeat
  \endgroup
  \par}

\newcommand{\method}{RoboProbe}

\newcommand{\mSolid}{Solidity}
\newcommand{\mContact}{Contact fidelity}
\newcommand{\mPersist}{Object persistence}
\newcommand{\mRobotDef}{Robot deformation}
\newcommand{\mObjDef}{Object deformation}
\newcommand{\sSolid}{Solidity}
\newcommand{\sContact}{Contact}
\newcommand{\sPersist}{Persistence}
\newcommand{\sRobotDef}{Robot def.}
\newcommand{\sObjDef}{Object def.}

\usepackage{array}
\usepackage{multirow}
\usepackage{tabularx}
\usepackage{colortbl}
\usepackage{pifont}
\usepackage{stfloats} %
\newcommand{\dashsep}[1]{\arrayrulecolor{gray!55}\cmidrule[0.3pt](lr){#1}\arrayrulecolor{black}}

\newcommand{\noagent}[1]{\,{\color{gray}\scriptsize(#1)}}
\newcolumntype{Y}{>{\centering\arraybackslash}X}
\newcolumntype{L}{>{\raggedright\arraybackslash}X}
\definecolor{tabgray}{HTML}{F4F6F8}
\definecolor{tabblue}{HTML}{EBF3FA}
\definecolor{tabgreen}{HTML}{EAF6EE}
\newcommand{\cmark}{\textcolor{ForestGreen}{\ding{51}}}
\newcommand{\xmark}{\textcolor{BrickRed}{\ding{55}}}
\newcommand{\best}[1]{\textbf{#1}}
\newcommand{\second}[1]{\underline{#1}}
\newcommand{\na}{\textcolor{gray}{--}}

\makeatletter
\renewcommand\paragraph{\@startsection{paragraph}{4}{\z@}{0.6ex \@plus .2ex \@minus .1ex}{-0.6em}{\normalfont\normalsize\bfseries}}
\makeatother

\usepackage{listings}
\lstnewenvironment{promptbox}{\lstset{basicstyle=\ttfamily\scriptsize,breaklines=true,breakatwhitespace=false,columns=fullflexible,keepspaces=true,frame=single,framerule=0.3pt,rulecolor=\color{black!30},backgroundcolor=\color{black!3},framesep=3pt,xleftmargin=3pt,xrightmargin=3pt,aboveskip=4pt,belowskip=6pt,showstringspaces=false,escapeinside={(*@}{@*)}}}{}

\definecolor{cvprblue}{rgb}{0.21,0.49,0.74}
\usepackage[pagebackref,breaklinks,colorlinks,allcolors=cvprblue]{hyperref}

\def\confName{CVPR}
\def\confYear{2027}

\title{RoboProbe: Benchmarking Physical Plausibility of Robot World Models}

\author{Zhuoying Yang\textsuperscript{*} \qquad Jiachen Ni\textsuperscript{*} \qquad Yichen Guo \qquad
Xiaoshan Wu\textsuperscript{\dag} \qquad Mengyu Wang\textsuperscript{\dag}\\
{\small \textsuperscript{*}Equal contribution \qquad \textsuperscript{\dag}Co-senior authors}
}

\begin{document}

\twocolumn[{%
\renewcommand\twocolumn[1][]{#1}%
\maketitle
\begin{center}
  \centering
  \includegraphics[width=\textwidth]{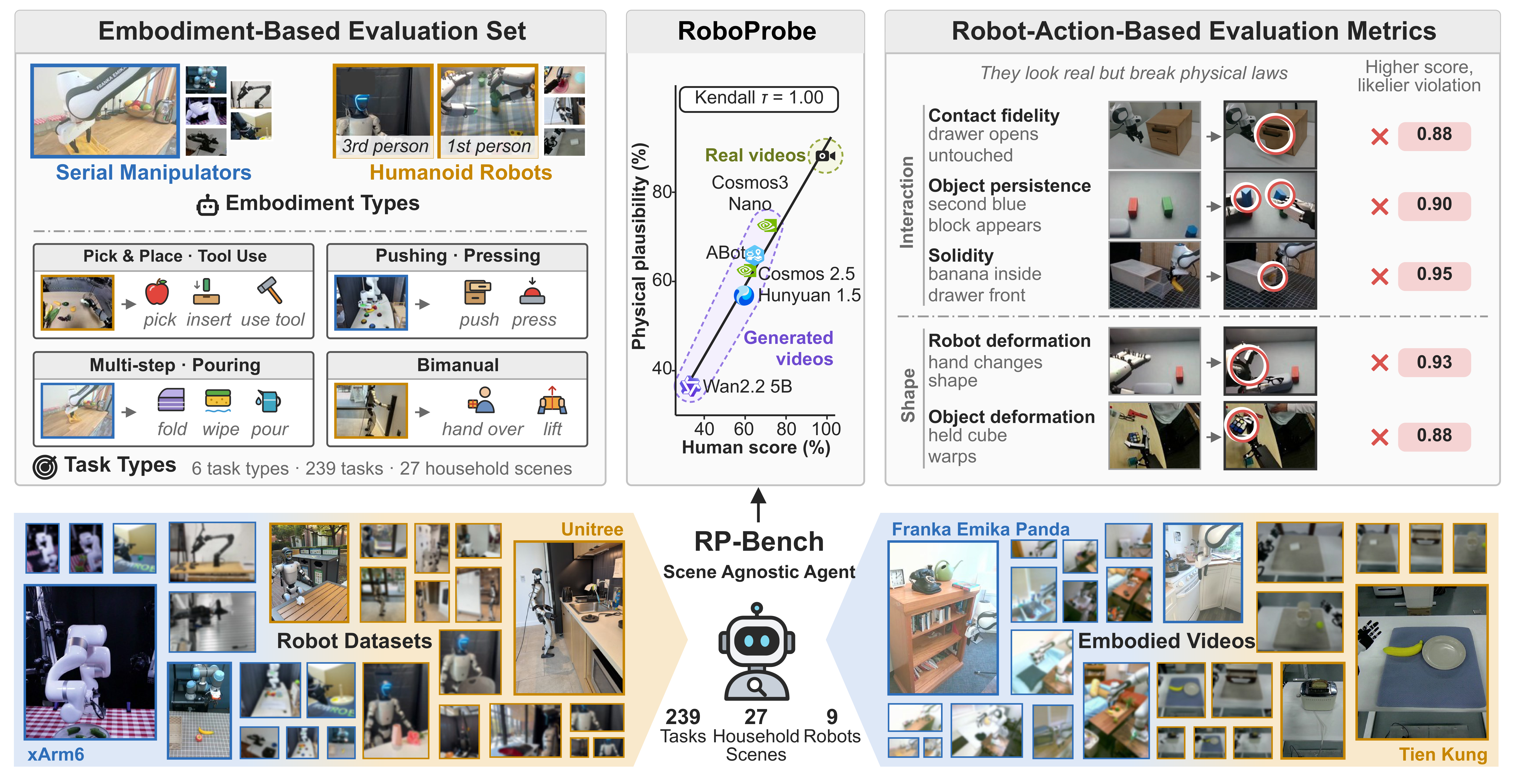}
  \captionof{figure}{\textbf{Generated robot videos look real but break physical laws.} Left, the benchmark covers serial manipulators and humanoids in first- and third-person views. Middle, for each generator, the share of videos that \method{} finds plausible against the share the annotators do on the 130 shared tasks; the rankings agree (Kendall $\tau{=}1.00$). Right, one generated video per violation type, with the violation circled in red and the metric's score, a higher score meaning a likelier violation.%
  }
  \label{fig:teaser}
\end{center}
\vspace{0.5em}
}]

\begin{abstract}
Evaluating the physical plausibility of generated videos is critical for reliable robotic planning with world models. These videos inform action proposal and assessment before real-world execution, so undetected physical violations can lead to erroneous assessments of action feasibility and safety.
Yet current video generators often produce physically implausible interactions. Detecting such violations directly with vision-language models (VLMs) remains challenging as physical verification requires evidence-grounded spatiotemporal reasoning beyond high-level semantic understanding.
We develop five law-specific metrics for evaluating physical plausibility by combining spatiotemporal evidence from segmentation, tracking, 3D geometry, or appearance comparisons with targeted VLM reasoning: \mContact{}, \mPersist{}, \mSolid{}, \mRobotDef{}, and \mObjDef{}.
Building on this metric suite, we introduce \textbf{\method{}}, an agentic evaluation framework that coordinates metric execution and selectively refines evaluations through constrained actions.
To evaluate the physical plausibility of robot world models, we further introduce \textbf{RP-Bench}, with 761 human-labeled generated videos and 130 real demonstrations of 239 manipulation tasks from 12 robot datasets.
On RP-Bench, all five metrics achieve AUROCs above 0.85 and outperform the evaluated learned judges, prior evaluators, and a frames-only Gemini baseline.
Applying these metrics to five video generators reveals distinct patterns of physical violations across models, in both violation types and frequencies.
\end{abstract}

\section{Introduction}
\label{sec:intro}

Embodied World Models (EWMs) generate future visual rollouts of robot-environment interactions to support planning, synthetic data generation, and policy evaluation~\cite{cosmos,cosmos25,cosmos3,abotphysworld,lvp,unipi}. Because these rollouts inform physical decisions, mere perceptual realism is insufficient: a visually convincing rollout can still violate physics, undermining the relationship between robot actions and environmental responses and misleading downstream planning and policy evaluation.
In order for the EWMs to be reliable tools for robot agents, evaluating EWMs therefore requires answering a critical question: \emph{are the generated robot actions and environmental responses physically plausible?}  

Early evaluations of generated videos measure pixel-level fidelity and temporal consistency~\cite{vbench,fvd,ewmbench}, which say little about whether an interaction obeys physics.
Most recent physical-plausibility benchmarks instead ask a large vision-language model (VLM) to judge the generated video directly~\cite{videophy2,rbench,pai}.
VLMs are strong at high-level semantic understanding, but detecting these violations requires precise spatiotemporal reasoning about 3D geometry, which remains a weakness of current VLMs.
Simulator-based benchmarks~\cite{robowmbench,kinebench} provide more explicit physical measurements, but they are limited to supported environments and vulnerable to action-extraction noise.
Existing 3D geometric evaluators either score a single object in isolation~\cite{pdi} or need simulator ground truth and leave robot--object contact and interpenetration to a VLM~\cite{robophys3d}.

To close this gap, we introduce \textbf{\method{}}, which grounds physical-plausibility judgments in spatiotemporal evidence measured with segmentation, tracking, and 3D geometry.
We design five law-specific metrics covering \mContact{}, \mPersist{}, \mSolid{}, \mRobotDef{}, and \mObjDef{}.
Beyond the fixed metrics, we adapt our measurement to diverse scenes, tasks, and atypical cases with an agentic framework that reasons over each video's scene, robot type, and entities.
It reviews and selectively refines the evidence through constrained actions.

To systematically probe whether current evaluators can catch these violations, we curate RP-Bench, with 761 labeled generated videos and 130 real ones from 12 robot datasets.
The six task types are pick-and-place, pushing and pressing, multi-step, bimanual, tool-use, and pouring tasks.
On RP-Bench, state-of-the-art proprietary VLMs provide meaningful signals but remain limited in precise detection and spatiotemporal evidence grounding.
Judging from sampled frames, they fall furthest behind on brief, geometric violations such as surface penetration.
By measuring this evidence directly, \method{} outperforms the strongest VLM, Gemini 4 Argon~\cite{gemini}, on all five violation types (average AUROC 0.894 vs.\ 0.828).
Its reported frame lies a median of 0.5 to 7 frames from the human annotation.
Our main contributions are as follows:

\begin{itemize}
\item \textbf{Geometry-Grounded Physical Plausibility Metrics:}
We develop five geometry-grounded metrics that translate physical laws into concrete checks of object states and interactions over time.

\item \textbf{Self-Reviewing Agentic Evaluation:}
We propose \method{}, an agentic framework that coordinates the
proposed metrics to assess physical plausibility in complex and
diverse generated videos. By combining tool-derived evidence with
targeted VLM reasoning and self-review, \method{} grounds its
assessments in specific physical evidence rather than on holistic semantic judgments.

\item \textbf{RP-Bench for Robotic Physical Plausibility:}
We introduce RP-Bench, a human-annotated benchmark comprising
761 generated videos and 130 real demonstrations spanning 239 tasks from 12 robot datasets.
On RP-Bench, \method{} outperforms a frames-only VLM baseline on all five violation types.
\end{itemize}

\section{Related Work}
\label{sec:related}
\begin{figure*}[b]
\centering
  \includegraphics[width=\textwidth]{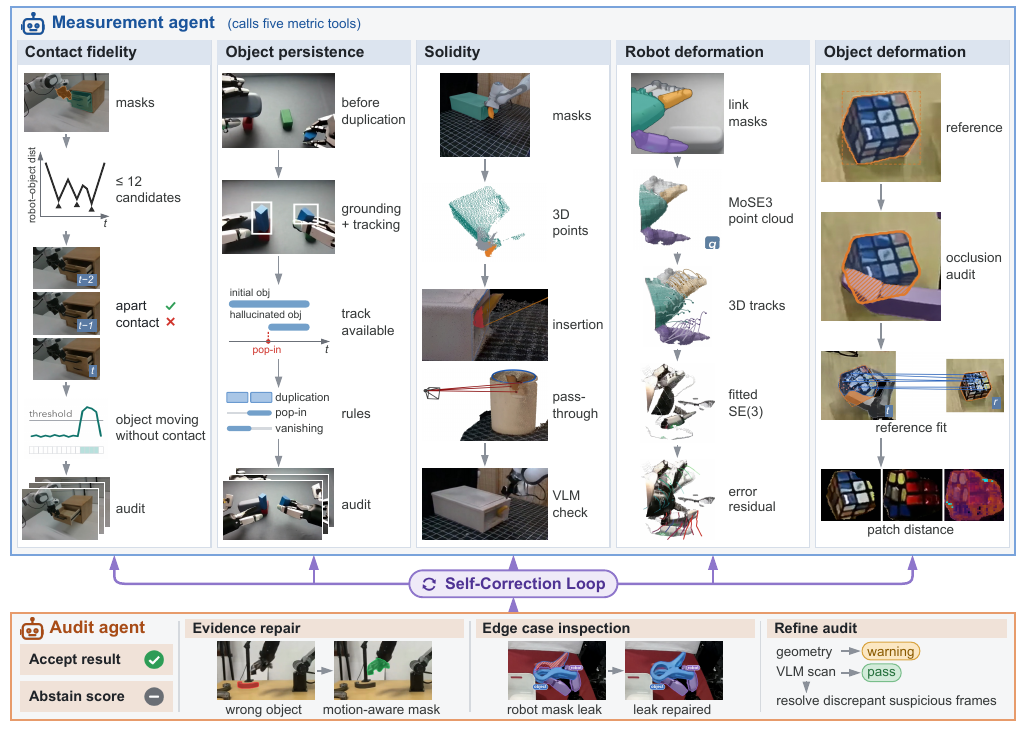}
  \caption{\textbf{\method{} runs its metrics as tools of agents.} A measurement agent calls five metric tools, one per law. An audit agent then checks the evidence behind each score, without seeing scores or labels, and either accepts the result, abstains, or starts evidence repair, edge-case inspection or audit refinement, after which the metric reruns.}
  \label{fig:method}\label{fig:agent}\label{fig:metrics}\label{fig:solidity}\label{fig:contact}\label{fig:persistence}
\end{figure*}

\subsection{Embodied World Models}
\label{sec:related_ewm}
Video generation models trained on internet-scale data learn to predict how a scene evolves~\cite{sora,wan,cogvideox}.
Conditioned on robot actions, language, or the current observation, they become embodied world models (EWMs) that let a robot agent imagine the outcome of an action before executing it~\cite{cosmos,cosmos25,cosmos3,abotphysworld,lvp,ctrlworld}.
EWMs now support model-based planning~\cite{unipi}, policy evaluation~\cite{worldeval1,worldeval2,worldgym}, and VLA training.
In each use, the agent acts on what the generated video shows, so an EWM is a reliable verifier of actions only if its rollouts are physically plausible.

\subsection{Open-Loop Visual Benchmarks}
\label{sec:related_openloop}
The recent advancement of VLMs makes them a natural choice as an intelligent evaluation tool, and a majority of recent benchmarks directly ask VLMs or learned judges whether a video is physically plausible~\cite{videophy2,worldmodelbench,pai,phyworldbench,rbench,h2rbench}.
VideoPhy-2~\cite{videophy2} and WorldModelBench~\cite{worldmodelbench} train dedicated judges to rate physical commonsense and law adherence, while RBench~\cite{rbench} and PAI-Bench~\cite{pai} score embodied videos along physical and task dimensions.
VLMs and learned judges are strong at high-level semantic understanding but remain limited in spatiotemporal reasoning and in evaluating precise geometric details.
Neither shows where or when a violation occurs.
\method{} instead grounds each judgment in spatiotemporal evidence from segmentation, tracking, and 3D geometry, combined with targeted VLM reasoning, and reports the frames where a violation occurs.

\subsection{Closed-Loop Execution Benchmarks}
\label{sec:related_sim_geo}
Closed-loop benchmarks evaluate an EWM by extracting robot actions from the generated video and executing them in a simulator, so task success measures whether the rollout is executable~\cite{robowmbench,kinebench}.
World-in-World~\cite{worldinworld} instead places world models in a closed planning loop that scores candidate actions without extracting them.
KineBench~\cite{kinebench} replaces learned inverse dynamics models with explicit 6D end-effector pose recovery.
However, these protocols are confined to environments that a simulator supports with a compatible action interface, and their verdicts inherit the noise of action extraction.
A failed execution may come from an inverse dynamics model or pose tracker that fails on unseen trajectories or deformed robots, which limits their utility on arbitrary generated videos.
\method{} closes this gap by measuring physical plausibility directly on the video, using VLM reasoning to adapt to each dynamic scene, robot type, and set of entities, without a simulator or a learned action extractor.

\subsection{3D Geometric Evaluation}
\label{sec:related_geo}
Another line of work lifts generated videos to 3D and measures geometry directly.
PDI-Bench~\cite{pdi} scores the scale, motion, and rigidity of a single tracked object, while RoboPhys-3D~\cite{robophys3d} compares rollouts with a matched simulator ground truth.
However, both measure entities one at a time: PDI-Bench never evaluates interaction physics between objects, and RoboPhys-3D hands contact and interpenetration back to a VLM.
Yet many violations in manipulation arise between entities.
\method{} closes this gap with law-specific metrics that each test one law of the robot--object interaction with geometric evidence.

\section{Method}
\label{sec:method}

To evaluate whether EWM rollouts are physically plausible, we introduce \method{}, an evidence-grounded evaluation framework with three components.
(1) Five law-specific metrics, \mContact{}, \mPersist{}, \mSolid{}, \mRobotDef{} and \mObjDef{}, where a higher score means more violation (\cref{sec:metrics}).
(2) An agentic framework that understands and adapts to diverse manipulation scenes, and reviews the evidence behind every score, repairing faulty evidence and rerunning the step (\cref{sec:agent}).
(3) RP-Bench, with 761 human-labeled generated videos and 130 real demonstrations over 239 tasks of six types (\cref{sec:rpbench}).
\Cref{fig:method} shows the pipeline.

\subsection{Evidence-grounded physical metrics}
\label{sec:metrics}

\paragraph{\mContact{}.}\label{sec:contact} 
An object should start to move only after the robot touches it. \mContact{} flags object motion that no plausible contact precedes, as \cref{fig:metric_vis}a shows. A gripper in front of an object overlaps it in the image without actually touching it, so we judge contact with a VLM on close crops. We detect object motion separately with point tracks and then pair the two in time. For motion, SAM3~\cite{sam3} masks the object, and CoTracker3~\cite{cotracker3} tracks the object contour and marks the frames in which it moves. For contact, RobotSeg~\cite{robotseg} masks the end-effector, and the set $\mathcal{C}$ holds up to 12 candidate frames where the two masks come closest. At each candidate $t$, Qwen3-VL-8B~\cite{qwen3vl} sees crops of frames $t{-}2$, $t{-}1$ and $t$ and returns the probabilities $P_C$ of contact and $P_U$ that contact is unclear. The candidate score and the contact-plausibility score of the video are
\begin{equation}
  s_t = 100\,\bigl(P_C + \tfrac12 P_U\bigr), \qquad S = \max_{t\in\mathcal{C}^{+}} s_t ,
  \label{eq:contact}
\end{equation}
where $\mathcal{C}^{+}\subseteq\mathcal{C}$ keeps the candidates that object motion follows within $[t, t{+}36]$, so one plausible contact before the motion suffices. A Gemini~\cite{gemini} audit of 16 frames gives the probability $a$ that an object moves without plausible contact. With $\rho$ the percentile rank over the evaluated videos, the violation score is $V = \tfrac12\,[\rho(a) + 1 - \rho(S)]$. If the masks, the motion or an eligible candidate are missing, $S$ takes the median over videos and the audit alone orders the video. \Cref{sec:supp_contact} gives the settings.

\paragraph{\mPersist{}.}\label{sec:persistence}
By object permanence~\cite{spelke1992}, an object should never duplicate, appear from nowhere or vanish. Tracking sees the video densely but follows only the manipulated object, while a VLM sees every object but only 16 frames. The two are complementary, so \mPersist{} uses both. Starting from the target in the first frame, GroundingDINO~\cite{groundingdino} finds boxes on every third frame, SigLIP~\cite{siglip} keeps those most similar to the target, SAM3~\cite{sam3} masks them, and matching on DINOv2~\cite{dinov2} features links the masks into tracks that bridge brief occlusions. Three events are scored on the tracks. Duplication $e_{\mathrm{dup}}$ is two target masks of similar size without overlap in one frame. Pop-in $e_{\mathrm{pop}}$ is a new target track that starts away from the original while the original is still visible. Vanishing $e_{\mathrm{van}}$ is a stable track that ends inside the image before the video ends, with no track continuing it.  A Gemini~\cite{gemini} audit of 16 frames stamped with their indices gives the probability $a$ that any object appears from nowhere or is duplicated. With $\rho$ the percentile rank as above and $k$ running over the three events, the tracking score and the violation score are
\begin{equation}
  T = \max_{k}\,\rho(e_k), \qquad V = \tfrac12\,[\rho(a) + \rho(T)] ,
  \label{eq:persist}
\end{equation}
so one clear event is enough to raise $T$. Objects other than the target thus enter $V$ only through the audit, and vanishing only through tracking. \Cref{sec:supp_persist} gives the settings.

\paragraph{\mSolid{}.}\label{sec:solidity}
An object should not penetrate a solid surface without clearing its edge. Generated videos may violate this law by letting one object pass through a solid surface or become embedded in it (\emph{insertion}), as \cref{fig:metric_vis}b shows. A penetration can last a few frames in a small part of the image, so geometry first finds the frames at risk and a VLM then judges them. SAM3~\cite{sam3} segments the moving and the static object in a zoomed crop, and MoSE3~\cite{mose3} reconstructs both as 3D point clouds. We fit surfaces to the static object and locate their free edges, the boundary an object must clear to pass the surface. On an opaque surface, we rely on occlusion, which is more stable than monocular depth. The camera ray through each lowest visible pixel $u$ of the object meets the surface at $\mathbf{x}^{*}(u)$, and while $\mathbf{x}^{*}(u)$ lies inside the free edge the object has not cleared the surface, so its progressive occlusion by that surface marks a pass-through. On a transparent surface we use depth instead, marking an uncleared object that moves from in front of the surface to on or behind it. For an insertion, with $d(\mathbf{x})$ the signed point-to-surface distance, $X_t$ the object's points in frame $t$ and $\ell$ the static object's size, the shares of the object on the two sides are
\begin{equation}
  f_t^{\pm} = \frac{1}{|X_t|}\,\bigl|\{\mathbf{x}\in X_t : \pm d(\mathbf{x}) > \tau\ell\}\bigr| ,
  \label{eq:solid}
\end{equation}
where the margin $\tau\ell$ drops points too close to the surface to trust, and a frame is at risk when both shares are substantial. Gemini~\cite{gemini} judges the frames at risk with the safe frames before them, and its penetration probability is the score. \Cref{sec:supp_solid} gives the settings.

\paragraph{\mRobotDef{}.}\label{sec:deformation}\label{sec:robot_rigidity}
Joints articulate, but individual robot links must remain rigid. Generated videos may bend, stretch or locally deform links. We identify such violations by measuring whether the observed 3D motion of each link can be explained by a single rigid transformation. We mask each robot link separately. For each link $k$, we select the earliest reference frame $q_k$ with sufficiently stable visibility and use its reconstructed point cloud as the reference geometry. MoSE3~\cite{mose3} tracks the link's 3D points $\mathbf{x}_i(t)$ in the reference camera's coordinates across the subsequent frames, and we estimate the link length $L_k$ from the principal-axis extent of the reference point cloud. For each later frame $t$, we estimate the rigid transformation $(\hat{\mathbf{R}}_t,\hat{\mathbf{T}}_t)$ that best aligns the reference points with their tracked positions by robust Kabsch fitting~\cite{kabsch1976} over dense, reliable, visible 3D points. We treat the displacement of each tracked point that the best-fitting $\mathrm{SE}(3)$ transformation cannot explain as a deformation error signal, normalized by the link length,
\begin{equation}
  r_i^k(t)=\frac{\left\|\hat{\mathbf{R}}_t\mathbf{x}_i(q_k)+\hat{\mathbf{T}}_t-\mathbf{x}_i(t)\right\|_2}{L_k}.
  \label{eq:robot_residual}
\end{equation}
To capture localized deformation while reducing sensitivity to tracking outliers, we take the 90th percentile of residuals over query points $Q_k(t)$ in each valid frame,
\begin{equation}
  s_k(t)=\operatorname{P}_{90}\left(\{r_i^k(t)\}_{i\in Q_k(t)}\right).
  \label{eq:robot_frame}
\end{equation}
Unlike the dense points used for alignment, these scoring points come from farthest-point sampling of pooled SIFT, Shi-Tomasi and grid candidates, to ensure spatial coverage. Since deformation may last only a few frames, we take the maximum frame-level score to capture brief violations,
\begin{equation}
  S_k^{\mathrm{deform}}=\max_{t\in\mathcal{T}_k}s_k(t).
  \label{eq:robot_link}
\end{equation}
The fixed reference frame also captures accumulated deformation, and the score is invariant to global rigid motion and to the global scale of the reconstruction.

\paragraph{\mObjDef{}.}\label{sec:object_rigidity}
A rigid object should keep its shape and appearance while the robot manipulates it. Generated videos may violate this law by squashing or morphing the object, as \cref{fig:metric_vis}c shows. Since manipulated objects are small and often hidden by the robot, we compare appearance instead of reconstructing 3D shape. The reference is the earliest frame that shows the unoccluded object cleanly. In ten later frames, we surgically remove the robot's pixels from the object mask and estimate the 2D rotation, scale, and translation that align the reference silhouette with the visible pixels, pairing each visible part with the same part in the reference. AnomalyDINO~\cite{anomalydino} then matches every DINOv2~\cite{dinov2} patch of the frame to its nearest patch among rotated copies of the reference crop. To capture localized deformation, we average the largest 1\% of patch distances per frame, and since deformation may be brief, we take the maximum over frames. For cups whose side facing the camera changes, newly exposed surfaces invalidate the comparison, so a VLM rates the frames against the reference instead. The final score averages this score equally with Gemini's~\cite{gemini} rating of the whole video.

\subsection{Agentic coordination and self-review}
\label{sec:design}\label{sec:agent}

\paragraph{Agentic Scene Understanding.}
World-model rollouts span diverse scenes, objects, and robot embodiments where hardcoded heuristics fail. A \emph{Measurement agent} uses general-purpose VLMs to parse task targets, infer reasonable physical properties of individual objects, and classify embodiments by general kinematics (serial manipulator vs.\ humanoid). It then extracts link masks, object masks, tracks, and 3D point clouds and runs the five metric tools of \cref{sec:metrics}.

\paragraph{Audit agent Self-Correction Loop.}
On generated edge cases that break typical evaluation assumptions, a fixed evaluation pipeline can become \emph{falsely confident}, outputting scores when it is tracking the wrong object or using a broken reference. To prevent such silent failures, an \emph{Audit agent} verifies the visual evidence behind each score before the score is final. It sees sampled frames and mask overlays but no scores or labels, and accepts the result or picks one action from the closed set in \cref{fig:method}. It never edits a score; every revised score comes from rerunning the metric.
(i)~\textbf{Refine audit} reinspects high-risk frames when the geometric and VLM scores of contact or object persistence disagree.
(ii)~\textbf{Evidence repair} ensures the pipeline tracks the true manipulated target. When the robot manipulates an object other than the task target, a single-frame detector can ground the wrong object. The agent therefore inspects frames across the rollout to identify the object that actually moves, and boxes it on its earliest unobstructed view.
(iii)~\textbf{Edge-case inspection} ensures object deformation is scored against a valid reference. Fixed masking heuristics fail when the object is initially occluded, fragmented, disappears, or bleeds into the robot mask. From mask overlays and trajectory statistics (e.g., object and robot mask overlap ratios, area growth), the agent diagnoses the failure mode and triggers targeted repairs, such as subtracting robot-mask leakage.
(iv)~\textbf{Abstention} is reserved for rollouts where valid geometric evidence cannot be recovered even after repair.

\section{RP-Bench}
\label{sec:rpbench}

RP-Bench has two purposes. It tests whether an evaluator agrees with human judgment on five violation types, and compares video generators on robot manipulation. Existing physical benchmarks assess general physical reasoning~\cite{videophy,videophy2,phyworldbench,phyground}, and robot video benchmarks collect human preferences or video-level scores~\cite{rbench,robojudge}. RP-Bench labels every generated video for the five laws of \cref{sec:metrics} and keeps each task's real demonstration as reference.

\paragraph{Tasks.}
We curate 239 tasks from 12 public robot datasets. They cover six task types: pick-and-place, pushing and pressing, multi-step tasks, bimanual coordination, tool use and pouring. Nine robot platforms of three types perform them, single-arm manipulators, a dual-arm mobile manipulator and humanoids. The camera view is first-person in 138 tasks and third-person in 101, and the scenes include 27 household scenes.

\paragraph{Videos.}
Each task comes with its real demonstration. We draft the prompt from that video and its first frame, and human annotators verify and filter all samples. Three physical-AI world models, ABot-PhysWorld~\cite{abotphysworld}, Cosmos 2.5~\cite{cosmos25} and Cosmos3~Nano~\cite{cosmos3}, and two general video generators, HunyuanVideo 1.5~\cite{hunyuanvideo15} and Wan2.2~5B~\cite{wan}, then generate 761 videos, and \cref{tab:generators} gives the count per generator. The evaluation set adds the real demonstrations of the 130 tasks that all five generators share, 891 videos in total. \Cref{sec:supp_eval} lists the source datasets.

\paragraph{Labels.}
Two annotators independently label each generated video for the five violation types defined in \cref{sec:metrics} and resolve disagreements through discussion. Before discussion they agree on 85.7\% to 94.6\% of the 761 generated videos per type, with Cohen's $\kappa$ from 0.67 to 0.86, as \cref{tab:supp_agreement} shows. For \mRobotDef{} we exclude 95 videos with an agent-identity mismatch, in which another robot arm or a human hand performs the task, since link deformation of the original robot does not apply. Seven ambiguous videos are excluded, and the 130 real demonstrations are not annotated and are treated as violation-free.

\section{Experiments}
\label{sec:exp}
\begin{table*}[t]
\centering
\caption{\textbf{On RP-Bench, \method{} scores above learned judges, prior evaluators and a frames-only VLM on all five types.} Per-video AUROC against our labels, with videos / positives per column. Avg.\ averages the five types, bold marks the best result and underline the best baseline. The last block gives ours / Gemini on MiraBench's own labels. $^\dagger$Trained on four of our generators, $^\ddagger$0.5 by construction, and \na{} marks a type the judge asks nothing about or a run not done.}
\label{tab:human}\label{tab:mirabench}
\footnotesize
\setlength{\tabcolsep}{3pt}
\setlength{\aboverulesep}{0pt}\setlength{\belowrulesep}{0pt}\renewcommand{\arraystretch}{1.05}
\begin{tabular*}{\textwidth}{@{\extracolsep{\fill}}l c ccc cc@{}}
\toprule
\multirow{2}{*}{Method} & \multirow{2}{*}{Avg.\,$\uparrow$} & \multicolumn{3}{c}{Interaction} & \multicolumn{2}{c}{Shape} \\
\cmidrule(lr){3-5}\cmidrule(l){6-7}
& & \sContact\,$\uparrow$ & \sPersist\,$\uparrow$ & \sSolid\,$\uparrow$ & \sRobotDef\,$\uparrow$ & \sObjDef\,$\uparrow$ \\
& & \noagent{884 / 311} & \noagent{880 / 177} & \noagent{884 / 154} & \noagent{767 / 480} & \noagent{802 / 247} \\
\midrule
\rowcolor{tabgray}
\multicolumn{7}{@{}l}{\textit{Frames-only VLM}} \\
Gemini 4 Argon~\cite{gemini} & \second{0.828} & \second{0.800} & \second{0.919} & \second{0.781} & \second{0.795} & \second{0.843} \\
Claude Sonnet 5.5 & 0.769 & 0.744 & 0.911 & 0.683 & 0.770 & 0.735 \\
Gemini 3.8 Flash & 0.746 & 0.728 & 0.829 & 0.654 & 0.757 & 0.764 \\
\rowcolor{tabgray}
\multicolumn{7}{@{}l}{\textit{Learned video judges}} \\
RoboJudge-9B~\cite{robojudge}$^\dagger$ & 0.669 & 0.698 & 0.729 & 0.687 & 0.609 & 0.622 \\
PhyJudge-9B~\cite{phyground} & \na & 0.609 & 0.717 & 0.627 & \na & 0.599 \\
WorldModelBench~\cite{worldmodelbench}, VILA-2B judge & \na & 0.500$^\ddagger$ & \na & 0.532 & \na & 0.574 \\
\rowcolor{tabgray}
\multicolumn{7}{@{}l}{\textit{Prior evaluators}} \\
RBench~\cite{rbench}, physical & 0.620 & 0.640 & 0.676 & 0.600 & 0.615 & 0.571 \\
PDI~\cite{pdi}, final score & 0.569 & 0.585 & 0.580 & 0.556 & 0.520 & 0.604 \\
VideoCon-Physics~\cite{videophy} & 0.547 & 0.536 & 0.553 & 0.526 & 0.592 & 0.528 \\
PAI~\cite{pai}, motion smoothness & 0.503 & 0.492 & 0.487 & 0.517 & 0.504 & 0.517 \\
\dashsep{1-7}
\textit{Best prior sub-score} & \textit{0.644} & \textit{0.640} & \textit{0.676} & \textit{0.617} & \textit{0.659} & \textit{0.627} \\
\rowcolor{tabgray}
\multicolumn{7}{@{}l}{\textit{Ours}} \\
\method{}, metrics only & 0.892 & 0.880 & 0.957 & 0.894 & 0.851 & 0.877 \\
\textbf{\method{}, agent in loop} & \best{0.894} & \best{0.889} & \best{0.957} & \best{0.894} & \best{0.851} & \best{0.881} \\
\rowcolor{tabblue}
Metrics only $-$ Gemini & +0.064 & +0.080 & +0.038 & +0.113 & +0.056 & +0.034 \\
\rowcolor{tabblue}
Metrics only $-$ best prior & +0.248 & +0.240 & +0.281 & +0.277 & +0.192 & +0.250 \\
\midrule
\rowcolor{tabgray}
\multicolumn{7}{@{}l}{\textit{Temporal localization error, median $|$reported $-$ annotated frame$|$}\,$\downarrow$ \textit{(positives with a reported frame)}} \\
\method{} / middle-frame guess & \best{3.9} / 21.5 & \best{4} / 20 & \best{5} / 31.5 & \best{3} / 20 & \best{0.5} / 23 & \best{7} / 13 \\
\midrule
\rowcolor{tabgray}
\multicolumn{7}{@{}l}{\textit{External benchmark, ours / Gemini on its own labels}} \\
MiraBench~\cite{mirabench} & \na & \best{0.763} / 0.748 & \na & \best{0.787} / 0.736 & \na & \best{0.788} / 0.780 \\
& & \noagent{80 / 8} & & \noagent{70 / 27} & & \noagent{139 / 69} \\
\bottomrule
\end{tabular*}
\end{table*}

\begin{figure*}[t]
\centering
  \includegraphics[width=\textwidth]{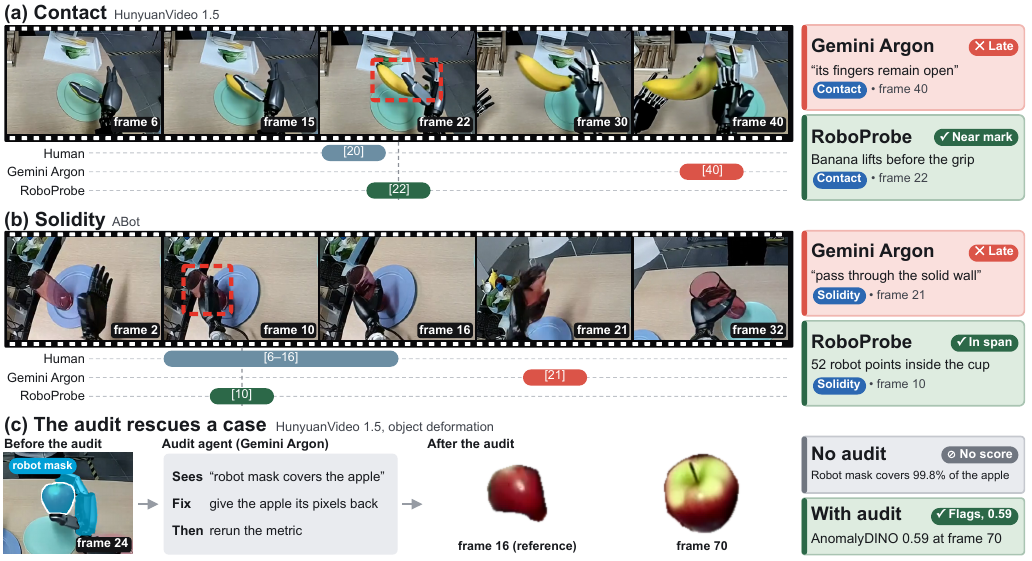}
  \caption{\textbf{\method{} localizes contact and solidity violations, and its audit rescues a missed case.} (a, b) Lanes show the human mark, the frame our metric reports and the frame Gemini 4 Argon names. Our frame comes from the geometry: in (b), robot points found inside the cup. The red box marks that frame. (c) The robot mask covers the apple, so no frame can be compared until the audit agent's edge-case inspection gives the apple its pixels back.}
  \label{fig:metric_vis}
\end{figure*}

\subsection{Setup}
\label{sec:setup}

\paragraph{Datasets.}
We test on RP-Bench from \cref{sec:rpbench}, and the header of \cref{tab:human} gives the labeled videos and positives per type. The metrics were developed on RoboWM68, a separate set of 333 videos that LVP~\cite{lvp}, Cosmos 2.5 and Cosmos 3 generate for RoboWM-Bench scenes~\cite{robowmbench}. For external validation we use MiraBench~\cite{mirabench}, which grades 186 first-person humanoid videos from four generators on 16 physical-consistency items as no, minor, clear or severe violation; clear and severe ones count as positives. On 88 videos from three of the generators we test \mContact{} on the contact-causality item and \mSolid{} on the interpenetration item, with no violation as negative and minor ones left out. We test \mObjDef{} on the colour item over 139 of the 186 videos it scores, with all other grades negative.

\paragraph{Baselines.}
The baselines form the three groups of \cref{tab:human}. The frames-only VLMs are Gemini 4 Argon~\cite{gemini}, Gemini 3.8 Flash~\cite{gemini38flash} and Claude Sonnet 5.5~\cite{claudesonnet55}, each given the same 16 uniformly sampled frames and violation-specific question. The learned judges are PhyJudge-9B~\cite{phyground}, which scores individual physical laws, WorldModelBench~\cite{worldmodelbench}, which gives a binary judgment per law, and RoboJudge-9B~\cite{robojudge}, which gives one video-level score. The prior evaluators are the RBench~\cite{rbench} physical score, VideoCon-Physics~\cite{videophy}, PAI~\cite{pai} motion smoothness and the PDI~\cite{pdi} final score. We also report the best of their 14 sub-scores per type, chosen on our labels and therefore an optimistic reference. \Cref{sec:supp_eval} gives all settings.

\paragraph{Evaluation.}
For each violation type we report the AUROC of a metric's score against the binary human label, and \cref{sec:supp_eval} gives task-grouped 95\% bootstrap confidence intervals~\cite{efron1979}. Each metric is evaluated on the labeled videos it scores, and every baseline is scored on the same labels in the same way. For \mRobotDef{} we report the mean AUROC over the 11 robot links. For temporal localization, an annotator who did not see the metrics' outputs marked the frame where each violation starts on a sample of the positive videos, or its interval for the two deformation types. We report the median distance in frames between this mark and the frame a metric reports, against always answering the middle frame.

\paragraph{Implementation.}
The metrics run on public perception models and paid Gemini audit and judge calls that cost 0.04 to 0.07 USD each. Qwen3-VL-8B~\cite{qwen3vl} runs locally, and \cref{sec:supp_eval} lists each step's model and measured cost.

\subsection{Results}
\label{sec:human}

\paragraph{Comparison.}
\Cref{tab:human} compares \method{} with VLM-based judges and prior evaluators on RP-Bench. Across the five violation types, \method{} achieves an average AUROC of 0.894, compared with 0.828 for the strongest baseline, Gemini 4 Argon, and 0.746--0.769 for the other two VLM baselines.The paired 95\% confidence intervals for the improvements exclude zero on all three interaction types. Compared with learned judges and prior evaluators, which achieve maximum AUROCs of 0.729 and 0.676, respectively, \method{} consistently achieves higher performance across the evaluated violation types. This trend also holds when comparing against the best prior sub-score for each type, selected using our evaluation labels, with improvements ranging from 0.192 to 0.281 AUROC.

\begin{table*}[t]
\centering
\definecolor{rOurs}{HTML}{1A549B}\definecolor{rGem}{HTML}{8A9097}
\def\rO#1{\textcolor{rOurs}{#1}}\def\rG#1{\textcolor{rGem}{#1}}
\caption{\textbf{Violation rates of video generators.} Each cell is the percentage of a generator's videos with the violation. Hum.\ is the share the annotators labeled, {\color{rOurs}Ours} the share our metric flags and {\color{rGem}Gem.}\ the share Gemini flags, so a reliable metric stays near Hum. Overall averages the five types, with the lowest generator rate in bold. Real demonstrations contain no violation, so their rates are false alarms.}
\label{tab:generators}
\footnotesize
\setlength{\tabcolsep}{2pt}
\setlength{\aboverulesep}{0pt}\setlength{\belowrulesep}{0pt}\renewcommand{\arraystretch}{1.3}
\begin{tabular*}{\textwidth}{@{}l@{\extracolsep{\fill}}c@{\extracolsep{\fill}\hspace{6pt}}c@{\extracolsep{0pt}\hspace{5pt}}c@{\hspace{5pt}}c@{\extracolsep{\fill}\hspace{6pt}}c@{\extracolsep{0pt}\hspace{5pt}}c@{\hspace{5pt}}c@{\extracolsep{\fill}\hspace{6pt}}c@{\extracolsep{0pt}\hspace{5pt}}c@{\hspace{5pt}}c@{\extracolsep{\fill}\hspace{6pt}}c@{\extracolsep{0pt}\hspace{5pt}}c@{\hspace{5pt}}c@{\extracolsep{\fill}\hspace{6pt}}c@{\extracolsep{0pt}\hspace{5pt}}c@{\hspace{5pt}}c@{\extracolsep{\fill}\hspace{6pt}}>{\columncolor{tabgray}[4pt][5pt]}c@{\extracolsep{0pt}\hspace{5pt}}>{\columncolor{tabgray}[0pt][5pt]}c@{\hspace{5pt}}>{\columncolor{tabgray}[0pt][0pt]}c@{}}
\toprule
Model & Videos & \multicolumn{3}{c}{\sContact} & \multicolumn{3}{c}{\sPersist} & \multicolumn{3}{c}{\sSolid} & \multicolumn{3}{c}{\sRobotDef} & \multicolumn{3}{c}{\sObjDef} & \multicolumn{3}{>{\columncolor{tabgray}[4pt][0pt]}c@{}}{Overall} \\
\cmidrule(l{2pt}){3-5}\cmidrule(l{2pt}){6-8}\cmidrule(l{2pt}){9-11}\cmidrule(l{2pt}){12-14}\cmidrule(l{2pt}){15-17}\cmidrule(l{2pt}){18-20}
& & {\scriptsize\rO{Ours}} & {\scriptsize\rG{Gem.}} & {\scriptsize Hum.} & {\scriptsize\rO{Ours}} & {\scriptsize\rG{Gem.}} & {\scriptsize Hum.} & {\scriptsize\rO{Ours}} & {\scriptsize\rG{Gem.}} & {\scriptsize Hum.} & {\scriptsize\rO{Ours}} & {\scriptsize\rG{Gem.}} & {\scriptsize Hum.} & {\scriptsize\rO{Ours}} & {\scriptsize\rG{Gem.}} & {\scriptsize Hum.} & {\scriptsize\rO{Ours}} & {\scriptsize\rG{Gem.}} & {\scriptsize Hum.} \\
\midrule
\multicolumn{17}{@{}l}{\textit{Physical-AI world models}} & & & \\
ABot-PhysWorld~\cite{abotphysworld} & 130 & \rO{31.5} & \rG{31.5} & 36.9 & \rO{16.9} & \rG{16.2} & 13.8 & \rO{32.3} & \rG{31.5} & 14.6 & \rO{50.8} & \rG{69.2} & 75.0 & \rO{41.2} & \rG{40.3} & 32.8 & \rO{34.5} & \rG{37.7} & 34.6 \\
Cosmos 2.5~\cite{cosmos25} & 132 & \rO{43.9} & \rG{47.0} & 43.2 & \rO{22.9} & \rG{21.2} & 15.9 & \rO{28.8} & \rG{35.6} & 11.4 & \rO{61.5} & \rG{76.0} & 91.7 & \rO{31.9} & \rG{39.5} & 29.4 & \rO{37.8} & \rG{43.9} & 38.3 \\
Cosmos3~Nano~\cite{cosmos3} & 239 & \rO{23.0} & \rG{27.6} & 19.7 & \rO{17.7} & \rG{15.9} & 12.6 & \rO{29.7} & \rG{27.6} & 10.5 & \rO{52.8} & \rG{67.3} & 73.8 & \rO{31.9} & \rG{30.0} & 23.7 & \rO{\best{31.0}} & \rG{\best{33.7}} & \best{28.1} \\
\midrule
\multicolumn{17}{@{}l}{\textit{General video generators}} & & & \\
HunyuanVideo~1.5~\cite{hunyuanvideo15} & 130 & \rO{51.5} & \rG{66.2} & 44.8 & \rO{38.5} & \rG{40.8} & 30.4 & \rO{38.4} & \rG{30.4} & 21.6 & \rO{40.7} & \rG{35.2} & 55.6 & \rO{48.3} & \rG{48.3} & 46.7 & \rO{43.5} & \rG{44.2} & 39.8 \\
Wan2.2~5B~\cite{wan} & 130 & \rO{73.8} & \rG{87.7} & 80.5 & \rO{55.8} & \rG{63.8} & 56.2 & \rO{66.4} & \rG{63.3} & 53.1 & \rO{69.6} & \rG{75.5} & 82.4 & \rO{53.4} & \rG{54.2} & 57.6 & \rO{63.8} & \rG{68.9} & 66.0 \\
\midrule
\multicolumn{17}{@{}l}{\textit{Real videos}} & & & \\
Real demonstrations & 130 & \rO{17.7} & \rG{8.5} & 0.0 & \rO{12.3} & \rG{0.8} & 0.0 & \rO{10.0} & \rG{3.8} & 0.0 & \rO{12.6} & \rG{15.7} & 0.0 & \rO{8.4} & \rG{13.4} & 0.0 & \rO{12.2} & \rG{8.4} & 0.0 \\
\bottomrule
\end{tabular*}
\end{table*}

\paragraph{Ablation.}
We examine the contributions of geometry-guided evidence selection and the audit agent. For \mSolid{}, we compare the same Gemini model and base prompt using either 16 uniformly sampled frames or frames selected by our geometric analysis. Geometry-guided selection increases AUROC from 0.781 to 0.894, suggesting that identifying relevant frames is important for detecting spatial violations. The audit agent primarily helps recover cases where the initial evidence is incomplete or unreliable. For \mContact{}, it repairs segmentation in 52 videos, improving AUROC on these cases from 0.791 to 0.923. For \mObjDef{}, it recovers clean reference frames in 27 videos, increasing the number of annotated violations with a geometric score from 235 to 247 out of 261. The recovered geometric scores rank all 12 additional violations above the recovered negatives. 

\paragraph{Temporal localization.}
Beyond video-level scores, \method{} identifies frames associated with physical violations. As shown in \cref{tab:human}, the median distance between the reported frame and the human annotation ranges from 0.5 to 7 frames across violation types, compared with 13 to 31.5 frames for a middle-frame baseline. Depending on the violation type, \method{} localizes 47\% to 60\% of violations within five frames of the annotation, compared with 20\% to 44\% for Gemini.

\paragraph{Generalization.}\label{sec:mirabench}
We further evaluate \method{} on MiraBench using its independent annotations. On the three evaluated categories, \method{} achieves AUROCs of 0.763, 0.787 and 0.788 for \mContact{}, \mSolid{} and \mObjDef{} on the colour item, respectively, compared with 0.748, 0.736 and 0.780 for Gemini. The largest improvement is again observed for \mSolid{}, although the gains on the other two categories are relatively modest. We also examine performance across video generators. When evaluated only on generated videos, \mContact{} and \mPersist{} achieve AUROCs of 0.889 and 0.959. Within individual generators, these metrics outperform Gemini in all 20 comparisons reported in \cref{tab:supp_pergen}. 

\subsection{Benchmarking generators}
\label{sec:generators}

\Cref{tab:generators} compares violation rates estimated by \method{} and Gemini against human annotations across five video generators. We classify violations using thresholds that maximize Youden's index on RP-Bench, applying the same procedure to Gemini. Averaged across the five violation types, the two general video generators exhibit higher violation rates than the three physical-AI world models. The overall rates estimated by \method{} preserve the human generator ranking, differing by at most 3.7 percentage points per generator, compared with 5.6 points for Gemini. \mContact{} also preserves the human ranking, while \mPersist{} and \mSolid{} do so on the 130 tasks shared by all five generators.

Performance varies across violation types. \mSolid{} detects 92\% of annotated penetration violations but overestimates violation rates, while \mRobotDef{} generally underestimates them. Because thresholds are selected on the evaluation set, the reported rates support relative generator comparisons rather than independently calibrated absolute violation frequencies. Selecting thresholds using the other generators changes estimated rates by at most 6.2 percentage points.

\section{Conclusion}
\label{sec:conclusion}

We presented \method{}, a framework for evaluating five physical properties of robot-generated videos through law-specific measurements and VLM reasoning. On RP-Bench, all five metrics outperform the reported learned judges, prior evaluators and a frames-only Gemini baseline. For \mContact{} and \mPersist{}, combining targeted measurements with VLM audits improves agreement with human labels. On the tasks all generators share, \mContact{}, \mPersist{} and \mSolid{} rank the generators as human labels do. Beyond benchmarking, the metrics could screen generated training videos or serve as rewards for fine-tuning video generators.

Our evaluation has several limitations. \mPersist{} tracks only the manipulated object, \mRobotDef{} excludes agent-identity mismatches, and \mObjDef{} excludes soft and articulated objects. All five metrics also flag 8 to 18\% of the real demonstrations, which we treat as violation-free. The external evaluation covers only \mContact{}, \mSolid{} and \mObjDef{} on MiraBench. Finally, several components rely on proprietary Gemini calls, adding cost and limiting reproducibility.

\label{lastcontentpage}

\section*{LLM Usage Statement}

The models gemini-4-argon, Gemini 3.8 Flash, Claude Opus 5.5 and Qwen3-VL-8B are components of \method{}, as \cref{sec:method,sec:supp_deform} describe. gemini-4-argon, Gemini 3.8 Flash and Claude Sonnet 5.5 are the frames-only baselines, and Qwen3.8-27B and Qwen3-VL-8B answer the RBench questions in \cref{sec:human}. The scored videos come from the five generators of \cref{sec:rpbench} and LVP. We used Claude and GPT to edit the text and to create images for the figures.

{
    \small
    \bibliographystyle{ieeenat_fullname}
    \bibliography{main}

\begin{thebibliography}{69}
\providecommand{\natexlab}[1]{#1}
\providecommand{\url}[1]{\texttt{#1}}
\expandafter\ifx\csname urlstyle\endcsname\relax
  \providecommand{\doi}[1]{doi: #1}\else
  \providecommand{\doi}{doi: \begingroup \urlstyle{rm}\Url}\fi

\bibitem[Alabdulmohsin et~al.(2023)]{sovit}
Ibrahim Alabdulmohsin et~al.
\newblock Getting {ViT} in shape: Scaling laws for compute-optimal model
  design.
\newblock In \emph{NeurIPS}, 2023.

\bibitem[{Anthropic}(2026)]{claudesonnet55}
{Anthropic}.
\newblock Claude sonnet 5.5.
\newblock \url{https://www.anthropic.com/claude/sonnet}, 2026.

\bibitem[Asim et~al.(2025)Asim, Wewer, Wimmer, Schiele, and Lenssen]{met3r}
Mohammad Asim, Christopher Wewer, Thomas Wimmer, Bernt Schiele, and Jan~Eric
  Lenssen.
\newblock {MEt3R}: Measuring multi-view consistency in generated images.
\newblock In \emph{CVPR}, pages 6034--6044, 2025.

\bibitem[Bai et~al.(2025)]{qwen3vl}
Shuai Bai et~al.
\newblock {Qwen3-VL} technical report.
\newblock \emph{arXiv preprint arXiv:2511.21631}, 2025.

\bibitem[Bai et~al.(2026)]{tavp}
Yongjie Bai et~al.
\newblock Learning to see and act: Task-aware virtual view exploration for
  robotic manipulation.
\newblock In \emph{CVPR}, 2026.

\bibitem[Bansal et~al.(2024)]{videophy}
Hritik Bansal et~al.
\newblock {VideoPhy}: Evaluating physical commonsense for video generation.
\newblock \emph{arXiv preprint arXiv:2406.03520}, 2024.

\bibitem[Bansal et~al.(2025)]{videophy2}
Hritik Bansal et~al.
\newblock {VideoPhy-2}: A challenging action-centric physical commonsense
  evaluation in video generation.
\newblock \emph{arXiv preprint arXiv:2503.06800}, 2025.

\bibitem[Benjamini and Hochberg(1995)]{benjamini1995}
Yoav Benjamini and Yosef Hochberg.
\newblock Controlling the false discovery rate: A practical and powerful
  approach to multiple testing.
\newblock \emph{Journal of the Royal Statistical Society: Series B
  (Methodological)}, 57\penalty0 (1):\penalty0 289--300, 1995.

\bibitem[Brooks et~al.(2024)Brooks, Peebles, et~al.]{sora}
Tim Brooks, William Peebles, et~al.
\newblock Video generation models as world simulators.
\newblock OpenAI Technical Report, 2024.

\bibitem[Cadene et~al.(2026)]{lerobot}
Remi Cadene et~al.
\newblock {LeRobot}: An open-source library for end-to-end robot learning.
\newblock In \emph{ICLR}, 2026.

\bibitem[Carion et~al.(2025)]{sam3}
Nicolas Carion et~al.
\newblock {SAM} 3: Segment anything with concepts.
\newblock \emph{arXiv preprint arXiv:2511.16719}, 2025.

\bibitem[Chen et~al.(2025)]{lvp}
Boyuan Chen et~al.
\newblock Large video planner enables generalizable robot control.
\newblock \emph{arXiv preprint arXiv:2512.15840}, 2025.

\bibitem[Chen et~al.(2026)]{abotphysworld}
Yuzhi Chen et~al.
\newblock {ABot-PhysWorld}: Interactive world foundation model for robotic
  manipulation with physics alignment.
\newblock \emph{arXiv preprint arXiv:2603.23376}, 2026.

\bibitem[Cheng et~al.(2026)]{mose3}
Jiahuan Cheng et~al.
\newblock {MoSE3}: Learning world-space {SE(3)} at every pixel.
\newblock In \emph{NeurIPS}, 2026.

\bibitem[Damm et~al.(2025)]{anomalydino}
Simon Damm et~al.
\newblock {AnomalyDINO}: Boosting patch-based few-shot anomaly detection with
  {DINOv2}.
\newblock In \emph{WACV}, 2025.

\bibitem[Deng et~al.(2026)]{rbench}
Yufan Deng et~al.
\newblock Rethinking video generation model for the embodied world.
\newblock \emph{arXiv preprint arXiv:2601.15282}, 2026.

\bibitem[Du et~al.(2023)]{unipi}
Yilun Du et~al.
\newblock Learning universal policies via text-guided video generation.
\newblock In \emph{NeurIPS}, 2023.

\bibitem[Efron(1979)]{efron1979}
B. Efron.
\newblock Bootstrap methods: Another look at the jackknife.
\newblock \emph{The Annals of Statistics}, 7\penalty0 (1):\penalty0 1--26,
  1979.

\bibitem[Eldar et~al.(1997)]{eldar1997}
Y. Eldar et~al.
\newblock The farthest point strategy for progressive image sampling.
\newblock \emph{{IEEE} Transactions on Image Processing}, 6\penalty0
  (9):\penalty0 1305--1315, 1997.

\bibitem[Fang et~al.(2026)]{molmoact2}
Haoquan Fang et~al.
\newblock {MolmoAct2}: Action reasoning models for real-world deployment.
\newblock \emph{arXiv preprint arXiv:2605.02881}, 2026.

\bibitem[Field and Welsh(2007)]{fieldwelsh2007}
C.~A. Field and A.~H. Welsh.
\newblock Bootstrapping clustered data.
\newblock \emph{Journal of the Royal Statistical Society Series B: Statistical
  Methodology}, 69\penalty0 (3):\penalty0 369--390, 2007.

\bibitem[{Fourier ActionNet Team} and Mu(2025)]{actionnet}
{Fourier ActionNet Team} and Yao Mu.
\newblock {ActionNet}: A dataset for dexterous bimanual manipulation.
\newblock \url{https://actionnet.pages.dev}, 2025.

\bibitem[{Gemini Robotics Team} et~al.(2025){Gemini Robotics Team},
  Choromanski, et~al.]{worldeval2}
{Gemini Robotics Team}, Krzysztof Choromanski, et~al.
\newblock Evaluating {Gemini} robotics policies in a {Veo} world simulator.
\newblock \emph{arXiv preprint arXiv:2512.10675}, 2025.

\bibitem[{Google DeepMind}(2026{\natexlab{a}})]{gemini}
{Google DeepMind}.
\newblock Gemini 4 {Argon} model card.
\newblock \url{https://deepmind.google/models/model-cards/gemini-4-argon/},
  2026{\natexlab{a}}.

\bibitem[{Google DeepMind}(2026{\natexlab{b}})]{gemini38flash}
{Google DeepMind}.
\newblock Gemini 3.8 flash model card.
\newblock \url{https://deepmind.google/models/model-cards/gemini-3-8-flash/},
  2026{\natexlab{b}}.

\bibitem[Gu et~al.(2026)]{phyworldbench}
Jing Gu et~al.
\newblock {PhyWorldBench}: A comprehensive evaluation of physical realism in
  text-to-video models.
\newblock In \emph{ICLR}, 2026.

\bibitem[Guo et~al.(2025)]{ctrlworld}
Yanjiang Guo et~al.
\newblock {Ctrl-World}: A controllable generative world model for robot
  manipulation.
\newblock \emph{arXiv preprint arXiv:2510.10125}, 2025.

\bibitem[Holland and Welsch(1977)]{hollandwelsch1977}
Paul~W. Holland and Roy~E. Welsch.
\newblock Robust regression using iteratively reweighted least-squares.
\newblock \emph{Communications in Statistics - Theory and Methods}, 6\penalty0
  (9):\penalty0 813--827, 1977.

\bibitem[Huang et~al.(2024)]{vbench}
Ziqi Huang et~al.
\newblock {VBench}: Comprehensive benchmark suite for video generative models.
\newblock In \emph{CVPR}, 2024.

\bibitem[Jiang et~al.(2026)]{robowmbench}
Feng Jiang et~al.
\newblock {RoboWM-Bench}: A benchmark for evaluating world models in robotic
  manipulation.
\newblock \emph{arXiv preprint arXiv:2604.19092}, 2026.

\bibitem[Kabsch(1976)]{kabsch1976}
W. Kabsch.
\newblock A solution for the best rotation to relate two sets of vectors.
\newblock \emph{Acta Crystallographica Section A}, 32\penalty0 (5):\penalty0
  922--923, 1976.

\bibitem[Karaev et~al.(2024)]{cotracker3}
Nikita Karaev et~al.
\newblock {CoTracker3}: Simpler and better point tracking by pseudo-labelling
  real videos.
\newblock \emph{arXiv preprint arXiv:2410.11831}, 2024.

\bibitem[Lee et~al.(2025)]{molmoact}
Jason Lee et~al.
\newblock {MolmoAct}: Action reasoning models that can reason in space.
\newblock \emph{arXiv preprint arXiv:2508.07917}, 2025.

\bibitem[Li et~al.(2025{\natexlab{a}})]{worldmodelbench}
Dacheng Li et~al.
\newblock {WorldModelBench}: Judging video generation models as world models.
\newblock \emph{arXiv preprint arXiv:2502.20694}, 2025{\natexlab{a}}.

\bibitem[Li et~al.(2025{\natexlab{b}})]{worldeval1}
Yaxuan Li et~al.
\newblock {WorldEval}: World model as real-world robot policies evaluator.
\newblock \emph{arXiv preprint arXiv:2505.19017}, 2025{\natexlab{b}}.

\bibitem[Liang et~al.(2023)]{robo360}
Litian Liang et~al.
\newblock {Robo360}: A {3D} omnispective multi-material robotic manipulation
  dataset.
\newblock \emph{arXiv preprint arXiv:2312.06686}, 2023.

\bibitem[Lin et~al.(2026)]{phyground}
Juyi Lin et~al.
\newblock {PhyGround}: Benchmarking physical reasoning in generative world
  models.
\newblock \emph{arXiv preprint arXiv:2605.10806}, 2026.

\bibitem[Liu et~al.(2023)]{groundingdino}
Shilong Liu et~al.
\newblock Grounding {DINO}: Marrying {DINO} with grounded pre-training for
  open-set object detection.
\newblock \emph{arXiv preprint arXiv:2303.05499}, 2023.

\bibitem[Liu et~al.(2026)Liu, Zhu, Zhang, Fan, Bai, and Li]{kinebench}
Zeyu Liu, Zhangzhe Zhu, Yang Zhang, Chenyou Fan, Chenjia Bai, and Xuelong Li.
\newblock {KineBench}: Benchmarking embodied world models via {IDM}-free
  kinematic grounding.
\newblock \emph{arXiv preprint arXiv:2607.19876}, 2026.

\bibitem[Liu et~al.(2021)]{swin}
Ze Liu et~al.
\newblock {Swin Transformer}: Hierarchical vision transformer using shifted
  windows.
\newblock In \emph{ICCV}, 2021.

\bibitem[Lowe(2004)]{lowe2004}
David~G. Lowe.
\newblock Distinctive image features from scale-invariant keypoints.
\newblock \emph{International Journal of Computer Vision}, 60\penalty0
  (2):\penalty0 91--110, 2004.

\bibitem[Ma et~al.(2026)]{robojudge}
Siyuan Ma et~al.
\newblock {RoboJudge}: Multimodal language models as judges for embodied video
  generation.
\newblock \emph{arXiv preprint arXiv:2610.27782}, 2026.

\bibitem[Mann and Whitney(1947)]{mannwhitney1947}
H.~B. Mann and D.~R. Whitney.
\newblock On a test of whether one of two random variables is stochastically
  larger than the other.
\newblock \emph{The Annals of Mathematical Statistics}, 18\penalty0
  (1):\penalty0 50--60, 1947.

\bibitem[Mei et~al.(2026)Mei, Huang, Ci, and Shou]{robotseg}
Haiyang Mei, Qiming Huang, Hai Ci, and Mike~Zheng Shou.
\newblock {RobotSeg}: A model and dataset for segmenting robots in image and
  video.
\newblock In \emph{CVPR}, 2026.

\bibitem[Milkey et~al.(2026)]{physicslens}
Isaiah Milkey et~al.
\newblock {PhysicsLENS}: Diagnosing physical property blindness in video
  generation models.
\newblock \emph{arXiv preprint arXiv:2610.01162}, 2026.

\bibitem[{NVIDIA} et~al.(2025{\natexlab{a}}){NVIDIA}, Agarwal, et~al.]{cosmos}
{NVIDIA}, Niket Agarwal, et~al.
\newblock Cosmos world foundation model platform for physical {AI}.
\newblock \emph{arXiv preprint arXiv:2501.03575}, 2025{\natexlab{a}}.

\bibitem[{NVIDIA} et~al.(2025{\natexlab{b}}){NVIDIA}, Ali, et~al.]{cosmos25}
{NVIDIA}, Arslan Ali, et~al.
\newblock World simulation with video foundation models for physical {AI}.
\newblock \emph{arXiv preprint arXiv:2511.00062}, 2025{\natexlab{b}}.

\bibitem[{NVIDIA} et~al.(2026){NVIDIA}, Aditi, Agarwal, et~al.]{cosmos3}
{NVIDIA}, Aditi, Niket Agarwal, et~al.
\newblock Cosmos 3: Omnimodal world models for physical {AI}.
\newblock \emph{arXiv preprint arXiv:2606.02800}, 2026.

\bibitem[{NVIDIA GEAR Lab}(2025)]{gr1100}
{NVIDIA GEAR Lab}.
\newblock {GR1-100}.
\newblock
  \url{https://huggingface.co/datasets/nvidia/PhysicalAI-Robotics-GR00T-GR1},
  2025.

\bibitem[Oquab et~al.(2023)]{dinov2}
Maxime Oquab et~al.
\newblock {DINOv2}: Learning robust visual features without supervision.
\newblock \emph{arXiv preprint arXiv:2304.07193}, 2023.

\bibitem[Quevedo et~al.(2025)]{worldgym}
Julian Quevedo et~al.
\newblock {WorldGym}: World model as an environment for policy evaluation.
\newblock \emph{arXiv preprint arXiv:2506.00613}, 2025.

\bibitem[{Qwen Team}(2026)]{qwen38}
{Qwen Team}.
\newblock {Qwen3.8-27B} model card.
\newblock \url{https://huggingface.co/Qwen/Qwen3.8-27B}, 2026.

\bibitem[Rong et~al.(2026)Rong, Shi, Ma, Cao, Wang, Zhang, Mu, Zhai, and
  Liu]{h2rbench}
Dingyi Rong, Yue Shi, Chaofan Ma, Jiezhang Cao, Zongrui Wang, Zeyu Zhang, Yao
  Mu, Guangtao Zhai, and Ning Liu.
\newblock {H2R-Bench}: Benchmarking human-to-robot manipulation video
  generation in world models.
\newblock \emph{arXiv preprint arXiv:2608.13049}, 2026.

\bibitem[Shi and Tomasi(1994)]{shitomasi1994}
Jianbo Shi and Carlo Tomasi.
\newblock Good features to track.
\newblock In \emph{CVPR}, 1994.

\bibitem[Spelke et~al.(1992)Spelke, Breinlinger, Macomber, and
  Jacobson]{spelke1992}
Elizabeth~S. Spelke, Karen Breinlinger, Janet Macomber, and Kristen Jacobson.
\newblock Origins of knowledge.
\newblock \emph{Psychological Review}, 99\penalty0 (4):\penalty0 605--632,
  1992.

\bibitem[{Tencent Hunyuan Foundation Model Team}(2025)]{hunyuanvideo15}
{Tencent Hunyuan Foundation Model Team}.
\newblock {HunyuanVideo} 1.5 technical report.
\newblock \emph{arXiv preprint arXiv:2511.18870}, 2025.

\bibitem[Unterthiner et~al.(2018)Unterthiner, van Steenkiste, Kurach, Marinier,
  Michalski, and Gelly]{fvd}
Thomas Unterthiner, Sjoerd van Steenkiste, Karol Kurach, Raphael Marinier,
  Marcin Michalski, and Sylvain Gelly.
\newblock Towards accurate generative models of video: A new metric \&
  challenges.
\newblock \emph{arXiv preprint arXiv:1812.01717}, 2018.

\bibitem[{Wan Team} et~al.(2025){Wan Team}, Wang, et~al.]{wan}
{Wan Team}, Ang Wang, et~al.
\newblock {Wan}: Open and advanced large-scale video generative models.
\newblock \emph{arXiv preprint arXiv:2503.20314}, 2025.

\bibitem[Wang et~al.(2026)Wang, Chen, Xu, Li, Zeng, Chou, Lu, Peng, Jiao, and
  Claudel]{robophys3d}
Tianyi Wang, Jiazhou Chen, Yiming Xu, Xiangyu Li, Tianyi Zeng, Chih-Hsien Chou,
  Ning Lu, Liang Peng, Junfeng Jiao, and Christian Claudel.
\newblock {RoboPhys-3D}: A comprehensive embodied world model evaluation via
  {3D} reconstruction.
\newblock \emph{arXiv preprint arXiv:2608.28718}, 2026.

\bibitem[Wu et~al.(2026)]{pdi}
Jiaxin Wu et~al.
\newblock Quantitative video world model evaluation for geometric-consistency.
\newblock \emph{arXiv preprint arXiv:2605.15185}, 2026.

\bibitem[Wu et~al.(2025{\natexlab{a}})]{robomind}
Kun Wu et~al.
\newblock {RoboMIND}: Benchmark on multi-embodiment intelligence normative data
  for robot manipulation.
\newblock In \emph{Robotics: Science and Systems (RSS)}, 2025{\natexlab{a}}.

\bibitem[Wu et~al.(2025{\natexlab{b}})]{robocoin}
Shihan Wu et~al.
\newblock {RoboCOIN}: An open-sourced bimanual robotic data collection for
  integrated manipulation.
\newblock \emph{arXiv preprint arXiv:2511.17441}, 2025{\natexlab{b}}.

\bibitem[Yang et~al.(2026)]{mirabench}
Tianzhuo Yang et~al.
\newblock {MiraBench}: Evaluating action-conditioned reliability in robotic
  world models.
\newblock \emph{arXiv preprint arXiv:2605.29360}, 2026.

\bibitem[Yang et~al.(2025)Yang, Teng, Zheng, Ding, Huang, Xu, Yang, Hong,
  Zhang, Feng, Yin, Zhang, Wang, Cheng, Xu, Gu, Dong, and Tang]{cogvideox}
Zhuoyi Yang, Jiayan Teng, Wendi Zheng, Ming Ding, Shiyu Huang, Jiazheng Xu,
  Yuanming Yang, Wenyi Hong, Xiaohan Zhang, Guanyu Feng, Da Yin, Yuxuan Zhang,
  Weihan Wang, Yean Cheng, Bin Xu, Xiaotao Gu, Yuxiao Dong, and Jie Tang.
\newblock {CogVideoX}: Text-to-video diffusion models with an expert
  transformer.
\newblock In \emph{ICLR}, 2025.

\bibitem[Yue et~al.(2025)Yue, Huang, Liao, Chen, Zhou, Chen, Yao, and
  Ren]{ewmbench}
Hu Yue, Siyuan Huang, Yue Liao, Shengcong Chen, Pengfei Zhou, Liliang Chen,
  Maoqing Yao, and Guanghui Ren.
\newblock {EWMBench}: Evaluating scene, motion, and semantic quality in
  embodied world models.
\newblock \emph{arXiv preprint arXiv:2505.09694}, 2025.

\bibitem[Zhai et~al.(2023)]{siglip}
Xiaohua Zhai et~al.
\newblock Sigmoid loss for language image pre-training.
\newblock In \emph{ICCV}, 2023.

\bibitem[Zhang et~al.(2025)Zhang, Jiang, Dai, Lu, Uzunoglu, Zhang, Wei, Wang,
  Patel, Liang, Khashabi, Peng, Chellappa, Shu, Yuille, Du, and
  Chen]{worldinworld}
Jiahan Zhang, Muqing Jiang, Nanru Dai, Taiming Lu, Arda Uzunoglu, Shunchi
  Zhang, Yana Wei, Jiahao Wang, Vishal~M. Patel, Paul~Pu Liang, Daniel
  Khashabi, Cheng Peng, Rama Chellappa, Tianmin Shu, Alan Yuille, Yilun Du, and
  Jieneng Chen.
\newblock {World-in-World}: World models in a closed-loop world.
\newblock \emph{arXiv preprint arXiv:2510.18135}, 2025.

\bibitem[Zhao et~al.(2025)]{humanoideveryday}
Zhenyu Zhao et~al.
\newblock Humanoid everyday: A comprehensive robotic dataset for open-world
  humanoid manipulation.
\newblock \emph{arXiv preprint arXiv:2510.08807}, 2025.

\bibitem[Zhou et~al.(2026)]{pai}
Fengzhe Zhou et~al.
\newblock {PAI-Bench}: A comprehensive benchmark for physical {AI}.
\newblock In \emph{CVPR}, 2026.

\end{thebibliography}
}

\clearpage
\setcounter{page}{1}
\nocite{humanoideveryday,physicslens,robojudge,mannwhitney1947,benjamini1995}%
\maketitlesupplementary

\noindent This supplement continues the numbering of the main paper. Sections 1--6, Figures 1--3, Tables 1--2 and numbered equations refer to the main paper, and citation numbers refer to its reference list.

\section{Evaluation paradigms}
\label{sec:supp_paradigms}
\Cref{tab:paradigms} compares the capabilities of existing evaluation paradigms with \method{}.

\begin{table}[t]
\centering
\footnotesize
\setlength{\tabcolsep}{3pt}
\renewcommand{\arraystretch}{1.1}
\begin{tabularx}{\linewidth}{@{}>{\raggedright\arraybackslash}Xcccc@{}}
\toprule
\textbf{Paradigm / Benchmark} &
\begin{tabular}[b]{@{}c@{}}\textbf{Scene-}\\\textbf{Agnostic}\end{tabular} &
\begin{tabular}[b]{@{}c@{}}\textbf{Geometric}\\\textbf{Constraints}\end{tabular} &
\begin{tabular}[b]{@{}c@{}}\textbf{Interaction}\\\textbf{Physics}\end{tabular} &
\begin{tabular}[b]{@{}c@{}}\textbf{Temporal}\\\textbf{Diagnosis}\end{tabular} \\
\midrule
VLM Judges (VideoPhy-2, RBench, PAI-Bench)~\cite{videophy2,rbench,pai} & \cmark & \xmark & \xmark & \xmark \\
Closed-Loop Evaluation (RoboWM-Bench, World-in-World)~\cite{robowmbench,worldinworld} & \xmark & \cmark & \xmark & \xmark \\
3D Consistency (MEt3R, PDI-Bench)~\cite{met3r,pdi} & \cmark & \cmark & \xmark & \cmark \\
Reference-Grounded 3D (RoboPhys-3D)~\cite{robophys3d} & \xmark & \cmark & \xmark & \xmark \\
\midrule
\rowcolor{tabblue}
\textbf{\method{} (Ours)} & \cmark & \cmark & \cmark & \cmark \\
\bottomrule
\end{tabularx}
\caption{\textbf{\method{} has all four capabilities.} Capabilities of existing evaluation paradigms for embodied world models.%
}
\label{tab:paradigms}
\end{table}

\paragraph{\method{} joins VLM reasoning with geometric evidence.}
No earlier paradigm in \cref{tab:paradigms} covers all four columns. VLM judges work on any scene but measure no geometry, and simulator and 3D methods measure geometry but need a supported environment or do not test the interaction between robot and object.
In \method{}, the agents and VLM audits handle open-world scenes and unusual cases, and the 2D and 3D geometry gives the measurements and the frames where a violation happens.
\method{} therefore needs no simulator and reports, frame by frame, violations of the interaction between robot and object as well as of the consistency of each object and robot link.

\section{Metric details}
\label{sec:supp_method}

\subsection{\mSolid{}}
\label{sec:supp_solid}

\textbf{Identifying the interacting pair.} \mSolid{} evaluates whether solid objects interact in a physically plausible way, so the first step is to find which solid objects interact. Gemini 3.8 Flash~\cite{gemini38flash} looks at six frames spread across the video and identifies the pair of interest: the object that the robot moves and, when there is one, the static object that it approaches, contacts or enters. Seeing the whole video lets it find the pair even when the two objects become visible at different times. Gemini then selects an early frame in which both are visible and marks each object in it. From these marks, SAM3~\cite{sam3} segments the two objects and tracks them forward and backward through the video. For the robot--object test, the pair is the manipulating hand or gripper and the object it moves, and we track the hand in the same way.

\noindent \textbf{Reconstructing the pair in 3D.} To resolve small gaps between the moving object and the static object, we reconstruct both in a common coordinate frame and at a common scale with MoSE3~\cite{mose3}. We use the same tight crop in every frame so that the two objects occupy more of the reconstruction grid while retaining static context for camera estimation. We compared zoomed and full-frame reconstructions on more than 30 development videos and measured both on seven of them, where the zoom lowered the static scene's frame-to-frame jitter from 0.47\% to 0.35\% of the camera distance and the share of smeared points at the object's edge from 13\% to 7\% (medians; lower on all seven). We chose MoSE3 because, on RoboWM development videos, the relative geometry of the interacting objects appeared more consistent in its reconstructions than in those of VGGT (Wang et al., CVPR 2025). The crop covers the tracked pair across the video, with padding of half the manipulated object's reference box size. It is at least 208\,px wide and expands toward 40\% static context, up to the full frame width. We resize to 518\,px in width, using Lanczos for upsampling (at most $2.5\times$) and area resampling for downsampling. A separate reconstruction pass includes hand masks for the robot--object test.

\noindent \textbf{Fitting surfaces and free edges.} We fit the static object's surfaces in 3D on the frames where it is clear of the moving object and the robot: a cylindrical wall and a top for a round object, a top and sides for a box-like one. Gemini 3.8 Flash then checks the fitted flat surfaces against the video. It sees one frame with the surfaces drawn on it and says, for each, whether the object is solid there or open, as at the top of an open bin. Open surfaces are removed, and the edges they leave are the free edges. The remaining surfaces are impenetrable: the moving object can reach their other side only by clearing a free edge.

\noindent \textbf{Choosing occlusion or depth evidence.} An object can remain visible on either side of glass or clear plastic, so visibility alone is an unreliable cue for crossing a transparent surface. We use the pixelwise glass predictions of L+GNet (Ojala et al., arXiv:2603.03718, 2026) to flag likely transparent surfaces and test crossings there using relative 3D depth. We use occlusion when the static object hides the moving object and the tested surface is opaque and can occlude it from the camera's viewpoint. Otherwise, we use relative depth.

\noindent \textbf{Insertion threshold.} In \cref{eq:solid}, $\ell$ is the size of the static object: we set $\ell=H$ and $\tau=0.10$, where $H$ is the static object's extent above its support plane (99th percentile of its points). A point therefore counts on a side only when it lies more than $0.10H$ from the surface. $X_t$ holds the moving object's reconstructed points within the tested surface's extent and must contain at least 20. Both shares are substantial when $f_t^{+}\ge 0.10$ and $f_t^{-}\ge 0.10$ in three consecutive reconstructed frames while the object's lowest point has not cleared the free edge. A pass-through by depth requires at least half of $X_t$ more than $0.15H$ in front of the surface and later at least half at or behind it for three frames, without clearing the free edge in between.

\noindent \textbf{Frames for the judge.} The geometry selects frames that show the interaction going from no penetration to penetration. Each frame gets a risk in $[0,1]$: 1 where a test is met, otherwise the fraction of its threshold reached. For the object--object pair and for the robot--object pair we take up to six frames each: the last safe frame, the approach, the first risky frame and the frame before it, and the frames that follow. The robot--object risk counts robot points more than $0.2r$ inside the manipulated object's fitted body, with $r$ its inradius. These frames are cropped to twice the reconstruction window, so the interaction fills the image. The rest of the 16 frames are full views farthest in time from them. Gemini 4 Argon, at temperature 0, receives the 768\,px frames in temporal order with frame numbers and timestamps.

\noindent \textbf{Development set.} We developed \mSolid{} on 48 labelled videos: 41 RoboWM videos and 7 videos that we generated with Wan~3.0 Video Prime from prompts asking for a penetration. Wan~3.0 Video Prime is a separate model from the Wan2.2~5B generator of RP-Bench, and none of the 48 videos is in RP-Bench. On the 20 RoboWM videos used to compare frame selection (9 positives), with the judge and base prompt fixed, geometry-guided frames raise AUROC from 0.636 with 16 uniform frames to 0.843.

\subsection{\mContact{}}
\label{sec:supp_contact}

The object name is chosen once per task and shared by the task's three videos. For the 130 tasks with HunyuanVideo~1.5 and Wan2.2~5B videos, the same selector runs once per task on those two videos and the real video. Qwen3-VL-8B~\cite{qwen3vl} sees nine images, the frames at 15\%, 50\% and 85\% of each generator's video, with a long side of 640\,px. The videos of the evaluation set come without object names. The object name is part of the benchmark protocol and is fixed before any metric runs. The selector sees only these frames, the dataset name, the file name and a task hint built from it. It sees no label, score or audit output, and its prompt tells it not to judge physical correctness (\cref{sec:supp_prompts}), so the name only fixes which object the metric tracks. The rest of the metric uses only the name, so another source, such as the instruction of a single rollout, can supply it. The frames-only Gemini needs no name, since its \mContact{} prompt asks about any object in the scene. For motion, CoTracker3 follows 60 contour points of the object, cancelling camera motion with background points, and a frame is moving above $\max(0.002, \mathrm{median}+6\,\mathrm{MAD})$. Candidate times use the 2D mask distance per frame, which is 0 when the masks overlap and otherwise the smallest pixel distance divided by the object radius. The candidates are the closest frame in each of six time bins and the six closest frames overall, without duplicates, so a video has up to 12 candidate times $t$. Qwen3-VL-8B rates the crops at $t{-}2$, $t{-}1$ and $t$. Each crop is a square around the union of the object and gripper masks with 80\,px padding and a side of at least 360\,px. The candidate search ignores motion as well, and the prompt tells the VLM not to use it. The motion window looks forward from $t$, because pairing backward would instead reward a gripper that catches up with an object after it moved.

The percentile ranks of the fusion are computed over all 891 videos of the evaluation set. Ranking each video instead only against the videos of the four other task folds or of the other generators gives AUROCs of 0.880 and 0.889.

With the audit agent in the loop, EVIDENCE\_REPAIR gives a geometric score to 52 more videos, all labelled and 20 positive.

\paragraph{Remaining \mContact{} settings.}
Every frame is decoded, and SAM3, RobotSeg and CoTracker3 receive the frames at the video's native resolution, each model resizing them internally, and the masks are stored at that resolution. The violation frame that \mContact{} reports is compared with the onset marked by the annotator (\cref{sec:supp_timing}), in \cref{tab:human} and \cref{tab:supp_timing}. A video whose geometric score abstains (no target grounding, at the motion step, or no candidate time) has no geometric frame, and \cref{tab:supp_timing} gives it the frame of the audit.

\subsection{\mPersist{}}
\label{sec:supp_persist}

Tracking looks for the object name that the \mContact{} naming step picks and takes the target in the first frame as its reference. Detection runs on every third frame and on the last one. GroundingDINO SwinB~\cite{groundingdino,swin} proposes at most 12 boxes with a low box threshold of 0.10 and non-maximum suppression at 0.5, and SigLIP so400m~\cite{siglip,sovit} keeps the three boxes most similar to the reference. SAM3~\cite{sam3} turns them into masks from box prompts and DINOv2-base~\cite{dinov2} gives each an appearance feature. A mask that lies at least 90\% inside another counts as the same object. The greedy matcher links objects in nearby frames at a cost of $0.75\,(1 - \text{similarity}) + 0.25\,d$, with $d$ the distance between box centers, requires a similarity of at least 0.25 and bridges gaps of up to six frames, so a brief occlusion does not count as vanishing.

The three rules use the following constants. Duplication scores the strongest frame with two target-like masks of similar size and no overlap. Pop-in needs a new target-like track that starts mid-video while the original is present, overlaps it by at most 10\%, lies at least one object size away and coexists with it for at least six frames. Vanishing needs a stable track, seen at least five times and alive for at least 12 frames, that ends at least six frames before the video ends, inside the frame rather than at its edge, with no track continuing it. Each rule's score becomes a percentile rank over the 887 videos of the evaluation set that \mPersist{} scores, because the rules have different scales, and the tracking score is the largest of the three ranks.

The \mPersist{} audit stamps each frame with its frame number, uses a long side of 768\,px and runs at temperature 0. Its question covers objects that appear from nowhere or are duplicated and leaves out disappearance. Objects in the first frame, normal reveals, the robot, shadows and lighting do not count, and a call without a valid answer is marked incomplete, never 0. The answer gives a probability, a frame, an object and a reason, and the audit costs about 0.07~USD per video. Both ranks are computed within the 887 scored videos of the evaluation set. Ranked only against held-out videos in three ways, \mPersist{} reaches 0.956, 0.959 and 0.959. The fusion has no learned weights, and its equal weights were never tuned.

\paragraph{Remaining \mPersist{} settings.}
Four NVIDIA GR1 videos get no tracking score, because GroundingDINO finds no target in their first frame and tracking then has no reference, and the fused score needs both parts. All four are labelled, so \mPersist{} is evaluated on 880 of the 884 labelled videos. SAM3 takes the GroundingDINO boxes as prompts, and detection and masking receive the sampled frames at native resolution, each model resizing them internally. The violation frame that \mPersist{} reports is the frame of the rule with the highest rank: the strongest duplication frame, the first frame of the popped-in track, or the last frame in which the vanishing object is seen. The audit's frame does not change it. To pick the rule for the frame, the three ranks are computed within the tracking run that processed the video; the evaluation set was tracked in two runs, of 501 and 390 videos.

\subsection{\mRobotDef{}}
\label{sec:supp_deform}

\textbf{Which links we test.} A robot is built from rigid links connected by joints: the joints move, and each link keeps its shape. \mRobotDef{} tests this on three links that every manipulator has, whatever its number of joints.

\noindent The upper arm and the forearm are the two long links that meet at the elbow. They carry the hand to its position and give the arm its reach: the two links measure 0.316\,m and 0.384\,m of the Franka Panda's 0.855\,m reach, 0.425\,m and 0.392\,m of the UR5e's 0.850\,m, 0.289\,m and 0.343\,m of the xArm~6's 0.700\,m, and 0.285\,m and 0.251\,m of the Piper's 0.626\,m. As the largest rigid bodies of the arm, they cover the most pixels and give the rigid fit the most points. The short parts around the joints mainly rotate and are not tested. On a Franka, the upper arm is link~2 and the forearm is link~5.

\noindent The palm is the rigid base of the end effector, to which the fingers are attached, and the link closest to the manipulated object. On a two-finger gripper, it is the gripper body.

\noindent On a humanoid, each side has a forearm and a palm; in first-person views we add the thumb and the index finger, and in views that the VLM classifies as third-person, the upper arm. With the three links of a single arm, \mRobotDef{} covers 13 links, and one definition applies to every robot in RP-Bench. Third-person humanoid views are rare in RP-Bench: the two humanoid upper arms carry a score and a label on 5 and 10 videos, with 1 and 2 positives, so the mean AUROC of \cref{tab:human} is taken over the other 11 links.

\noindent \textbf{Classifying the kinematic structure.} A robot is a serial manipulator (one arm on a base) or a humanoid (a torso with two arms), and the two have different links and examples. For the videos of five source datasets, NVIDIA GR1, ActionNet, RoboMIND, TAVP and the RoboCOIN stack-block task, the robot type follows from the dataset. For the others, one Gemini 3.8 Flash~\cite{gemini38flash} call classifies the robot from the first, middle and last frames, the generation prompt and a description of both structures. On the 133 videos of ABot-PhysWorld, Cosmos~2.5 and Cosmos3~Nano where this call decided the robot type, it was correct on all 133. A second call checks that the robot of the first frame is still the one acting (\cref{sec:supp_coverage}).

\noindent \textbf{Anchor frame.} All prompts are placed on one frame, the anchor, and SAM3 propagates the masks from it in both directions. The anchor is the frame where the robot's structure is most complete and clear, because that is where a box separates one link from the next. Gemini 3.8 Flash picks it from three frames shortlisted by link visibility, pose, sharpness and overlap for an arm, and from ten evenly spaced frames for a humanoid.

\noindent \textbf{Per-link masks.} Each link is masked on the anchor frame in three steps. Claude Opus 5.5 draws a box around every visible link, guided by human-drawn boxes on example images of the same viewing direction. The examples show robots of a similar kinematic structure and are chosen by viewing direction alone, so they need not show the robot of the video. Gemini 3.8 Flash places positive and negative points inside each box, guided by a human example of the same link. SAM3 segments the link from the box and the points, and Opus 5.5 reviews the mask: it accepts it, or gives new points for one more SAM3 pass. For an arm, a link that matches a rigid-motion cluster of MoSE3 takes its box and points from that cluster. In first-person views, Flash also places and checks the points of the thumb and the index finger. No target video is annotated.

\noindent \textbf{Mask quality.} The pipeline produces link masks and a score for 823 of the 891 videos, 314 single arms and 509 humanoids from all 12 source datasets. A robotics expert inspected the link masks of 150 randomly sampled videos and judged 148 (98.67\%) to match the robot's rigid links.

\noindent \textbf{Scoring settings.} The reference frame $q$ of a link is the first frame whose mask area reaches a third of the link's 90th-percentile area for three consecutive frames, or the anchor frame if none does, and a link with fewer than 64 mask pixels at $q$ is skipped. The candidate points are the union of SIFT~\cite{lowe2004} keypoints, Shi--Tomasi corners~\cite{shitomasi1994} and a uniform grid, and farthest-point sampling~\cite{eldar1997} reduces them to $n = \max(100, A/64.84)$ for a mask area $A$. MoSE3 runs once per reference frame at a long side of 518\,px and returns 3D trajectories and visibility in the camera frame of $q$, and for humanoids it re-tracks the thumb and index finger in a zoomed window. The depth filter uses the depth of the MoSE3 points at $q$ to remove sparse outliers by a 32-nearest-neighbour density with a MAD rule, depth islands detached from the link and pixels on depth edges, and a frame needs 80 retained pixels and half of the mask. The $\mathrm{SE}(3)$ fit uses the dense, depth-filtered interior link pixels visible at $t$, while the residual is read at the $n$ query points. It is Kabsch with Cauchy weights in iteratively reweighted least squares~\cite{hollandwelsch1977}, with at most 10 iterations, and needs $\max(32, 0.3N)$ visible pixels of the $N$ link pixels that span half the link length along its principal axis. The residual is $r_i(t) = \lVert R_t x_i(q) + T_t - x_i(t) \rVert / L$, with $L$ the 2nd to 98th percentile extent along that axis at $q$, and a frame value needs at least five points.

\subsection{\mObjDef{}}
\label{sec:supp_objdef}

\textbf{Why we compare appearance at selected frames.} The robot moves an object by holding it, so the hand covers the object during the manipulation that \mObjDef{} evaluates. On the 802 RP-Bench videos that \mObjDef{} scores (87,018 frames), the hand hides at least 20\% of the object's silhouette at some frame in 55.1\% of the videos and at least 50\% in 30.0\%, and more than 5\% of it in 27.8\% of the frames after the object starts to move. Following the same surface points through the video, as \mRobotDef{} does on a link (\cref{sec:supp_deform}), needs those points in view throughout. We therefore compare the object's appearance in ten later frames with one early reference frame, on the pixels the hand does not cover.

\noindent \textbf{Occlusion test.} Pixel areas decide whether the hand hides the object in frame $t$. $O_t$ is the object mask and $G_t$ the hand mask, the union of the palm and finger masks of \mRobotDef{}. The expected silhouette $E$ is the object mask of the most recent frame clear of the hand, shifted onto frame $t$. The hidden share is $|(E \setminus O_t) \cap G_t| / |E|$: only pixels of $E$ that are missing from the object mask and lie under the hand count, so the hand must have taken the object's place. A frame is occluded when the hidden share reaches 0.2, the hand accounts for at least 0.6 of the missing pixels $E \setminus O_t$, and at least 0.2 of the object's CoTracker3~\cite{cotracker3} points lie under the hand, for two consecutive frames or more.

\noindent \textbf{Frames and pairing.} The ten frames follow quotas of 2, 2, 2 and 4 over the last four fifths of the video, by largest visible area, and always include the first frame after each occlusion. The visible pixels $O_t \setminus G_t$ are paired with the reference by the rotation, scale and shift that lay the reference silhouette over them with an IoU of at least 0.75, or by mapping the box of $E$ onto the box of the reference object. When both apply, the one closer to a rigid 3D fit of the object's MoSE3~\cite{mose3} tracks is kept. A frame is paired when at least 80\% of its visible pixels land on the reference object. This pairs 90.5\% of the 87,018 frames of the 802 videos, and the 3D fit alone pairs 48.4\%.

\noindent \textbf{Score.} AnomalyDINO~\cite{anomalydino} uses DINOv2 ViT-B/14~\cite{dinov2} at 448\,px, eight rotations of the reference crop and the cosine distance. A cup-like container whose side facing the camera changes is detected by one Gemini 3.8 Flash~\cite{gemini38flash} call on the reference and six later views. Flash then rates each later frame against the reference from 0 to 1, and the video takes the highest rating. The rating takes the place of the AnomalyDINO score directly, and both are 0 for an object identical to the reference. On the 439 videos of ABot-PhysWorld, Cosmos~2.5 and Cosmos3~Nano scored this way, AnomalyDINO spans 0.24 to 0.87 (393 videos) and Flash 0.15 to 0.95 (46 videos). Gemini 4 Argon's rating, from 16 frames of the whole video, is divided by 100 onto this scale before the average.

\subsection{Agent}
\label{sec:supp_agent}

Every metric runs in two settings: with the audit agent in the loop and without it (\cref{tab:human}). Without the agent, the five metrics score each video directly. With the agent in the loop, the audit agent inspects the evidence behind a video's score. It sees only the geometric evidence, sampled frames with mask overlays, and never a score, verdict or label. It accepts the evidence or picks one action of \cref{sec:agent}; the two below repair the evidence and rerun the metric. A revised score always comes from rerunning the metric.

EVIDENCE\_REPAIR verifies the two masks that \mContact{} measures from: the manipulated object and the end effector. Gemini 4 Argon sees five frames of the video, the task prompt and the state of the two masks, and picks one action from a closed set: box the object, re-mask the robot, both, or abstain. It also names the object and the earliest frame that shows it clearly. Gemini 3.8 Flash boxes the object on that frame and SAM3 tracks the box in both directions. To re-mask the robot, SAM3 segments the whole robot, and this mask takes the place of the end-effector mask, so contact by any part of the robot counts. \mContact{} then reruns unchanged on these object and robot masks, with the same target phrase.

EDGE\_CASE\_INSPECTION verifies the two conditions that \mObjDef{} measures from: a reference frame that shows the whole object free of the robot, and later frames whose object mask lies on the object. Gemini 4 Argon receives the geometry of both masks as numbers and as images. A summary gives the frame where the object first moves, the growth of the object's mask area, and the frames where a mask is empty or the robot mask covers the object. A table gives, for every fifth frame and every frame shown, the object and robot pixels, the share of the object under the robot mask, the visible object pixels and the silhouette-fit IoU against the reference. Up to 14 frames follow, each shown plain and with the object outlined and the robot mask filled: frame 0, the reference, the frames where a mask changes and frames across the video. Argon names the cause with its frame numbers and chooses repairs from a closed set: move the reference to the earliest frame that shows the whole object; re-mask the robot from Flash boxes of every robot part; mark frames without a robot as valid; return the object's pixels from the robot mask to the object; or re-mask the object from a Flash box. SAM3 tracks every new box, and the code places a new reference on the earliest whole, robot-free frame. The occlusion test, the MoSE3 pass and the pairing then rerun on the repaired masks, and \mObjDef{} scores the video as any other. In both actions the agent decides only where the evidence is; whether a violation occurs is left to the metric. \cref{sec:supp_agent_run} gives one complete execution.

REFINE\_AUDIT is the third action, for \mContact{} and \mPersist{}, whose scores fuse a geometric or tracking score with a Gemini audit of 16 uniformly sampled frames. When the two disagree on a video, the agent picks the event frame to inspect, and the Gemini audit reruns on frames around it before the metric reruns.

\section{Evaluation details}
\label{sec:supp_eval}

\subsection{RP-Bench source datasets}
Each task is one demonstration of a source dataset, and its first frame and text prompt condition the generators. NVIDIA GR1~\cite{gr1100} contributes 92 tasks, MolmoAct household~\cite{molmoact,molmoact2} 27, RoboCOIN~\cite{robocoin} 21, Robo360~\cite{robo360} 20, RoboMIND~\cite{robomind} 18, ActionNet~\cite{actionnet} 15, SO-101 and PiperX 10 each, UR5e 9, TAVP~\cite{tavp} 8 and the two Piper datasets 5 and 4, 239 in total. The SO-101, UR5e and Piper datasets are community datasets on Hugging Face recorded with LeRobot~\cite{lerobot}. The GR1 and ActionNet humanoids, the ten Tien Kung humanoid tasks of RoboMIND and the dual-arm mobile robot of RoboCOIN give 138 first-person tasks, and the other 101 tasks show a single arm from a third-person camera, among them the Franka arms of RoboMIND and TAVP, the xArm6 of Robo360 and the SO-101, Piper, PiperX and UR5e arms. When a demonstration has several cameras, its real video is the camera whose first frame best matches the generation input. The 130 shared tasks are the tasks that ABot-PhysWorld, Cosmos~2.5 and Cosmos3~Nano all cover, and HunyuanVideo~1.5 and Wan2.2~5B were run on these tasks. Cosmos~2.5 has 2 more tasks and Cosmos3~Nano 109 more.

The selection covers diverse robots and scenes, and tasks at three levels of difficulty. For GR1, Robo360 and TAVP we take every available video of the robot. MolmoAct, SO-101, RoboMIND and eleven RoboCOIN repositories contribute one or two episodes per distinct task. A demonstration is kept when the robot is clearly visible in its first ten frames, and a demonstration with several steps is cut to its first complete manipulation. The scenes range from laboratory tabletops to home kitchens, bathrooms and living rooms. By their prompts, we assign each task one of three levels: simple actions, pick-and-place tasks that place an object onto a surface, into a container or with a precise fit, and tasks that reorient an object, use a tool or pour, take several steps or need both hands. 26 tasks involve an articulated object and 18 a deformable one.

\subsection{Video generation protocol}
Every generator receives the task's first frame, resized and cropped to its input size, and the text prompt, and every video uses seed 0. ABot-PhysWorld runs 50 steps with guidance 5.0 at 864$\times$480 or 656$\times$480. Cosmos~2.5 runs 35 steps with guidance 7.0, and Cosmos3~Nano 35 steps with guidance 6.0, both at 1280$\times$704 or 960$\times$704. Wan2.2~5B runs 50 steps with guidance 5.0 at the Cosmos sizes, with its default negative prompt, and HunyuanVideo~1.5 50 steps with guidance 6.0 at 480p. A portrait input swaps width and height.

The number of sampling steps and the guidance scale of every generator are the official default inference settings released by its developers.

A human reviewer checks every input before generation. A first frame is kept when the robot is large enough in the image, mostly inside the camera view and clearly lit, so that every generator starts from a frame that shows the robot it has to move. A prompt that describes a long sequence of steps is shortened to one or two actions, for example pressing one or two buttons instead of several.

Cosmos3~Nano was run on all 239 tasks, Cosmos~2.5 on 132 and ABot-PhysWorld on 130, and HunyuanVideo~1.5 and Wan2.2~5B on the 130 tasks that the other three share. NVIDIA GR1, RoboMIND, ActionNet, RoboCOIN, SO-101, PiperX, MolmoAct household, UR5e, TAVP, Robo360 and two Piper datasets contribute 176, 54, 45, 41, 30, 30, 27, 27, 24, 20, 15 and 12 of the videos of ABot-PhysWorld, Cosmos~2.5 and Cosmos3~Nano. The 130 tasks that HunyuanVideo~1.5 and Wan2.2~5B also render, each with its real video, come from NVIDIA GR1, RoboMIND, ActionNet, SO-101, PiperX, RoboCOIN, UR5e, TAVP and the two Piper datasets, 41, 18, 15, 10, 10, 10, 9, 8, 5 and 4 of them.

\subsection{Human annotation protocol}
The annotators label each video for the five laws, following the definitions of \cref{sec:metrics}.

\noindent \textbf{\mContact{}.} A video is positive when the interaction between the robot and an object is physically implausible: the object starts to move or accelerates on its own, without the robot applying a force to it.

\noindent \textbf{\mPersist{}.} A video is positive when an object duplicates, appears from nowhere or vanishes while in view.

\noindent \textbf{\mSolid{}.} A video is positive when two solid objects penetrate each other, or when the robot hand, itself a solid object, penetrates another solid.

\noindent \textbf{\mRobotDef{}.} The annotator takes the robot of the first frame as the undeformed robot and gives one label per link. A link is positive when it does not keep its shape and appearance through the video. Changes of viewpoint, motion, rotation and shadow do not count as a deformation. On a single arm, the labelled links are the palm of the gripper, the forearm and the upper arm. On a humanoid, the palm, the forearm and the upper arm of each arm are labelled, and in first-person views, where the hands are large, the thumb and the index finger of each hand as well.

\noindent \textbf{\mObjDef{}.} The annotator takes the manipulated object of the first frame as the reference. A video is positive when the object does not keep its shape and appearance through the video, again apart from changes of viewpoint, motion, rotation and shadow.

An undecided annotator marks the label as unsure.

\begin{center}
\begin{minipage}{\columnwidth}
\centering
\captionsetup{type=table}
\caption{\textbf{Inter-annotator agreement on RP-Bench.} Labels of the two annotators on the 761 generated videos, before discussion and before any exclusion. Agreement is the share of videos with the same label.}
\label{tab:supp_agreement}
\footnotesize
\setlength{\tabcolsep}{2pt}
\begin{tabular*}{\columnwidth}{@{\extracolsep{\fill}}lccccc@{}}
\toprule
Violation type & A pos. & B pos. & Both pos. & Agreement (\%) & Cohen's $\kappa$ \\
\midrule
\sContact & 345 & 319 & 295 & 90.3 & 0.803 \\
\sPersist & 191 & 190 & 170 & 94.6 & 0.856 \\
\sSolid & 152 & 197 & 130 & 88.3 & 0.671 \\
\sRobotDef & 512 & 509 & 456 & 85.7 & 0.676 \\
\sObjDef & 287 & 256 & 228 & 88.6 & 0.751 \\
\bottomrule
\end{tabular*}
\end{minipage}
\end{center}

\Cref{tab:supp_agreement} gives the agreement of the two annotators before they resolved disagreements. All five $\kappa$ lie in the substantial (0.61--0.80) or almost-perfect (above 0.80) range of Landis and Koch (1977). Seven generated videos whose annotator marked the label as unsure are left out of \mContact{}, \mPersist{} and \mSolid{}, leaving 884 labelled videos.

The real videos count as free of violations.

\subsection{Coverage and abstention}
\label{sec:supp_coverage}
\mContact{} tests that an object starts to move only after the robot touches it, so its geometric score applies when a robot manipulates an object that stays in the video, which holds for 583 of the 884 labelled videos. The other 301 videos take the median geometric score in the fusion, so \mContact{} covers all 884. In 123 of them the robot does not meaningfully execute the task: it does not approach the correct object, or it does not set the object in motion. In 94 the target object disappears or deforms beyond recognition, so no target can be grounded. Of the remaining 84, 35 show a new robot arm or a human hand doing the task, and the other 49 have no grounded target or no object motion after a candidate time. \mSolid{} also covers all 884, and \mPersist{} covers 880: it abstains on four NVIDIA GR1 videos (\cref{sec:supp_persist}) whose target is not identified in the first frame.

\mRobotDef{} is evaluated on 767 of the 891 videos. It leaves out 95 agent mismatches, in which a new robot arm or a human hand does the task while the robot of the first frame never moves, so its link deformation does not apply; 22 videos on which the two annotators still disagree after discussion, because a large change of viewpoint or lighting makes the degree of deformation ambiguous; and 7 videos that our metric does not score.

\mObjDef{} is scored on the 802 videos our metric scores. It abstains when the manipulated object deforms by nature, a soft object such as a towel or a cable in 30 videos and an articulated one such as a drawer in 51, so the law of a rigid shape does not apply. Of the other 8 videos, the object itself disappears in 6, which is an \mPersist{} violation, 1 has no object mask, and the annotator leaves 1 unlabelled as ambiguous. \mRobotDef{} abstains when its agent check (\cref{tab:supp_calls}) flags an agent mismatch, when no valid mask is found or when no link can be scored, with no VLM fallback. \mObjDef{} abstains for a non-rigid object or no grounding or valid mask. Every frames-only Gemini call of \mContact{} and \mPersist{} returned a score, so no video is missing from those baselines.

An agent mismatch leaves the other four metrics scored, and their robot masks are built so that the robot that acts in the video is the one they measure. For \mObjDef{}, the robot mask tells where the robot hides the object. On a serial manipulator, RobotSeg detects every gripper on every tenth frame and on the last one, and the SAM3 tracker follows each detected gripper forward and backward through the video. The gripper that touches the object on the frames where the object moves, and moves in the same direction, is taken as the end effector that handles the object. On any other robot, SAM3 on the text ``robot'' masks the robot from two seed frames, forward and backward, and the union of its instances is the robot mask. \mSolid{} uses the robot mask only to tell where the robot hides the surfaces and the objects, built the same way per structure: RobotSeg's robot mask followed by the SAM3 tracker on a serial manipulator, and SAM3 on the text ``robot'' on every frame on any other robot. \mContact{} measures from the end-effector mask, and when the audit agent finds that this mask does not cover the robot, EVIDENCE\_REPAIR re-masks the robot with SAM3 on the text ``robot'' (\cref{sec:supp_agent}).

\subsection{Baselines}

\paragraph{Frames-only VLMs.}
The main frames-only VLM is gemini-4-argon, which receives 16 frames sampled uniformly over the video at 768 px width. Gemini 3.8 Flash~\cite{gemini38flash} and Claude Sonnet 5.5~\cite{claudesonnet55} receive the same frames and the same prompts. For \mContact{}, Gemini returns the probability that an object moves without plausible contact, and for \mPersist{} the probability that an object appears from nowhere or is duplicated. Disappearance is rare among the \mPersist{} positives of our development set, RoboWM68, so, to keep the frames-only VLM focused, the question leaves out vanishing, as in the audit of our \mPersist{} (\cref{sec:supp_persist}); our metric scores vanishing through tracking (\cref{sec:persistence}). An object vanishes in 4 of the 163 reviewed \mPersist{} positives of RP-Bench. Leaving these videos out moves the AUROC of Gemini from 0.919 to 0.921 and leaves ours at 0.957, so the comparison in \cref{tab:human} does not rest on them. For \mSolid{}, Gemini at temperature 0 gets the rule that two solids never share space, sees the frames at 768 px width and returns the probability that one solid passes through another.

For \mRobotDef{}, one call per video sends the 16 uniform frames plus the reference frame of each scored link, each labelled with its frame number and time. The prompt names the robot type and locates every link by a box and an interior point in its reference frame. It states that robot links are rigid and move only at their joints, and asks how much each link's shape changes during the video. Bending, stretching, melting, merging, gaining or losing parts, and turning into a different part all count. Joint motion, occlusion, viewpoint change, blur and a change of colour or lighting alone do not. Gemini returns a degree from 0 to 100 per link at the worst moment, anchored at 0, 25, 50, 75 and 100, and the degree divided by 100 is the link's score.

For \mObjDef{}, Gemini sees the 16 uniform frames, with the object named and located by a box and an interior point in the first frame, and without the generation prompt. The prompt states that an object keeps its shape and appearance unless something acts on it. It asks for the degree, from 0 to 100, of any change in shape or appearance that the object's material and what acts on it cannot explain. Seeing the object from another side, a newly visible part, occlusion, transparency, lighting, blur and the whole object vanishing do not count, and an unjudgeable answer scores 0.

The \mContact{} and \mPersist{} calls are also the audits of our two fused metrics, and the \mObjDef{} rating enters our \mObjDef{} score (\cref{sec:supp_objdef}), so the gain over Gemini measures what our evidence adds.

\paragraph{Prior evaluators.}
Before computing AUROC, every prior score is oriented so that a larger value means more error. PAI~\cite{pai} reports four sub-scores, and motion smoothness is the best of them in every column of \cref{tab:human}. PAI motion smoothness saturates near 1, so its differences are about 0.001 to 0.002 in absolute units, and its ranking carries the signal while no stable threshold exists yet. The best-sub-score row takes, per column, the highest AUROC among the 14 sub-scores of RBench~\cite{rbench}, VideoCon-Physics~\cite{videophy}, PAI and PDI~\cite{pdi}, so it is chosen on the same labels it is scored on. PDI is scored with the object as its target. RBench's three questions are answered by Qwen3.8-27B~\cite{qwen38} with thinking off. With this judge RBench physical reaches 0.620 on average in \cref{tab:human}, and with Qwen3-VL-8B~\cite{qwen3vl} 0.529 to 0.537 on the evaluation set. The WorldModelBench judge answers no to its \mContact{} question for every video, so all its \mContact{} scores tie and its AUROC is 0.5. A cell of \cref{tab:human} is empty where a method asks nothing about the type: PhyJudge-9B has no call about robot links, and WorldModelBench no question about persistence or robot links.

\paragraph{Scoring and tests.}
\mRobotDef{} is the mean of the per-link AUROCs over 11 links, where a video-level score counts for each link of its video.

\begin{table*}[t]
\centering
\caption{\textbf{Without the real videos, our metrics stay above Gemini on all five types.} AUROC with its 95\% interval on the labelled generated videos, with the fusions ranked over the generated videos only. Videos and positives are those our metric scores. The last column is the paired 95\% interval of ours minus Gemini.}
\label{tab:supp_generated}
\footnotesize
\setlength{\tabcolsep}{2pt}
\begin{tabular*}{\textwidth}{@{\extracolsep{\fill}}l cc ccc@{}}
\toprule
Violation type & Videos & Positives & Ours & Gemini & Ours $-$ Gemini \\
\midrule
\sContact & 754 & 311 & \best{0.892} [0.865, 0.915] & 0.780 [0.746, 0.813] & $[+0.084, +0.138]$ \\
\sPersist & 750 & 177 & \best{0.956} [0.941, 0.969] & 0.912 [0.885, 0.937] & $[+0.024, +0.065]$ \\
\sSolid & 754 & 154 & \best{0.885} [0.856, 0.913] & 0.759 [0.716, 0.800] & $[+0.084, +0.170]$ \\
\sRobotDef & 640 & 480 & \best{0.859} [0.839, 0.875] & 0.776 [0.743, 0.803] & $[+0.052, +0.116]$ \\
\sObjDef & 683 & 247 & \best{0.876} [0.845, 0.905] & 0.831 [0.795, 0.864] & $[+0.028, +0.064]$ \\
\bottomrule
\end{tabular*}
\end{table*}

\begin{table*}[t]
\centering
\caption{\textbf{Within every generator, our metrics score above Gemini.} AUROC of ours / Gemini on the generated videos of each generator, with the videos and positives that ours scores in parentheses. For \sRobotDef{}, the parentheses give the videos, the videos with a deformed link and the number of links averaged, those with at least five positive and five negative videos in the row. $^*$No RoboWM68 video comes from this generator; the 7 videos used to develop \mSolid{} were generated with Wan~3.0 Video Prime and are not in RP-Bench (\cref{sec:supp_solid}). A pooled row ranks its videos together. Scores are those of the agent-in-loop row of \cref{tab:human}, with the fusions ranked over all 891 videos.}
\label{tab:supp_pergen}
\footnotesize
\setlength{\tabcolsep}{2pt}
\begin{tabular*}{\textwidth}{@{\extracolsep{\fill}}l ccccc@{}}
\toprule
Generator & \sContact & \sPersist & \sSolid & \sRobotDef & \sObjDef \\
\midrule
ABot-PhysWorld$^*$ & 0.836\,/\,0.683 (130 / 48) & 0.947\,/\,0.843 (130 / 18) & 0.866\,/\,0.756 (130 / 19) & 0.887\,/\,0.857 (120 / 90, 11) & 0.869\,/\,0.829 (119 / 39) \\
Cosmos 2.5 & 0.798\,/\,0.662 (132 / 57) & 0.926\,/\,0.872 (131 / 20) & 0.816\,/\,0.562 (132 / 15) & 0.870\,/\,0.749 (96 / 88, 9) & 0.952\,/\,0.870 (119 / 35) \\
Cosmos3~Nano & 0.904\,/\,0.807 (239 / 47) & 0.934\,/\,0.928 (237 / 30) & 0.903\,/\,0.708 (239 / 25) & 0.854\,/\,0.753 (214 / 158, 10) & 0.855\,/\,0.816 (207 / 49) \\
HunyuanVideo~1.5$^*$ & 0.921\,/\,0.756 (125 / 56) & 0.955\,/\,0.920 (125 / 38) & 0.851\,/\,0.832 (125 / 27) & 0.851\,/\,0.646 (108 / 60, 8) & 0.805\,/\,0.762 (120 / 56) \\
Wan2.2~5B$^*$ & 0.885\,/\,0.731 (128 / 103) & 0.977\,/\,0.851 (127 / 71) & 0.883\,/\,0.777 (128 / 68) & 0.856\,/\,0.736 (102 / 84, 10) & 0.909\,/\,0.859 (118 / 68) \\
\midrule
Pooled, $^*$ & 0.893\,/\,0.775 (383 / 207) & 0.966\,/\,0.901 (382 / 127) & 0.882\,/\,0.806 (383 / 114) & 0.878\,/\,0.791 (330 / 234, 11) & 0.862\,/\,0.820 (357 / 163) \\
Pooled, Cosmos & 0.870\,/\,0.748 (371 / 104) & 0.931\,/\,0.906 (368 / 50) & 0.873\,/\,0.656 (371 / 40) & 0.854\,/\,0.753 (310 / 246, 11) & 0.893\,/\,0.840 (326 / 84) \\
\bottomrule
\end{tabular*}
\end{table*}

\subsection{Intervals and tests}

For \mContact{}, \mPersist{} and \mSolid{}, the bootstrap~\cite{efron1979,fieldwelsh2007} resamples whole tasks, so the videos of a task stay together. It uses $B{=}2000$ draws, and paired differences reuse the same draws. For \mRobotDef{} and \mObjDef{}, the bootstrap also resamples whole tasks, with $B{=}2000$ draws and seed 0, and for \mRobotDef{} each draw recomputes the mean of the per-link AUROCs over the links with at least five positive and five negative videos. The paired 95\% intervals of the gain over Gemini are $[+0.054,+0.105]$ for \mContact{}, $[+0.017,+0.058]$ for \mPersist{}, $[+0.074,+0.153]$ for \mSolid{}, $[+0.026,+0.086]$ for \mRobotDef{} and $[+0.020,+0.050]$ for \mObjDef{}, and the interval of the audit agent's gain on \mContact{}, $[+0.002,+0.017]$, lies above zero.

On the 52 \mContact{} videos whose evidence the agent repairs, 20 of them positive, the AUROC rises from 0.791 to 0.923. With the agent in the loop, \mObjDef{} scores 27 more videos, 12 of them positive, and it ranks all 12 above the 15 negatives.

The task-grouped intervals behind \cref{sec:human} are $[0.85, 0.90]$ for our \mContact{} metric and $[0.77, 0.83]$ for Gemini on 884 videos of the evaluation set with 311 positives. For \mPersist{}, ours spans $[0.94, 0.97]$ on the 880 videos it scores with 177 positives, and Gemini $[0.90, 0.94]$ on the same videos. For \mSolid{}, with 154 positives, ours spans $[0.87, 0.92]$ and Gemini $[0.74, 0.82]$. For \mRobotDef{}, on 767 videos with 480 positives, the mean over the 11 links spans $[0.83, 0.87]$ for ours and $[0.77, 0.82]$ for Gemini. For \mObjDef{}, on 802 videos with 247 positives, ours spans $[0.85, 0.90]$ with the metrics only and $[0.85, 0.91]$ with the agent in the loop, and Gemini $[0.81, 0.87]$. Within the evaluation set, on the 583 labelled videos with 182 positives where the geometric \mContact{} score does not abstain, it reaches 0.803 alone, against 0.762 for Gemini on the same videos, and the full metric 0.883, and on the 301 abstaining videos the full metric reaches 0.849. RoboJudge was trained on videos of ABot-PhysWorld, Cosmos3~Nano, Wan2.2~5B and HunyuanVideo~1.5, four of our five generators.

\subsection{External dataset details}
From MiraBench~\cite{mirabench}, \cref{tab:human} uses the contact, interpenetration and colour items, which map to \mContact{}, \mSolid{} and \mObjDef{}. Positives are grade C or D on these items, against grade A for the first two and grades A and B for colour, and \mContact{} ran on 88 of its videos, those of three of the four generators. \Cref{tab:human} counts, for each item, the videos graded A, C or D on it, since grade B (minor) is left out: 80 videos with 8 positives for \mContact{} and 70 videos with 27 positives for \mSolid{}. Our \mObjDef{} law covers the appearance as well as the shape of a rigid object (\cref{sec:object_rigidity}), so MiraBench's colour item falls under it. On MiraBench, \mObjDef{} scores 139 videos and abstains on 47 with a soft or articulated object, 43 and 4 of them. \Cref{tab:supp_mira} gives the AUROC of \mObjDef{} against MiraBench's colour, size and material items on the videos it scores, 0.788, 0.808 and 0.803.

\begin{center}
\begin{minipage}{\columnwidth}
\centering
\captionsetup{type=table}
\caption{\textbf{\mObjDef{} on three appearance items of MiraBench.} AUROC of our \mObjDef{} score against each item, with grades C and D as positives and grades A and B as negatives.}
\label{tab:supp_mira}
\footnotesize
\setlength{\tabcolsep}{2pt}
\begin{tabular*}{\columnwidth}{@{\extracolsep{\fill}}lccc@{}}
\toprule
MiraBench item & Videos & Positives & Ours \\
\midrule
Colour & 139 & 69 & 0.788 \\
Size & 139 & 75 & 0.808 \\
Material & 136 & 76 & 0.803 \\
\bottomrule
\end{tabular*}
\end{minipage}
\end{center}

\subsection{Temporal localization details}
\label{sec:supp_timing_details}
The temporal annotator marked onsets on a web page that steps through the video frame by frame and has a button for a video without a visible onset. The instructions read: for \mContact{}, mark the first frame where an object moves or changes without a plausible contact, for example it starts moving before the gripper touches it, or moves while it is not held; for \mPersist{}, mark the first frame where an object appears from nothing, is duplicated, or vanishes while in view; for \mSolid{}, mark the first frame where an object passes through, or becomes embedded in, a solid object or surface. The onset videos are random samples with a fixed seed: 80 \mContact{} and 80 \mPersist{} positives drawn among those on which both our metric and Gemini report a frame, of which 75 and 78 received an onset. For \mSolid{}, the onset set holds 40 positives. The \mContact{} and \mPersist{} samples compare both frames on the same videos, and \cref{tab:supp_timing} covers every \mContact{}, \mPersist{} and \mSolid{} positive with an onset mark. For \mPersist{}, the pool holds the positives with an extracted tracking frame at sampling time. We restrict the \mContact{} and \mPersist{} samples in this way to pair the two methods. A frame error exists only when a method reports a frame, so an unpaired sample would compare ours and Gemini's on different videos. The pairing leaves out the positives on which one method reports no frame, which \cref{tab:supp_timing} includes.

For \mRobotDef{} and \mObjDef{}, the annotator marks an interval instead of an onset. A deformation often develops gradually, while a \mContact{} violation starts at the frame in which the object begins to move, so an interval describes it better than a single frame. The annotator marks the interval in which the deformation is most severe, not the earliest one. Intervals are marked on the 480 \mRobotDef{} positives and the 247 \mObjDef{} positives of \cref{tab:human}.

\subsection{Development set construction and calibration}
RoboWM-Bench~\cite{robowmbench} is a closed-loop benchmark that executes a generated video as robot actions in a simulated copy of its scene. We build our development set on it to study how such a benchmark works and how well, and we generate the videos from RoboWM-Bench's own 68 inputs, each a first frame and an instruction, so that other baselines run on the same development videos. RoboWM68 combines 68 videos from LVP~\cite{lvp}, 61 from Cosmos 2.5~\cite{cosmos25} and 68 from each of three Cosmos 3~\cite{cosmos3} seeds, 333 videos in total.

The inputs cover eight tasks with 37 distinct instructions on a Franka arm. Closing a drawer, pressing a button, picking, pushing and pulling are single-step tasks, and discarding into a bin, placing on a plate and putting into a drawer are multi-step tasks, which RoboWM-Bench scores by sub-goals. The cubes, cups, banana and tape roll are rigid objects, and the drawer and the button are articulated.

We annotated the videos over several rounds, and in each round the annotator writes a description of the errors in the video. The first round also gives each LVP and Cosmos 2.5 video a main error type and a severity, and the second round records binary subtypes of task, contact, geometry and penetration errors for all generators. The violation types of our five laws, for example deformation and penetration, were identified step by step from these observations.

We map the finer labels of RoboWM68 onto our five types. \mSolid{} is any penetration, including penetration of an object by the robot, \mContact{} is any contact error, and \mPersist{} is the duplication or disappearance of an object, while \mRobotDef{} and \mObjDef{} keep their own labels. On RoboWM68, robot and object deformation hold 219 and 97 of the 333 videos. \Cref{tab:supp_dev} gives the number of positive and negative videos of each type in the annotation of LVP, Cosmos~2.5 and the first Cosmos~3 seed.

\begin{center}
\begin{minipage}{\columnwidth}
\centering
\captionsetup{type=table}
\caption{\textbf{Composition of the development set.} Positive / negative videos per violation type in the annotation of LVP, Cosmos~2.5 and the first Cosmos~3 seed, 194 videos without the input whose prompt does not match its first frame.}
\label{tab:supp_dev}
\footnotesize
\setlength{\tabcolsep}{2pt}
\begin{tabular*}{\columnwidth}{@{\extracolsep{\fill}}lcccc@{}}
\toprule
Violation type & LVP & Cosmos 2.5 & Cosmos 3 & Total \\
\midrule
\sContact & 26 / 41 & 33 / 27 & 25 / 42 & 84 / 110 \\
\sPersist & 6 / 61 & 5 / 55 & 9 / 58 & 20 / 174 \\
\sSolid & 13 / 54 & 10 / 50 & 9 / 58 & 32 / 162 \\
\sRobotDef & 48 / 19 & 37 / 23 & 50 / 17 & 135 / 59 \\
\sObjDef & 16 / 51 & 19 / 41 & 28 / 39 & 63 / 131 \\
\bottomrule
\end{tabular*}
\end{minipage}
\end{center}

We developed \mContact{} and \mPersist{} on labelled RoboWM68 videos. The \mContact{} geometric score was first developed on 46 of them with 11 positives, and its rank average with the audit uses equal weights that were never tuned. \mSolid{} was developed on the 48 videos of \cref{sec:supp_solid}, 41 of them from RoboWM, and its frame selection compared on 20 of the RoboWM videos with 9 positives.

\subsection{Implementation and cost}

Among the API calls, the \mContact{} and \mPersist{} audits each cost about 0.07 USD per video, about 62 USD for the 891 videos of the evaluation set, and one \mSolid{} judge call costs about 0.04 USD.

Masks come from SAM3~\cite{sam3} and RobotSeg~\cite{robotseg}, \mContact{} tracks motion with CoTracker3~\cite{cotracker3} and runs Qwen3-VL-8B~\cite{qwen3vl} locally to name its target and check contact, and \mPersist{} links GroundingDINO~\cite{groundingdino}, SigLIP~\cite{siglip} and DINOv2~\cite{dinov2} detections. \mSolid{} lifts its masks to 3D with MoSE3~\cite{mose3} and calls Gemini once per video. Both deformation metrics use MoSE3, and \mObjDef{} also CoTracker3 and AnomalyDINO~\cite{anomalydino}. Their router, light mask calls and \mObjDef{} fallback run on Gemini 3.8 Flash~\cite{gemini38flash}, and the \mObjDef{} rating on gemini-4-argon. The link boxes and mask reviews of \mRobotDef{} run on Claude Opus 5.5, about 4.2 calls per video (\cref{sec:supp_deform}). Checkpoints and code revisions are in \cref{sec:supp_versions}.

\subsection{Models and runtime}
On one NVIDIA H200, \mContact{} processes a chunk of 24 videos in under 30 minutes. Per video, it makes one Gemini audit call and scores up to 12 candidate times with Qwen3-VL-8B, three crops each, from the likelihoods of the answer letters without generating text, and it names its target once per task in each selector run. \mPersist{} makes one Gemini audit call of about 17.7k input tokens and no other VLM call, and the \mSolid{} judge is one call per video.

On one NVIDIA L40S, \mPersist{} processes a chunk of 24 videos in 9.5 to 16 minutes, 34\,s per video over 255 videos. On one NVIDIA H100 PCIe per run, \mRobotDef{} takes 2.2 GPU-minutes per video, 7.1 hours for 195 videos of HunyuanVideo~1.5, Wan2.2~5B and real robots, and its MoSE3 tracking reserves at most 34.3\,GiB per video. Over 358 such videos it makes 7.8 Gemini 3.8 Flash calls per video, for the agent check, the robot type, the anchor frame, the viewing direction and the points of each link, and 4.2 Claude Opus 5.5 calls. \mObjDef{} reuses the robot masks and takes 75\,s per video, 8.2 hours for 390 videos, with 2.5 Flash calls per video for the object box on the first frame, the check for an object with different sides, the points of the gripper fingers and the rating. \mSolid{} takes 97 H200-minutes for 70 MiraBench videos, 83\,s per video, and makes 493 Flash calls for them, 7.0 per video: one routes the robot, about two choose the frame that shows both objects, one describes the task, one boxes the manipulated object, one finds the static object and one checks the fitted surfaces.

With the audit agent in the loop, EVIDENCE\_REPAIR makes one Gemini 4 Argon call per inspected video and, on 78 of its 90 videos, seven Flash calls: one object box, four checks of the moved object and two motion-aware boxes. \mContact{} then runs on the repaired masks of 89 videos in four chunks, 25 to 27 minutes each on one H200 and 31 minutes on one L40S. EDGE\_CASE\_INSPECTION makes one Argon call per inspected video and at most two Flash box calls.

\section{Additional results}
\label{sec:supp_results}

\subsection{Results on generated videos only}
\label{sec:supp_generated}

\Cref{tab:supp_generated} repeats the evaluation of \cref{tab:human} without the real videos, with the fusions ranked over the generated videos only. Ours is above Gemini on all five types and every paired interval of ours minus Gemini lies above zero, so none of the gains depends on the real videos. Without the agent, \mContact{} reaches 0.882. With the ranks of \cref{tab:human}, which run over all 891 videos, \mContact{} reaches 0.890 and 0.880 on the same videos, and the other types keep their values. Ranked instead only against the videos of the other generators, as in \cref{sec:human}, \mContact{} and \mPersist{} reach 0.889 and 0.959 (\cref{sec:supp_contact,sec:supp_persist}).

\subsection{Average precision and results per generator}
\label{sec:supp_pergen}

Average precision gives the same order as AUROC. Ours and Gemini reach 0.824 and 0.636 on \mContact{}, 0.855 and 0.709 on \mPersist{}, 0.583 and 0.384 on \mSolid{} and 0.788 and 0.707 on \mObjDef{}, against positive rates of 0.352, 0.201, 0.174 and 0.308. \Cref{tab:supp_pergen} splits the AUROC of all five types by generator, and ours is above Gemini in all 25 cells, among them the 20 cells of \mContact{}, \mPersist{}, \mSolid{} and \mObjDef{} over the five generators. On the three generators that RoboWM68 does not contain, the paired intervals of ours minus Gemini are $[+0.080, +0.157]$ for \mContact{}, $[+0.040, +0.093]$ for \mPersist{}, $[+0.025, +0.127]$ for \mSolid{}, $[+0.044, +0.127]$ for \mRobotDef{} and $[+0.014, +0.072]$ for \mObjDef{}.

\subsection{\mRobotDef{} per link}
\label{sec:supp_robot_links}

\begin{table*}[t]
\centering
\caption{\textbf{On humanoid robots, our per-link AUROC is above Gemini on all eight links.} \mRobotDef{} AUROC per humanoid link on the 767 evaluated videos. Scored links are the links both methods score, as in \cref{tab:human}. All annotated links are every link with a human label, where a link a method does not score takes the lowest score. Pos and Neg count the positive and negative videos of each link.}
\label{tab:supp_robot_links}
\footnotesize
\setlength{\tabcolsep}{2pt}
\begin{tabular*}{\textwidth}{@{\extracolsep{\fill}}l ccc ccc@{}}
\toprule
 & \multicolumn{3}{c}{Scored links} & \multicolumn{3}{c}{All annotated links} \\
\cmidrule(lr){2-4}\cmidrule(l){5-7}
Link & Pos / Neg & Ours & Gemini & Pos / Neg & Ours & Gemini \\
\midrule
Left palm & 107 / 337 & \best{0.920} & 0.871 & 109 / 361 & \best{0.909} & 0.862 \\
Left thumb & 95 / 303 & \best{0.913} & 0.877 & 108 / 362 & \best{0.826} & 0.802 \\
Left index finger & 86 / 316 & \best{0.903} & 0.882 & 100 / 370 & \best{0.799} & 0.789 \\
Left forearm & 24 / 306 & \best{0.883} & 0.757 & 28 / 442 & \best{0.810} & 0.730 \\
Right palm & 191 / 197 & \best{0.792} & 0.752 & 200 / 270 & \best{0.816} & 0.774 \\
Right thumb & 155 / 177 & \best{0.849} & 0.710 & 192 / 278 & \best{0.765} & 0.673 \\
Right index finger & 153 / 191 & \best{0.804} & 0.662 & 178 / 292 & \best{0.773} & 0.675 \\
Right forearm & 64 / 266 & \best{0.821} & 0.716 & 76 / 394 & \best{0.766} & 0.702 \\
\midrule
Average, humanoid (8 links) & & \best{0.860} & 0.778 & & \best{0.808} & 0.751 \\
Average, all 11 links & & \best{0.851} & 0.795 & & \best{0.804} & 0.765 \\
\bottomrule
\end{tabular*}
\end{table*}

\Cref{tab:supp_robot_links} splits the \mRobotDef{} AUROC of \cref{tab:human} by link. 470 of the 767 evaluated videos show a humanoid, from NVIDIA GR1, ActionNet, RoboMIND and RoboCOIN. A humanoid carries eight labelled links, a palm, a thumb, an index finger and a forearm on each arm, against three on a single arm, so each video asks for a judgment on two hands and their fingers. Ours is above Gemini on all eight humanoid links. Over the links both methods score, the mean is 0.860 against 0.778, and over all annotated links it is 0.808 against 0.751. The largest margins are on the right index finger (0.804 against 0.662), the right thumb (0.849 against 0.710) and the left forearm (0.883 against 0.757).

\subsection{Temporal localization results}
\label{sec:supp_timing}

\begin{table}[t]
\centering
\caption{\textbf{Our reported frame lies within 5 and 10 frames of the human mark more often than Gemini's frame or the middle frame.} Share of the annotated positive videos per type on which a frame is reported (F) and whose frame lies within 5 or 10 frames of the mark. A video without a reported frame counts as a miss. The table covers every positive with a temporal mark, while the localization row of \cref{tab:human} is computed on a subset of them (\cref{sec:supp_timing_details}).}
\label{tab:supp_timing}
\footnotesize
\setlength{\tabcolsep}{2pt}
\begin{tabular*}{\linewidth}{@{\extracolsep{\fill}}l c ccc ccc cc@{}}
\toprule
 & & \multicolumn{3}{c}{Ours} & \multicolumn{3}{c}{Gemini} & \multicolumn{2}{c}{Middle} \\
\cmidrule(lr){3-5}\cmidrule(lr){6-8}\cmidrule(l){9-10}
Type & Videos & F & 5 & 10 & F & 5 & 10 & 5 & 10 \\
\midrule
\sContact & 298 & 0.95 & \best{0.58} & \best{0.73} & 0.78 & 0.38 & 0.48 & 0.11 & 0.23 \\
\sPersist & 173 & 1.00 & \best{0.49} & \best{0.61} & 0.88 & 0.36 & 0.58 & 0.09 & 0.17 \\
\sSolid & 152 & 0.91 & \best{0.58} & \best{0.71} & 0.74 & 0.28 & 0.39 & 0.13 & 0.20 \\
\sRobotDef & 480 & 1.00 & \best{0.69} & \best{0.74} & 0.99 & 0.31 & 0.41 & 0.22 & 0.33 \\
\sObjDef & 247 & 1.00 & \best{0.47} & \best{0.57} & 0.95 & 0.37 & 0.53 & 0.23 & 0.37 \\
\bottomrule
\end{tabular*}
\end{table}

The localization row of \cref{tab:human} is computed on a subset of the annotated positives of each type (\cref{sec:supp_timing_details}), the sampled videos with a reported frame: 75 for \mContact{}, 78 for \mPersist{}, 38 for \mSolid{}, 102 for \mRobotDef{} and 80 for \mObjDef{}. \Cref{tab:supp_timing} covers every positive with a temporal mark, 298, 173, 152, 480 and 247 videos, so its numbers differ slightly from those of the main paper. On all annotated positives the median frame error of our metrics is 3.5, 6, 3, 0 and 7 frames, and 4, 5, 3, 0.5 and 7 on the subsets. \Cref{tab:supp_timing} gives, for our metrics, for Gemini and for the middle frame, the share of the annotated videos on which a frame is reported and the share whose reported frame lies within 5 and within 10 frames of the human mark, where a video without a frame counts as a miss. On the subsets of the main paper, our frame lies within five frames of the human mark on 47\% to 60\% of the positives, against 20\% to 44\% for Gemini, and on all annotated positives on 47\% to 69\%, against 28\% to 38\%. Gemini's frame is the image it names among those it saw, mapped to the video frame that image shows. Marks are onset frames for \mContact{}, \mPersist{} and \mSolid{}, and intervals for the two deformation types, where the distance is taken to the nearest interval.

The onset was marked on 298 \mContact{} positives and on 173 \mPersist{} positives. In the \mContact{} and \mPersist{} rows of \cref{tab:supp_timing}, a video whose geometric score abstains takes the frame of the Gemini audit, which alone orders that video in the fused score. Over the 298 \mContact{} videos, our frame lies a median of 3.5 frames from the onset on 284 videos, against 6 for Gemini over 232 and 22.5 for the middle frame. The geometric frame alone, on the 172 videos where it exists, lies a median of 4 frames from the onset. On the other 112 videos our frame is the audit's frame, the same as Gemini's, so the difference between the two in the \mContact{} row comes from the 172 geometric frames. On the 120 videos where both the geometric frame and Gemini's frame exist, 63 of ours and 35 of Gemini's lie within 5 frames of the onset, and 81 and 50 within 10. The tracking frame of \mPersist{}, on the 171 of the 173 videos where it exists, lies a median of 6 frames from the onset. With a missing frame counted as a miss, our frame lies within five frames of the onset on 58\% and 49\% of the \mContact{} and \mPersist{} positives, against 38\% and 36\% for Gemini. The onset was marked on 152 \mSolid{} positives. On the 139 with a reported frame the median error of our metric is 3 frames, against 10 for Gemini over 113 and 23 for the middle frame.

\subsection{Violation rates per metric}
\label{sec:supp_generators}

In \cref{tab:generators} and \cref{tab:app_generators}, every metric flags at the threshold with the largest Youden index on the whole evaluation set, set separately for ours and for Gemini. Picking each generator's threshold only on the other generators and the real videos, with the fusions ranked the same way, changes our \mContact{}, \mPersist{}, \mRobotDef{} and \mObjDef{} rates by at most 6.2, 2.3, 0.9 and 5.9 points and leaves \mSolid{} unchanged, and the \mPersist{} ranking of the generators then matches the human one. On the 130 tasks that all five generators share, our \mContact{}, \mPersist{} and \mSolid{} rates order the generators as the human rates do, Kendall $\tau{=}1$ for each, against 1, 1 and 0.4 for Gemini. We also move the threshold of each of our metrics to the score percentile 2, 5 or 10 points below or above its Youden threshold, and to the value at which the share of flagged labelled videos equals the human share. At every one of these thresholds, the order of the five generators and the real videos by Overall rate matches the human order, Kendall $\tau{=}1$. Each \mSolid{}, \mRobotDef{} and \mObjDef{} cell of \cref{tab:generators} counts the videos that both score, given under Scored in \cref{tab:app_generators} for the three in turn, and each \mContact{} and \mPersist{} rate counts the videos its method scores, and the human share in parentheses every labelled video of the generator. \mRobotDef{} flags a video when its highest link score passes the threshold, and humans mark it when any link deforms. Agent mismatches are left out of \mRobotDef{} and kept in \mObjDef{}. The real videos are the 130 demonstrations of the evaluation set, counted as free of violations.

\begin{table*}[t]
\centering
\caption{\textbf{Violation rates per metric.} The share of each generator's videos that a metric flags, in percent, with the human share in parentheses, where lower is better. Overall is the mean over the five metrics and is the column of \cref{tab:generators}. Scored counts the videos that ours and Gemini both score for \sSolid{}, \sRobotDef{} and \sObjDef{} in turn.}
\label{tab:app_generators}
\footnotesize
\setlength{\aboverulesep}{0pt}\setlength{\belowrulesep}{0pt}\renewcommand{\arraystretch}{1.12}
\begin{tabular*}{\textwidth}{@{\extracolsep{\fill}}l c cccccc c@{}}
\toprule
Model & Videos & \sContact & \sPersist & \sSolid & \sRobotDef & \sObjDef & Overall & Scored \\
\midrule
\multicolumn{9}{@{}l}{\textit{Physical-AI world models}} \\
ABot-PhysWorld~\cite{abotphysworld} & 130 & 31.5 (36.9) & 16.9 (13.8) & 32.3 (14.6) & 50.8 (75.0) & 41.2 (32.8) & 34.5 (34.6) & 130 / 120 / 119 \\
Cosmos 2.5~\cite{cosmos25} & 132 & 43.9 (43.2) & 22.9 (15.9) & 28.8 (11.4) & 61.5 (91.7) & 31.9 (29.4) & 37.8 (38.3) & 132 / 96 / 119 \\
Cosmos3~Nano~\cite{cosmos3} & 239 & 23.0 (19.7) & 17.7 (12.6) & 29.7 (10.5) & 52.8 (73.8) & 31.9 (23.7) & 31.0 (28.1) & 239 / 214 / 207 \\
\midrule
\multicolumn{9}{@{}l}{\textit{General video generators}} \\
HunyuanVideo~1.5~\cite{hunyuanvideo15} & 130 & 51.5 (44.8) & 38.5 (30.4) & 38.4 (21.6) & 40.7 (55.6) & 48.3 (46.7) & 43.5 (39.8) & 125 / 108 / 120 \\
Wan2.2~5B~\cite{wan} & 130 & 73.8 (80.5) & 55.8 (56.2) & 66.4 (53.1) & 69.6 (82.4) & 53.4 (57.6) & 63.8 (66.0) & 128 / 102 / 118 \\
\midrule
\multicolumn{9}{@{}l}{\textit{Real videos}} \\
Real demonstrations & 130 & 17.7 (0.0) & 12.3 (0.0) & 10.0 (0.0) & 12.6 (0.0) & 8.4 (0.0) & 12.2 (0.0) & 130 / 127 / 119 \\
\bottomrule
\end{tabular*}
\end{table*}

\section{Prompts}
\label{sec:supp_prompts}

This section gives the main prompts whose answers enter the scores in \cref{tab:human}, copied from the code or the logged requests of the runs that produced the scores. Each boxed text is a verbatim prompt that a model received in our runs, quoted as data for reproducibility. Values in braces are filled at run time, and \texttt{\{k\}} is the number of frames. Unless stated otherwise, the frames follow the text in the same user message.

\subsection{Frames-only Gemini}

Gemini receives 16 frames sampled uniformly over the video, without the task prompt. For \mContact{}, \mPersist{} and \mSolid{}, it runs at temperature 0 and each frame is 768\,px on its long side with its frame number drawn on it.

\noindent\textbf{\mContact{}}\par\nopagebreak
\begin{promptbox}
These {k} images are frames from ONE AI-generated robot manipulation video, in time order.
Frame 1 is the first frame and Frame {k} the last; each image is labelled with its frame number.

Task: decide whether, at any point in the video, an object MOVES OR CHANGES WITHOUT A PLAUSIBLE PHYSICAL CONTACT:
for example it starts moving, lifts, slides or rotates before the robot's gripper or hand (or another object) touches
it; it is carried while the gripper is not actually holding it; or it follows the robot at a distance.

Do NOT count:
- objects that move only after the gripper, hand or another object visibly touches or pushes them;
- objects that fall or roll naturally after being released or pushed;
- camera motion, lighting changes, or the robot itself.

Answer with ONLY a JSON object:
{"probability": <integer 0-100: probability that at least one object moves or changes without a plausible contact>,
 "frame": <frame number where it first happens, or null>,
 "object": "<the object, or empty>",
 "reason": "<one short sentence>"}
\end{promptbox}

\noindent\textbf{\mPersist{}}\par\nopagebreak
\begin{promptbox}
These {k} images are frames from ONE AI-generated robot manipulation video, in time order.
Frame 1 is the first frame and Frame {k} the last; each image is labelled with its frame number.

Task: decide whether, at any point in the video, an object APPEARS OUT OF NOWHERE or gets DUPLICATED
(an extra copy of an object that was already there). Consider EVERY object in the scene, not only the one
the robot manipulates.

Do NOT count:
- objects already visible in Frame 1, even if they are moved, picked up, stacked, rotated or placed;
- objects that were hidden behind the robot or out of view and then become visible in a normal way;
- the robot itself, its shadow, reflections, or lighting changes.

Answer with ONLY a JSON object:
{"probability": <integer 0-100: probability that at least one object appears out of nowhere or is duplicated>,
 "frame": <frame number where it first happens, or null>,
 "object": "<the object that appears or is duplicated, or empty>",
 "reason": "<one short sentence>"}
\end{promptbox}

\noindent\textbf{\mSolid{}}\par\nopagebreak
\begin{promptbox}
These {k} images are frames from ONE AI-generated robot manipulation video, in time order,
each labelled with its frame number (Frame 1 is first).

Physical rule: two solid objects can never occupy the same space. A solid can become hidden only
(a) behind something that is in front of it (closer to the camera), or (b) by going through a real opening
(the open top of a cup / bin / box, an open drawer).

Question: anywhere in the video, does a solid object PASS THROUGH another solid (penetration)? For example:
gripper fingers going into a cup wall or through an object, an object entering a bin or drawer through its side
wall instead of its opening, an object sinking into the table, two objects merging into each other.
Do NOT count: normal occlusion, an object put into a container through its opening, ordinary grasp contact,
deformation of soft objects, blur, or objects appearing / disappearing for other reasons.

Answer with ONLY a JSON object:
{"probability": <integer 0-100: probability that at least one solid passes through another>,
 "frame": <frame number where it is clearest, or null>,
 "objects": "<which solid passes through which, or empty>",
 "reason": "<one or two short sentences citing what you see>"}
\end{promptbox}

For \mRobotDef{} and \mObjDef{}, each frame follows its own text label, such as ``Frame 7 (2.2 s):'', after the prompt. The \mRobotDef{} prompt lists one line per scored link with the frame that locates it, its box, an interior point and a short description of the link, and one JSON entry per link. For a single arm, its second paragraph opens with ``The robot is a single robot arm with a gripper. Joint housings between links belong to no link.'' and the joint motions read ``the arm moving, the wrist rotating, the gripper opening or closing''.

\noindent\textbf{\mRobotDef{}}\par\nopagebreak
\begin{promptbox}
The images below are 16 frames from ONE AI-generated robot video, in time order. Each image comes after its label "Frame k (t s)". Frame 1 is the first frame of the video, made from a real photo: there the robot has its true shape.

The robot has two arms with hands or grippers. Link names use the robot's own left and right. Rate these {n} links:
- {link}, located in Frame {f} ({t} s): box_2d [{box}], point_2d [{point}]. {description of the link}

Each link is located in one frame by box_2d [ymin, xmin, ymax, xmax] and point_2d [y, x], scaled 0-1000 to the image height and width; the point lies on the link. Follow each link through the other frames.

Physical rule: every link of a real robot is rigid. It moves and rotates only at its joints; its own shape and size never change.

Question: for each link, how much does its shape change during the video, compared with how it looks in Frame 1 and in the frame where it is located? Deformation means the link itself changes shape: it bends, stretches or shrinks, swells, melts or smears, twists, breaks or splits, merges with another part or with an object, gains or loses parts (extra or missing fingers, a part that fades away while it should be visible), or turns into a different kind of part (a hand becoming a claw).

Do NOT count: motion at the joints (the arm moving, the wrist rotating, the hand opening or closing, fingers bending at their knuckles), the link leaving the view or going behind something, perspective change as the link turns, motion blur, low resolution, a change of colour or lighting alone, and anything that happens to objects.

Degree, 0-100, at the worst moment of the video:
  0 = the same shape as in Frame 1 throughout
 25 = slight: a small or brief change, visible only when looking closely
 50 = clear: an obvious change of shape in part of the link or part of the video
 75 = strong: a large change of shape
100 = the link is destroyed or unrecognisable
Any value in between is allowed. Use null only when the link is never visible enough to judge.

Answer with JSON only, one entry for each of the {n} links ("what" = the change you see, or "none"; "frame" = the frame number of the worst moment, or null; "degree" = 0-100, or null):
{"links": {
  "{link}": {"what": "...", "frame": ..., "degree": ...},
}}
\end{promptbox}

\noindent\textbf{\mObjDef{}}\par\nopagebreak
\begin{promptbox}
The images below are 16 frames from ONE AI-generated robot video, in time order. Each image comes after its label "Frame k (t s)". Frame 1 is the first frame of the video, made from a real photo: there the object has its true shape and appearance.

Rate this object: {object}. It is located in Frame 1 by box_2d [ymin, xmin, ymax, xmax] = [{box}] and point_2d [y, x] = [{point}], scaled 0-1000 to the image height and width; the point lies on the object. Follow this object through the other frames.

Physical rule: an object keeps its shape and its appearance unless something acts on it. A rigid object (a cup, fruit, block, bottle, tool) never changes shape. A soft object (cloth, a sponge, a bag) bends or folds only where the robot or gravity pushes it. An articulated object (a drawer, a lid, scissors) moves only at its hinges or slides. No object changes its colour, texture, surface pattern, printing or material by itself.

Question: how much does the object change in shape or appearance during the video, in a way that its material and what acts on it cannot explain, compared with how it looks in Frame 1?
- Shape: it bends, dents, stretches or shrinks, swells, melts or smears, twists, breaks or splits without a cause, merges with the robot, another object or the table, gains or loses parts, a part fades away while it should be visible, or it turns into a different object.
- Appearance: its colour, texture, surface pattern, printing or label, or its material changes (for example, glass becoming opaque, metal becoming plastic), smears or flickers, or its surface blends into the robot, another object or the table.

Do NOT count: the object being moved, turned or tipped and so seen from another side (its outline then changes: a lying cup looks long, a cup seen from above looks round, a handle or the inside appears or disappears); a side or part of the object that comes into view for the first time and looks different; the object leaving the view or being hidden by the robot or another object; what is seen through a transparent object, refraction and reflections; shading, shadows, glare and changes of lighting; motion blur, low resolution and compression noise; the whole object vanishing or a new object appearing; and anything that happens to the robot or to other objects.

Degree, 0-100, at the worst moment of the video:
  0 = the same shape and appearance as in Frame 1 throughout
 25 = slight: a small or brief change, visible only when looking closely
 50 = clear: an obvious change in part of the object or part of the video
 75 = strong: a large change
100 = the object is destroyed, unrecognisable or turned into a different object
Any value in between is allowed. Use null only when the object is never visible enough to judge.

Answer with JSON only ("change" = "shape", "appearance", "both" or "none"; "what" = the change you see, or "none"; "frame" = the frame number of the worst moment, or null; "degree" = 0-100, or null):
{"object": {"change": "...", "what": "...", "frame": ..., "degree": ...}}
\end{promptbox}

\paragraph{Remaining baseline settings.}
The frames-only Gemini calls of \mContact{} and \mPersist{} run at temperature 0 with at most 2048 output tokens and no thinking level set. An unparseable answer, or a call that still fails after three attempts, leaves the video without a score rather than scoring it 0, and in our runs every one of these calls returned a score. A \mSolid{} call without a usable answer likewise gives no score, and a \mRobotDef{} link without a numeric answer is left out rather than scored 0. The frames-only calls of \mRobotDef{} and \mObjDef{} run at temperature 0 with medium reasoning effort and at most 16384 output tokens. Gemini 3.8 Flash and Claude Sonnet 5.5 receive the requests of Gemini 4 Argon for every metric, with the same prompt, frames and temperature. Their output limit is 16384 tokens for the two deformation metrics, 4096 for \mSolid{} and 2048 for \mContact{} and \mPersist{}, and the reasoning effort is medium for the two deformation metrics and not set for the other three.

\subsection{Our metrics}

\mContact{} names its target once per task and selector run with Qwen3-VL-8B, which sees three frames of each of the task's \texttt{\{N\}} videos with greedy decoding. The task hint is built from the file name.
\begin{promptbox}
You are selecting the target object for an automated robot-video
physical plausibility benchmark.

Matched task:
dataset: {dataset}
filename: {filename}
task hint: {task hint}

The images come from {N} different video generators performing the
SAME intended task.

Choose ONE concise physical target that should be tracked consistently
across all {n} generators.

Rules:
- Select the primary manipulated object or articulated part.
- Do NOT select robot, arm, gripper, camera, table, floor, or background.
- Prefer a concrete noun phrase usable as a text segmentation prompt.
- If the task operates an articulated object, the moving part is valid,
  e.g. "cabinet door".
- If several instances are manipulated as a category, use the category,
  e.g. "fruit" or "bowl".
- Do not judge whether the video is physically correct.
- Do not select a target based on which generator performs better.
- Output ONLY the target noun phrase.
- Maximum 5 words.
\end{promptbox}

Qwen3-VL-8B then rates the three crops at $t{-}2$, $t{-}1$ and $t$ of every candidate time with the prompt below. The score is $100\,(P_C + 0.5\,P_U)$, with $P_C$ and $P_U$ of \cref{eq:contact} taken from the likelihoods of the three answer letters.
\begin{promptbox}
You are examining three chronological HIGH-RESOLUTION ROI images
from a robot manipulation video.

Target object: {target}

Image 1 = t-2
Image 2 = t-1
Image 3 = candidate time t

IMPORTANT:
candidate time t is NOT known to be the target-response onset.
Do not assume that the target is responding at t.

Judge ONLY the visible spatial relationship between the robot
end-effector and the target object.

Use Images 1 and 2 to understand approach direction, local geometry,
gap closing, and depth ordering before Image 3.

Do NOT infer contact from target motion.
Do NOT infer contact from later consequences.
Mere 2D overlap is NOT sufficient evidence of physical contact.

Choose exactly ONE answer:

A

CLEARLY SEPARATED / NO CONTACT EVENT SUPPORTED BY t:

At Image 3, a visible gap remains between the closest robot and
target-object surfaces, or the images do not show the robot reaching
the target surface.

B

DEPTH AMBIGUOUS:

At Image 3, the robot and target appear extremely close or overlap in
2D projection, but the images do not reliably establish whether the
surfaces truly meet or whether a small 3D gap remains.

C

SURFACE-CONTACT EVENT SUPPORTED BY t:

At Image 3, a specific robot end-effector surface visibly meets the
target-object surface at a plausible contact location, with no visible
separating gap and sufficient visual evidence to support actual surface
contact. Earlier images may show the gap closing or the end-effector
approaching that same contact location.

Return exactly one character:

A, B, or C.
\end{promptbox}

Our \mSolid{} judge, gemini-4-argon at temperature 0, sends the frames-only \mSolid{} prompt above with the paragraph below inserted after its first two lines. \texttt{\{intro\}} reads ``The groups below follow'', and \texttt{\{lines\}} holds one line per frame group, for example ``Group B, the robot's hand or gripper and the object being moved: Frames 6, 7, 8, 12, 13, 14.'' A video without any group gets the prompt unchanged.
\begin{promptbox}
The frames are not evenly spaced. {intro} one pair of things more closely, around the moments they come
together, slightly enlarged around them:
{lines}
The other frames are spread over the rest of the video and show the whole view. The groups only say where to
look closely; they do not say whether anything is wrong.
\end{promptbox}

\paragraph{\mSolid{} perception prompts.} Gemini 3.8 Flash finds the two objects and the first frame that shows both with the prompt below, on six frames spread over the video with their longest side at most 1024\,px.
\begin{promptbox}
You look at frames from one video in which a robot (or a person) handles objects. The frames are in time order, spread evenly over the whole video from its first to its last frame, each labelled with its frame number. When the task has a text description, you get it too.

1. OBJECTS. Find the two objects of the main interaction:
- the PRIMARY object: the one object the robot (or person) moves during the video, the physical, non-robot thing its gripper or hand picks up, carries, places, pushes, pulls, slides or presses. When something is put into, onto or against another object, the primary object is the thing being put.
- the SECONDARY object: the one other object that the primary object is put into, onto, against or through, or that the robot works on together with it. Take the whole physical unit (a drawer with its cabinet, a bin with its rim). There is none when the primary object interacts with no other object; never choose an object only because it is in view.
- Never the robot or any part of it, a person, the table or floor that everything stands on, a wall or the background.

2. DO THE OBJECTS STAY THE SAME? Compare the objects in the scene from frame to frame, apart from the robot. Do the same objects stay in the scene from the first frame to the last, or does the scene change: an object comes into view later (lowered from above, brought in from outside the picture, uncovered, or appearing in the robot's gripper or hand), an object disappears, or an object changes its size, shape or colour, or turns into another one? Videos can be generated, so such changes happen.

3. FIRST FRAME OF EACH OBJECT. For each of the two objects, the first given frame that shows that very object: the same physical object, with the size, shape and colour it has when the objects interact. Follow it back frame by frame. An earlier object that only looks alike does not count: another object of the same kind elsewhere in the scene (for example a second cup or box that stays where it is), or an object of the same colour but of another size or shape that later turns into it or is replaced by it.

4. FIRST FRAME OF THE INTERACTION. The first given frame from which both objects of the interaction (the primary object alone when there is no secondary object) are in view as the very objects that interact. When the objects stay the same and both are in view from the start, this is frame 1. When the scene changes, it is the first frame after the change that brings in or changes an object of the interaction.

5. DESCRIBE each of the two objects as it looks from that frame on: a short name (2 to 5 words) and one sentence on its colour, shape, size and material and where it is, that tells it apart from similar objects.

Describe only what the frames show. Do not judge whether anything in the video is physically plausible.

Answer with one JSON object and nothing else:
{"consistent": true or false,
 "changes": "<what comes into view, disappears or changes, and in which frame; or none>",
 "reason": "<one or two sentences: what the robot does and with which objects>",
 "primary": {"name": "<short name>", "description": "<one sentence>", "first_frame": <number of the given frame>},
 "secondary": {"present": true, "name": "<short name>", "description": "<one sentence>", "first_frame": <number of the given frame>},
 "first_frame": <number of the given frame: the first frame of the interaction>}
When there is no secondary object: "secondary": {"present": false, "reason": "<one sentence>"}.
\end{promptbox}
A second call on five consecutive frames, with that frame as reference, finds the first of them that shows both objects.
\begin{promptbox}
You look at consecutive frames from one video, in time order, each labelled with its frame number. You get the two objects of the video's main interaction, named and described from frames across the whole video, and a reference frame from later in the video in which both are visible. Find the first of the given frames in which both of these same objects are visible (the primary object alone when there is no secondary object). An object that only looks alike (another box of the same kind, a block of the same colour but of another size or in another place) does not count. Answer 0 when none of the given frames shows both.

Describe only what the frames show. Do not judge whether anything in the video is physically plausible.

Answer with one JSON object and nothing else:
{"reason": "<one sentence>", "first_frame": <the number of the given frame, from 1 to the number of frames, or 0>}
\end{promptbox}
The second object is boxed on the first frame that shows both objects and on the frame halfway from it to the end of the video with the prompt below. The manipulated object is outlined in green on both frames.
\begin{promptbox}
You find the secondary object in a generated robot video. The video shows a robot doing the task of its generation prompt. The manipulated object (the thing the robot picks, holds or moves) has already been found; you get its name, and it is outlined in green on both frames. You find the one other physical object that the robot or the manipulated object is likely to touch or interact with during the task. You get the generation prompt, the manipulated object's name, the video's first frame and its middle frame.

1. WHICH OBJECT. Read the generation prompt for the task, then look at both frames.
- The secondary object is what the manipulated object is put into, onto or against (a bin, basket, box, drawer, bowl, plate, tray, shelf, container), or a thing the robot pushes, pulls, opens or closes as part of the task (a drawer, door, lid).
- When the task names no such object, choose the object the frames show the robot or the manipulated object most likely interacting with. When there is none (for example, the robot only lifts the object in the air), answer "present": false. Never choose an object only because it is in view.
- Take the whole physical unit: a drawer together with the cabinet or frame it slides in; a bin with its rim; a box with its lid.
- Never the robot or any part of it, a person, the table or floor that everything stands on, a wall or the background.
- Never the manipulated object outlined in green, any part of it, or the same object again under another name.

2. WHERE. On each of the two frames, say whether the object is visible. When it is, give:
- "box": a tight box around the whole visible object, every visible part of it (for a bag, box or container: all its walls, panels, flaps and handles, not only the face nearest the camera), [ymin, xmin, ymax, xmax];
- "positive": 1 to 3 points on the object itself, spread over its different visible parts (separate panels or sides), not on its edge;
- "negative": 0 to 3 points next to the object where a mask could spill: the robot, the manipulated object, the table beside it.
All coordinates are integers from 0 to 1000, relative to that frame (0 = top or left edge, 1000 = bottom or right edge). A point is [y, x].

3. ROLE. One of "receptacle" (things go into or onto it), "articulated" (the task moves a part of it), "support" (things rest against it), "other".

Answer with one JSON object and nothing else:
{"present": true, "name": "<short name>", "role": "<role>", "reason": "<one sentence: why the robot or the object interacts with it>",
 "first_frame": {"visible": true, "box": [ymin, xmin, ymax, xmax], "positive": [[y, x], ...], "negative": [[y, x], ...]},
 "middle_frame": {"visible": true, "box": [ymin, xmin, ymax, xmax], "positive": [[y, x], ...], "negative": [[y, x], ...]}}
For a frame where the object is not visible: {"visible": false, "reason": "<one sentence>"}.
When there is no secondary object: {"present": false, "reason": "<one sentence>"}.
\end{promptbox}
Fitted surfaces that are openings are dropped with the prompt below, on one frame at 896\,px shown plain and with the lettered surfaces drawn on it.
\begin{promptbox}
You look at one frame from a video, shown twice: plain, and with coloured, lettered regions drawn on it. The regions are surfaces that a program fitted to one object, the SECONDARY object (outlined in blue): flat sides of it, its top, or the top of a round object. You get the object's name and what each letter is.

For each lettered region, say whether it lies on a solid face of the object or on an opening:
- "solid": material that nothing can pass through: a wall, a side, a panel, a closed top, a lid, a shelf board, the top of a block or of a cabinet.
- "opening": open space through which things can be put into or taken out of the object: the open top of a bin, box, basket, bowl, cup or drawer, the open front of a shelf or cubby.
A region is drawn as a flat patch even where the object has open space: judge what the object has there, not the drawing. Use the plain frame to see the object itself.

Describe only what the frame shows. Do not judge whether anything in the video is physically plausible.

Answer with one JSON object and nothing else:
{"reason": "<one sentence: what the object is and where its openings are, if any>",
 "surfaces": {"A": "solid" or "opening", "B": "solid" or "opening", ...}}
\end{promptbox}
Whether the object is ever hidden inside the second one is decided on eight frames with both objects outlined.
\begin{promptbox}
You look at frames from one video in which a robot (or a person) handles objects. The frames are in time order, from the first to the last frame of the video. In each frame two objects are outlined: the PRIMARY object, the object being handled, in GREEN, and the SECONDARY object, the other object it interacts with, in BLUE. An outline covers only the part of an object that is visible in that frame; a part hidden behind something has no outline.

Answer one question about the whole video: does the primary object, at some point, go largely inside the secondary object, the way a thing goes into a container (a bin, box, basket, drawer, cabinet, cup or bowl), and does the secondary object's own sides, walls or front then hide it, entirely or in part, from the camera?

1. "inside": true when, at some point, most or all of the primary object is inside the secondary object (within its walls or sides, or below its top), the way a thing put into a container is inside it. False when the primary object only goes onto, next to, against, over or around the secondary object, or when only a small part of it goes into the secondary object while most of it stays outside.
2. "hidden": true when, while the primary object is inside, the secondary object's own sides, walls or front cover it, entirely or in part, so that less of it is visible from the camera. False when the primary object stays in plain view while inside (for example seen from above through an open top, or through clear walls), and false whenever "inside" is false.

Describe only what the frames show. Do not judge whether anything in the video is physically plausible.

Answer with one JSON object and nothing else:
{"reason": "<one or two sentences: where the primary object goes relative to the secondary object, and what covers it>", "inside": true or false, "hidden": true or false}
\end{promptbox}
A call of the \mSolid{} judge without a usable answer gives no score.

\subsection{Agent prompts and tools}
\label{sec:supp_agent_prompts}

This section gives the prompts of the measurement agent that masks the robot links for \mRobotDef{} (\cref{sec:supp_deform}) and the prompts of the audit agent (\cref{sec:supp_agent}). \Cref{tab:supp_calls} lists each call with its model, its settings, its input, its closed set of answers and the step that runs next, and the Gemini calls run at temperature 0. In the boxes, a line in square brackets marks the position of an image in the message, and an italic line stands for blocks printed in an earlier box.

\begin{table*}[t]
\centering
\caption{\textbf{Calls of the measurement agent and the audit agent.} For each call, the model, the reasoning effort, the output limit in tokens, the input, the answer with its closed set of choices and the step that runs next. Opus is Claude Opus 5.5, Flash is Gemini 3.8 Flash and Argon is Gemini 4 Argon. A dash means that no limit is set.}
\label{tab:supp_calls}
\footnotesize
\setlength{\tabcolsep}{3pt}
\begin{tabularx}{\textwidth}{@{}>{\raggedright\arraybackslash}p{0.15\textwidth} l l r >{\raggedright\arraybackslash}X >{\raggedright\arraybackslash}X >{\raggedright\arraybackslash}p{0.165\textwidth}@{}}
\toprule
Call & Model & Effort & Limit & Input & Answer & Next step \\
\midrule
\multicolumn{7}{@{}l}{\textit{Measurement agent}} \\
Agent check & Flash & low & 16384 & Generation prompt; first, middle and last frame & \texttt{same\_robot} or \texttt{agent\_mismatch} & \mRobotDef{} abstains on a mismatch (\cref{sec:supp_coverage}) \\
Robot type & Flash & low & 8192 & Generation prompt; first, middle and last frame & \texttt{serial\_manipulator} or \texttt{humanoid} & Masking pipeline of that structure \\
Planner, humanoid & Opus & high & 32768 & Example images with human-drawn boxes; anchor frame & Per link: \texttt{full}, \texttt{partial}, \texttt{occluded}, \texttt{out\_of\_frame} or \texttt{uncertain}; a box for \texttt{full} and \texttt{partial} & SAM3 box prompt \\
Planner, arm, clusters & Opus & high & 32768 & Example images with human-drawn boxes; anchor frame, plain and with the numbered clusters; cluster trajectory table & Per link: \texttt{assign}, \texttt{merge}, \texttt{split}, \texttt{emit} or \texttt{skip} & Mask and box from the clusters; box call for \texttt{emit} and \texttt{split} \\
Planner, arm, boxes & Opus & low & 8192 & One example image; anchor frame & A box per emitted link; a joint point per split & SAM3 box prompt \\
Points, humanoid & Flash & low & 8192 & Example crop with its points; anchor frame with the box; crop around the box & 1--3 positive and 0--4 negative points & SAM3 with the box and the points \\
Points, arm & Flash & low & 8192 & Example crop with its points; anchor frame with the box; crop around the box & 3 positive points, 2 for the gripper palm & SAM3 with the box and the points \\
Critic, humanoid & Opus & low & 16384 & Example crop; anchor frame with the box and the mask; crop, plain and with the mask and the numbered points & \texttt{accept}, or \texttt{refine} with the kept and the new points & One more SAM3 pass on a refine \\
Critic, arm & Opus & low & 16384 & Example crop; anchor frame with the box and the mask; crop, plain and with the mask and the points & \texttt{accept}, or \texttt{refine} with 1--3 positive and 0--5 negative points & One more SAM3 pass on a refine \\
\midrule
\multicolumn{7}{@{}l}{\textit{Audit agent}} \\
EVIDENCE\_\allowbreak REPAIR & Argon & high & \na & Five frames; task prompt; state of the two masks & \texttt{box\_object}, \texttt{mask\_robot}, \texttt{both} or \texttt{abstain}; the object and its frame & Object box call; SAM3 on the text ``robot'' \\
\quad Object box & Flash & high & \na & The named frame; object name; task prompt & A box, or null & SAM3 tracks the box in both directions; \mContact{} reruns \\
\quad Moved-object check & Flash & high & \na & Four frames; box of the mask on each & True or false for \texttt{moved\_object\_is\_masked}, \texttt{nothing\_moves} and \texttt{mask\_jumps} & Motion-aware box \\
\quad Motion-aware box & Flash & high & \na & The boxed frame and two later frames; object name; task prompt & A frame and a box, or null & \na \\
EDGE\_CASE\_\allowbreak INSPECTION & Argon & high & 32768 & Mask geometry as text; up to 14 frames, plain and with both masks & Causes A to E; actions \texttt{re\_reference}, \texttt{remask\_robot}, \texttt{robot\_absent}, \texttt{object\_owns\_overlap}, \texttt{remask\_object} or \texttt{no\_repair} & Repair by code or by the two box calls; \mObjDef{} reruns \\
\quad Object box & Flash & low & 4096 & One frame; object name & Visible or not; a box & SAM3 tracks the box \\
\quad Robot-part boxes & Flash & low & 4096 & One frame & A box for each visible robot part & SAM3 tracks each box \\
\bottomrule
\end{tabularx}
\end{table*}

\paragraph{Robot type.} Gemini 3.8 Flash receives the system prompt below and one user message with the generation prompt and the first, middle and last frame. It answers \texttt{serial\_manipulator} or \texttt{humanoid}, and the answer selects the masking pipeline of that structure.
\begin{promptbox}
ROLE. You route one AI-generated robot video to the analysis pipeline for its kind of robot. You see the text
prompt the video was generated from and three frames of it: the first, a middle and the last frame.

ROBOT KINEMATIC KNOWLEDGE (identical for every video)

SERIAL MANIPULATOR (a single robot arm)
- One arm: a single chain of rigid links joined by rotary joints, from a base outward:
  base -> shoulder -> upper arm -> elbow -> forearm -> wrist -> gripper or tool.
- The base stands on a table, the floor, a pedestal, a rail or a cart. There is no torso, no chest and no head: the
  first link rises straight from the base.
- The end of the arm carries one gripper (two fingers, parallel jaws, a suction cup or a tool), not a human-like hand.
- Seen close up, often only part of the arm is in view: the forearm, the wrist and the gripper entering from one edge
  of the image. It is still a serial manipulator when nothing in view is a torso with a second arm and the arm ends in
  a gripper, not a multi-fingered hand.
- Other robot arms that only stand in the scene do not change the type: classify the robot that does the task.

HUMANOID OR HUMANOID-LIKE ROBOT
- A torso (chest) with TWO arms, one attached at each side of the upper torso, usually with a head on top. The two
  arms are mirror images of each other: the same links in the same order.
- Below the torso there may be a pelvis and two legs, a fixed stand, or a wheeled base. A torso with two arms on a
  wheeled base (a humanoid-like mobile manipulator) counts as humanoid.
- Each arm: shoulder -> upper arm -> elbow -> forearm -> wrist -> hand. The hands are often dexterous hands with
  fingers, sometimes grippers.
- Often only one arm is in view, with the torso and the other arm outside the image. An arm that ends in a dexterous,
  multi-fingered (human-like) hand is a humanoid arm even then.
- From the robot's own head (first-person camera): two arms enter the image from the bottom edge, one on each side,
  with hands or grippers in front of the camera; the torso and the head are not in view.

HOW TO DECIDE
1. Find the robot that does the task (the text prompt says what it does).
2. Count its arms that attach to one shared body. Two arms on one torso, or two mirror-image arms entering from the
   bottom edge of a first-person view: humanoid. One arm on its own base: serial manipulator.
3. Look for a torso, a chest, a head, legs or human-like hands: any of them means humanoid.

ANSWER: which kind of robot does the task in this video? Exactly one of:
- "serial_manipulator": a single robot arm on a base;
- "humanoid": a humanoid or humanoid-like robot (a torso with two arms), including first-person views from its head.

OUTPUT: JSON only, with exactly these keys and, where listed, exactly these values (do not rename them):
{
  "robot_type": "serial_manipulator" | "humanoid",
  "evidence": ["short observations that decided it, e.g. one arm rising from a table-mounted base"],
  "reason": "one or two sentences"
}
\end{promptbox}

A second call checks that the robot of the first frame is still the one acting, and answers \texttt{same\_robot} or \texttt{agent\_mismatch}.
\begin{promptbox}
ROLE. You check one AI-generated robot video before it is analysed. An image-to-video model made the video from
its FIRST FRAME (a real photo of a robot at work) and a text prompt that describes what the robot should do. You see
the text prompt, the first frame, a frame from the middle of the video and its last frame.

THE MAIN AGENT is the robot that is shown acting in the first frame: the robot the text prompt is about and the video
focuses on. Describe it before you look at the later frames: its kind (one arm on a base, a humanoid with a torso and
two arms, ...), its colour and its gripper or hands, where it is mounted and from which side the camera sees it.

QUESTION. In the middle frame and in the last frame: is the main agent still the first-frame robot, and has no new,
distinct robot taken over its role?

A NEW AGENT is a distinct robot that is NOT in the first frame and emerges during the video from a different viewpoint
than the main agent: it reaches in from another side or edge of the image, with its own base or mounting and its own
angle, often looking larger and closer to the camera, and it takes over the main agent's role in the task (it does the
manipulation the text prompt gives the main agent). Meanwhile the main agent is still present somewhere in the image
(idle, at an edge, partly visible) or is hidden behind the new robot. It is a new agent even when it is the same kind
and colour as the main agent. Only robots can be new agents.

NOT a new agent (these all count as the SAME robot):
- the first-frame robot deformed: bent, stretched, melted, twisted, broken links, wrong joints, extra or missing
  parts, a changed gripper, changed colour, texture or lighting. A strange-looking version of the first-frame robot is
  still the first-frame robot; we are not checking its quality here;
- the main agent changing its appearance in place: its hand turning into a claw or a gripper, an arm turning into a
  different-looking arm, the whole robot turning into another robot from the same body, mount and position. This is
  deformation, however different the result looks;
- people, human arms or hands, whether they enter the scene or take part in the action: only robots are checked here;
- the first-frame robot partly or completely leaving the image, or hidden behind an object;
- robots, arms or people that are already visible in the first frame, moving or not;
- supporting figures: robots in the background that do not take part in the action;
- objects, tools, shadows and reflections.

VERDICT:
- "agent_mismatch" only when a new agent as defined above appears in the middle or last frame: a distinct robot from
  another viewpoint takes over the main agent's role while the main agent is still in the image or hidden behind it.
- "same_robot" otherwise, including when the main agent turns into a different-looking robot where it stands, when a
  person joins the action, and when the main agent has simply left the image and no new robot is there.

OUTPUT: JSON only, with exactly these keys and, where listed, exactly these values (do not rename them):
{
  "first_frame_robot": "one-sentence description of the main agent in the first frame",
  "later_frames": {
    "middle": {"first_robot": "visible" | "partly_visible" | "not_visible",
               "new_robots_or_humans": [{"kind": "robot" | "human", "description": "..."}]},
    "last": {"first_robot": "visible" | "partly_visible" | "not_visible",
             "new_robots_or_humans": []}
  },
  "verdict": "same_robot" | "agent_mismatch",
  "reason": "one or two sentences"
}
\end{promptbox}

Both calls receive the user message below.
\begin{promptbox}
TEXT PROMPT the video was generated from:
{generation prompt}
the FIRST frame (the input image): frame {index} of {count} ({W}x{H}):
[image: the first frame, PNG]
a MIDDLE frame: frame {index} of {count} ({W}x{H}):
[image: the middle frame, PNG]
the LAST frame: frame {index} of {count} ({W}x{H}):
[image: the last frame, PNG]
Return the JSON only.
\end{promptbox}

\paragraph{Link boxes.} On the anchor frame of a humanoid, Claude Opus 5.5 returns a status and a box for each arm link, as \texttt{box\_2d = [ymin, xmin, ymax, xmax]} on a scale of 0 to 1000 of the frame. Each box is the SAM3 box prompt of its link. The system prompt below is that of the front view, and the \texttt{VIEWPOINT} line and the names of the two arms follow the viewing direction of the video.
\begin{promptbox}
ROLE. You are the planner of an automatic link-masking pipeline for humanoid robots. On the anchor frame of a
video, you decide which of the robot's six arm links (upper arm, forearm and palm of each arm) are visible, and you
draw a box around each visible one. Later steps place click points inside each box and segment the link with SAM; they
never move your box, so a box on the wrong part gives a wrong mask.

HUMANOID KINEMATIC KNOWLEDGE (identical for every video)

1. Body. A humanoid robot has a torso with a head on top and two arms, one attached at each side of the upper
   torso. Below the torso there may be a pelvis and two legs, a fixed stand, or a wheeled base. The two arms are
   mirror images of each other: the same parts in the same order. When one arm is easier to read, use it to
   understand the other.

2. Arm chain. Each arm is a chain of rigid parts joined by rotary joints, from the body outward:
   shoulder -> upper arm -> elbow -> forearm -> wrist -> hand (palm and fingers).
   Identify an arm by following it from where it attaches to the body, never by where its hand is. Arms can cross
   or reach across the body; each arm keeps the identity of its own shoulder.
   A link is part of this arm only if it connects through the chain to this arm's shoulder; an arm or machine
   that crosses in front of the robot is not one of its links.

3. Long links and intermediate links. The rigid parts of an arm are of two kinds.
   - Long links: the upper arm and the forearm. Each is one rigid shell, clearly longer than it is wide: a
     cylinder, a rounded box or a tapered tube, often with a smooth cover.
   - Intermediate links: the short parts between and around the long links. The shoulder usually carries a stack
     of two or three short motor housings between the torso and the upper arm. The elbow is a short joint housing
     between the upper arm and the forearm. The wrist has one to three short segments between the forearm and the
     hand. An intermediate link is about as long as it is wide: a round disc, a cap or a short drum, often of a
     different colour or material than the long shells, with a seam or rotation axis at its ends.
   - How to see a joint. A rotary joint shows its axis: a round end cap, disc or ring facing the camera (often with
     a centre cover, screws or a seam around it), a rounded housing that bulges wider than the shells next to it,
     or the place where the arm bends. Such a housing is an intermediate link even when it has the same colour and
     surface as the long shell and seems to flow into it. A long shell ends where its outline narrows into, or
     meets, such a housing.
   - A long shell can carry a twist seam partway along its length. Then the longer rigid piece is the link and the
     short piece on the other side of the seam is an intermediate link.
   Intermediate links are never one of the target links: they are not boxed, never clicked as positive, and never
   folded into a neighbouring long link to make it look complete.

4. Target links.
   - upper_arm: the long shell between the shoulder stack and the elbow housing.
   - forearm: the long shell between the elbow housing and the wrist segments. It usually points toward what the
     hand is working on.
   - palm: the rigid body of the hand between the wrist and the finger roots, back and palm side. It ends at the
     knuckle line where the fingers start; the first segments of the fingers belong to the fingers. A hand may have
     five fingers, three fingers or two jaws; for a jaw gripper the palm is the gripper body without the jaws.
   - Fingers are not part of the palm.
   Cables, covers, labels and fittings fixed to a shell belong to that link.

5. Never a target: the head, the torso, the pelvis, legs, a stand or base, a harness or gantry above the robot,
   loose hanging cables, any object the robot holds or touches, people (a human arm or hand is never a robot link),
   and any other robot or machine.

6. Visibility. Usually only part of the robot is in view. A link may be cut by the image edge, hidden behind the
   torso, the other arm, a table or a held object, or seen end-on. A link pointing toward or away from the camera
   looks short; it is still a long link. Decide what a part is by where it sits in the chain (which joints it lies
   between), not by its apparent length. Only the visible part of a link is masked. A link you cannot identify with
   confidence is left out: a missing link is acceptable, a wrong one is not.

THIRD-PERSON VIEW. The camera looks at the robot from outside. The torso is in view, usually with the head; the
shoulders are at the top corners of the torso and the arms hang down from them or reach forward.
- An arm hanging beside the torso: upper arm and forearm lie along the side of the torso, one above the other, with
  the elbow at about waist height. The edge of the arm shell is the boundary with the torso.
- An arm reaching toward the camera: the forearm and hand cover part of their own upper arm, which may show only a
  short piece below the shoulder.
- An arm reaching forward or sideways over a table: the elbow bend is the boundary between upper arm and forearm.
- In an oblique, side or back view the far arm is partly or mostly hidden by the torso; often only its hand or
  forearm shows.
- Targets: palm, forearm and upper arm of each arm, six at most.

The torso, head, pelvis and legs are in view and are not targets.

VIEWPOINT (front): The robot faces the camera. Its right arm is on image-left and its left arm on image-right, at about the same depth. Name the arms `image_left` and `image_right`.

STATUSES (one per link):
- "full": the whole link is visible.
- "partial": only part of the link is visible (cut by the image edge or partly hidden); the box holds the visible part.
- "occluded": the link is hidden (by the torso, the other arm, a held object, a table, a person).
- "out_of_frame": the link is outside the image.
- "uncertain": you cannot tell which visible part is this link, or you are not sure.
confidence (0-1): how sure you are of the status and, for a boxed link, that the box holds this link and nothing else.
A link has a box_2d exactly when its status is "full" or "partial" and its confidence is at least 0.5. A partial link
is boxed only when its visible part can be named as that link; otherwise it is "uncertain" and gets no box.

BOX CONVENTION
- An upper-arm or forearm box encloses the visible part of that long shell and stops at the joint housings at its
  ends. It is not a box over the whole half of the arm.
- A palm box encloses the hand body. Fingers may fall inside the box; downstream points keep them out of the mask.
  The wrist segments and any held object stay outside the box.
- Boxes of one arm may overlap at a joint.
- A partly visible link gets a box around its visible part only.
- box_2d = [ymin, xmin, ymax, xmax], integers normalized to 0-1000 of the anchor frame.

EXAMPLES. Before the anchor frame you see human-annotated photos of other humanoid robots seen from the same
viewpoint. Each comes with its green boxes labelled, its exact boxes and a description of every box. They teach the
convention: what each box contains and what it leaves out. In them, "gripper" means the palm. A link without a box in
an example was not annotated there; that does not mean it is hidden.

FRAME TOOL. You may set "frame_request" to 2 or 3 to see runner-up frame 2 or 3 of the same video, each once. They are
context only: they show what the anchor hides (an arm behind an object, which arm is which). Every box is still drawn
on the anchor frame. When you request a frame, still give your best complete answer now; you will then see the frame
and answer again. Set "frame_request" to null when you are done.

Precision over recall: a missing link is acceptable, a wrong one is not.

OUTPUT: JSON only:
{
  "arms": [
    {"arm": "image_left",
     "links": [
       {"link": "upper_arm", "status": "partial", "confidence": 0.8,
        "visible_part": "white shell below the shoulder housings down to the dark elbow housing; lower half behind the table",
        "box_2d": [ymin, xmin, ymax, xmax]},
       {"link": "forearm", "status": "full", "confidence": 0.9, "visible_part": "...", "box_2d": [ymin, xmin, ymax, xmax]},
       {"link": "palm", "status": "occluded", "confidence": 0.7, "reason": "held object covers the hand"}
     ]},
    {"arm": "image_right", "links": ["... always the three links upper_arm, forearm, palm"]}
  ],
  "frame_request": null
}
- Both arms are always listed ("image_left" and "image_right"), each with exactly the three links upper_arm, forearm and palm.
  An arm out of view has three "out_of_frame" entries.
- "visible_part" names what the box starts and ends on; it is passed to the later steps. Every boxed link has one.
- A link without a box has a short "reason".
\end{promptbox}

In first-person views, the blocks below take the place of the \texttt{THIRD-PERSON VIEW} block and the line after it.
\begin{promptbox}
FIRST-PERSON VIEW. The camera is the robot's own head, looking down and forward at its hands. The robot's head,
face and the front of its torso are not in view; parts of its chest, shoulders or legs may appear along the bottom
edge and are not targets.
- The arms enter from the bottom edge, the left arm from the lower left and the right arm from the lower right,
  and reach into the scene.
- Along each arm from the bottom edge upward: upper arm (nearest the camera, large, often cut by the bottom or side
  edge and often blurred), elbow, forearm (often pointing away from the camera, so it looks shorter than it is),
  wrist, and the hand at the far end.
- Often only the forearms and hands are in view; the upper arms may be absent.
- The hands are seen from above or from behind: the back of the hand, the palm or the side of the hand may face the
  camera, and the fingers may point away, down or toward the other hand.
- Targets: palm, forearm and upper arm of each arm (six at most), plus the thumb and index finger of each hand,
  which a separate step handles. The thumb is the shorter, thicker finger set on the side of the palm near the
  wrist, able to close against the others; the index finger is the long finger next to the thumb.

- The arm labels are fixed: image_left is the robot's left arm and image_right its right arm.
- Upper arms are often cut by the bottom or side edge; box the part in view.
- Fingers are not boxed here. The palm box covers the hand body, and a thumb or index finger crossing it is left to
  the finger step.
\end{promptbox}

The user message shows the human-annotated examples of the same viewing direction and then the anchor frame.
\begin{promptbox}
EXAMPLES from the viewpoint bucket {bucket}:
[IMAGE: example {k}: the exemplar photo with its human-drawn boxes in green, each tagged "{arm} {link}"]
Example {k}: the {W}x{H} photo above, a different robot and scene. Its boxes, in the box format of your answer (arm labels as in this viewpoint):
{exemplar boxes as JSON}
What each box covers:
- {arm} {link}: {description}
ANCHOR FRAME {anchor} of the video ({W}x{H}), clean. Draw every box on this frame:
[IMAGE: the anchor frame, clean]
Return the JSON only.
\end{promptbox}

For a single arm, the planner first decides how each of the three links is formed from the numbered rigid-motion clusters of MoSE3 outlined on the anchor frame, with one of five verdicts per link: \texttt{assign}, \texttt{merge}, \texttt{split}, \texttt{emit} or \texttt{skip}. The code takes the mask and the box of an assigned, merged or split link from its clusters. The system prompt below is that of the front view, and item 7 follows the viewing direction of the video.
\begin{promptbox}
You are the planner of an automatic robot-link masking pipeline for a single-arm robot (serial manipulator).
For the three links upper_arm, forearm and gripper_palm, decide which ones are visible in the anchor frame and how each
visible link is formed from the numbered rigidity clusters drawn on it. Judge from what the image shows first; the clusters
are evidence, not answers. Code turns your decisions into masks and boxes. In this call you write NO coordinates and you
never see, predict or adjust a box.

KINEMATIC KNOWLEDGE (identical for every video)
1. The three links. gripper_palm = the rigid body between the wrist flange and the fingers; forearm = the long link between the elbow and the wrist, toward the gripper; upper_arm = the long link between the shoulder and the elbow, toward the base. Chain order: base -> upper_arm -> forearm -> gripper_palm.
   Where they usually are: the upper arm is connected to the base and often runs from the base up toward the ceiling.
   The forearm often sits a bit higher than the gripper, and its direction often aims toward the object on the table or
   ground that the robot is approaching.
2. Intermediate links. Many robots carry extra links between these three: wrist segments between the forearm and the
   gripper palm, a short elbow segment that joins the upper arm and the forearm between them, and tool-changer stacks
   above the palm. Intermediate links are not one of the three links: they are never boxed and
   never folded into a neighbouring link.
3. Joint housings (the round or cylindrical motor housings at each joint) belong to neither neighbouring link.
4. The gripper palm excludes the fingers and any grasped object; fingers are removed downstream. A cluster that covers the
   gripper body together with its fingers is the gripper_palm cluster; "fingers" is only for a cluster that holds nothing
   but fingers. Cables and labels fixed to a tube belong to that link.
5. Not every link is in view. The full chain is gripper_palm, forearm, upper_arm, but the camera often sees only part of
   it: only the gripper, or two of the three (typically the gripper and the forearm) when the upper arm is out of the
   image or hidden. Then only the links you actually see get a verdict that boxes them; the others are skipped.
6. Distal links accumulate motion: the base does not move, the gripper moves most.
7. Viewpoint (front_view): The base stands at the bottom. Proximal links rise from the base; distal links come toward the camera and may hang in front of, and partly hide, the proximal links.

CLUSTERS ARE EVIDENCE, NOT ANSWERS: preprocessing grouped the robot's pixels into numbered clusters of pixels that
move rigidly together over the whole video (a 3D motion model). A cluster is not a verified link. One link may be split into
several clusters (fragments of one tube); one cluster may span two links (a joint that did not move), or a link together
with a joint housing or an intermediate link; clusters can also be intermediate links, joint housings, cables, the base
or a grasped object. Badges with an "x" (e.g. "7x") mark extra clusters that lie mostly outside the robot mask (grasped
objects, cable loops). Clusters with no badge on the anchor are not visible there.

VERDICTS (one per link)
- assign: exactly one cluster is this link and nothing else: its outline follows the link you see, from one joint housing to the next. clusters: [id]
- merge: several clusters together are exactly this link (fragments of one straight tube, or one rigid group of the summary). clusters: [ids] OR group: "gN" (never both). Code checks the union is one straight link.
- split: one cluster spans exactly this link and its chain neighbour (the joint between them did not move). clusters: [id], split_with: the neighbour label. Give the neighbour the same cluster with verdict split and split_with pointing back.
- emit: the link is clearly visible as its own part, but no cluster or cluster set matches its outline (for example the only candidate cluster also covers a joint housing, an intermediate link or a neighbouring link). A later call draws its box from the image.
- skip: the link is not visible, occluded, or you are uncertain. status "occluded" or "uncertain", a reason code and a confidence.

RULES
1. Look at the clean anchor frame first. Decide which of the three links are actually visible, where each starts and ends
   (at its joint housings), and only then compare with the cluster outlines. Occlusion is expected.
2. Intermediate links are never assigned: their clusters go to unassigned_clusters with why "intermediate_link". Never merge
   them into a neighbour to make a link look complete.
3. Visual evidence over cluster boundaries: assign or merge only when the cluster outline matches the link you see. If the
   best cluster also covers a joint housing, an intermediate link, the gripper or part of a neighbouring link, do not assign
   it as it is: use split when it spans exactly this link and its chain neighbour, otherwise emit so the box is drawn from
   the image.
4. When uncertain, leave it out. Do not assign or emit a link you cannot clearly see as its own part, and never give a link
   a region that really belongs to another link, the elbow, a joint housing or an intermediate link just to complete the
   chain (a common mistake: labelling part of the forearm or the elbow as the upper arm when the upper arm is out of view).
   Skip it with status "occluded" or "uncertain", a reason code and a confidence, and leave its clusters unassigned; leaving
   a link or a cluster out is always acceptable. confidence (0-1) is how sure you are that the link is clearly visible as its
   own part and that your verdict is right. A skip is accepted when that confidence is below 0.5 or the
   reason is occluded_by_gripper or out_of_frame; a skip with higher confidence for another reason goes to the box-drawing
   call instead.
5. Unassigned clusters are allowed and expected (intermediate links, joint housings, grasped object, cables, base, background).
6. A merge is a set of cluster ids or one rigid group id; code computes its geometry.
7. A cluster belongs to at most one link, except the single cluster shared by a split pair.
8. No coordinates anywhere in this answer.
Precision over recall: a wrong link mask is far worse than a missing one.

Reason codes: occluded_by_gripper, out_of_frame, cluster_impure, low_confidence, ambiguous_label.
Unassigned "why" values: intermediate_link, grasped_object, cable, base, background, joint_housing, fingers, other_robot_part.

FRAME TOOL: you may set frame_request to 2 or 3 to see runner-up frame 2 or 3 with the same cluster outlines. Decisions
and cluster IDs still refer to the anchor frame; runner-ups only help you see what the anchor hides. Each runner-up can
be requested once. When you request one, still give your best complete answer now; you will then see the frame and
answer again. Use frame_request null when you are done.

OUTPUT: schema A, JSON only:
{
  "anchor_frame": <anchor frame index>,
  "links": [
    {"label": "upper_arm", "status": "occluded", "verdict": "skip", "clusters": [], "group": null, "split_with": null,
      "reason": "out_of_frame", "confidence": 0.1, "proximal_neighbor": "base", "distal_neighbor": "forearm"},
    {"label": "forearm", "status": "boxed", "verdict": "assign", "clusters": [3], "group": null, "split_with": null,
      "reason": null, "confidence": 0.9, "proximal_neighbor": "upper_arm", "distal_neighbor": "gripper_palm"},
    {"label": "gripper_palm", ...}
  ],
  "unassigned_clusters": [{"id": 7, "why": "grasped_object"}, {"id": 8, "why": "intermediate_link"}],
  "frame_request": null
}
links always has exactly three entries, one each for upper_arm, forearm and gripper_palm. A link you leave
out still gets its entry, with verdict "skip" (as upper_arm in the example above).
status is "boxed" for assign/merge/split/emit and "occluded" or "uncertain" for skip.
\end{promptbox}

Its user message shows the examples, the anchor frame and the trajectory table of the clusters.
\begin{promptbox}
SCHEMA EXAMPLES from the viewpoint bucket {bucket}. Each shows a different robot with its final link boxes drawn and labelled, the exact final answer (box_2d = [ymin, xmin, ymax, xmax] normalized to 0-1000) and a description of every box. They teach the convention of what each link contains and excludes; your own answer uses clusters, not boxes.
[IMAGE: schema example {k}: the exemplar photo with its human-drawn link boxes drawn and labelled, longest side at most 1024 px]
Example {k}: {W}x{H} image above. Final answer:
{exemplar boxes as JSON}
Box descriptions:
- {label}: {description}
ANCHOR FRAME {anchor} (clean), {W}x{H} pixels:
[IMAGE: the anchor frame, clean]
ANCHOR FRAME {anchor} with the robot mask outline (light grey) and numbered cluster outlines (badges off the body, a thin line leads to a dot on the cluster):
[IMAGE: the anchor frame with the robot mask outline and the numbered cluster outlines]
CLUSTER TRAJECTORY TABLE (source pixels; rotations and translation relative to frame {reference frame}; rot_world = 3D rotation angle, rot_inplane = rotation seen in the image, counter-clockwise positive; n/a = no tracks: the cluster was not visible at frame 0)
| frame | cluster | kind | centroid (px) | cluster box [x0,y0,x1,y1] | area % | rot_world (deg) | rot_inplane (deg) | trans (px) | touches | rel_rot_to_touching (deg) |
| --- | --- | --- | --- | --- | --- | --- | --- | --- | --- | --- |
| t{frame} ({role: anchor | runner-up 2 | runner-up 3, when the frame has one}) | c{cluster} | {kind} | ({cx}, {cy}) | [{x0}, {y0}, {x1}, {y1}] | {area %, 1 decimal} | {rot_world} | {rot_inplane} | {trans} | {c<id>, ... or -} | {c<id>: <deg> (*@$\cdot$@*) ... or -} |

SUMMARY
- rigid_groups (move as one rigid body over the whole video; merge candidates, use the group id): none
- code_merges: none
- joint_estimates (experimental, projected screw axis of touching pairs): none
Return schema A as JSON only.
\end{promptbox}

A second call writes the coordinates: a box for each link with the verdict \texttt{emit} and the joint point of each \texttt{split}.
\begin{promptbox}
[IMAGE: the first schema example of the bucket: the exemplar photo with its human-drawn link boxes drawn and labelled, longest side at most 1024 px]
EXAMPLE of the box convention (human annotation, a different scene from the same viewpoint, {W}x{H} image above with the boxes drawn and labelled). Its exact boxes:
{exemplar boxes as JSON}
What each box covers:
- {label}: {description}
Draw your boxes on the anchor frame below with the same convention.
Write coordinates only. The image is the anchor frame, {W}x{H} pixels. Convention: boxes as box_2d = [ymin, xmin, ymax, xmax] and points as point_2d = [y, x], integers normalized to 0-1000 of this image. Box only visible structure, loosely: a box may include background but must contain the whole visible link.
Items:
- label upper_arm, kind box: the long link between the shoulder housing and the elbow housing; both housings excluded.
- label forearm, kind box: the long link between the elbow housing and the first wrist housing; housings excluded.
- label gripper_palm, kind box: the rigid gripper body between the wrist flange and the fingers; fingers, grasped object and wrist housings excluded.
Return JSON only: {"items": [{"label": "...", "kind": "joint_point", "point_2d": [y, x]}, {"label": "...", "kind": "box", "box_2d": [ymin, xmin, ymax, xmax]}]} with exactly one item per line above.
[IMAGE: the anchor frame with the robot mask outline (light grey); for a joint_point item also the outline of that cluster; for a rejected box also that box, labelled]
\end{promptbox}

\paragraph{Points.} For each boxed link of a humanoid, Gemini 3.8 Flash returns one to three positive and up to four negative points on the scale of the crop around the box, and SAM3 segments the link from the box and the points. The system prompt names the link and the arm, repeats the kinematic knowledge, the view block and the viewpoint line of the planner, and adds the click rules, the card of the link and the planner's description of the box. It is shown for the forearm.
\begin{promptbox}
ROLE. You place click points for an interactive segmentation model (SAM) on ONE link of a humanoid robot: the forearm of the image_left arm (the robot's right arm). SAM gets the green box and your points and grows the link mask from them: positive points mark the link, negative points mark nearby parts that the mask must not take.

(*@{\rmfamily\itshape [kinematic, view and viewpoint blocks of the planner]}@*)

CLICK RULES. Positive points lie on the visible surface of the link, well inside its outline (not on an edge),
spread out along it. SAM grows the mask from the positives and carves the area around each negative point out of
the mask, so a negative point never goes on the link and keeps a clear distance from every positive. Use as few
negative points as the mask needs.

FOREARM. Positives along the middle of the long shell, away from both ends, never on a round housing or on the
wrist mechanism. Negatives, where the mask could spill: one in the centre of the elbow housing, one on the wrist
segments, one on the hand if it is close.

TRACE FIRST. Put positives only on the shell that connects to this arm's neighbouring links; a look-alike
in the box (another arm, an object, the torso) is not this link.

THE PLANNER'S DESCRIPTION of this box: "{visible_part}"
\end{promptbox}

The cards of the upper arm and the palm are below.
\begin{promptbox}
UPPER ARM. Positives along the middle of the long shell, away from both ends, never on a round housing or end cap.
Negatives, where the mask could spill: one in the centre of the shoulder housing at the body end, one in the centre
of the elbow housing at the far end (the rounded housing or end cap where the arm bends), one on the torso if the
arm lies against it.

PALM. Positives on the hand body (back of the hand or palm side), away from the finger roots and the wrist.
Negatives: at most two at finger tips, at the far end of the most extended fingers, never at the finger roots or on
the hand body; fingers curled against the palm get none. One on the wrist segment, in its centre, away from the
palm; a negative close to the palm needs a positive on the palm right next to it. One well inside a held object if
the mask could spill into it.
\end{promptbox}

The user message shows one human example of the same link, the anchor frame with the box and the crop around the box.
\begin{promptbox}
[IMAGE: the per-link exemplar: a 768-px crop around the exemplar box (padded like the target crop and grown to hold all its points) with its green box, green positive dots and red negative dots]
EXAMPLE (human annotation of a different robot and scene, the same viewpoint): the image above is a {crop W}x{crop H} crop around its {arm} forearm box (green). Green dots are its positive points (on the link); red dots are its negative points (on parts that must stay out of the mask).
Its positive points as [y, x] in 0-1000 of that crop: {positives [[y, x], ...]}
Its negative points as [y, x] in 0-1000 of that crop: {negatives [[y, x], ...]}
What its box covers ("gripper" means the palm): {description}
TARGET: the anchor frame ({W}x{H}) with the forearm box drawn in green, for context:
[IMAGE: the anchor frame with this link box in green]
TARGET CROP ({crop W}x{crop H}): a zoomed crop around the forearm box (green). Place the points in this crop:
[IMAGE: the target crop: the link box padded 25% per side (palm: 50%), longest side 768 px, only this box drawn in green]
Return JSON only: {"positive": [[y, x], ...], "negative": [[y, x], ...]} with 1-3 positive points inside the green box and 0-4 negative points inside the crop, integers [y, x] normalized to 0-1000 of the TARGET CROP. No other field.
\end{promptbox}

For a single arm, Flash places the positive points of each emitted link, three for the upper arm and the forearm and two for the gripper palm, with the message below.
\begin{promptbox}
You place positive click points for an interactive segmentation model (SAM) on ONE link of a single-arm robot (serial manipulator): the forearm. SAM gets the green box and your points and grows the link mask from them, so every point must lie on the link itself.
THE LINK: forearm = the long link between the elbow and the wrist, toward the gripper. What it covers (the human annotation convention): the long link between the elbow housing and the first wrist housing; housings excluded.
Place exactly 3 points on the forearm link itself, between the elbow housing and the first wrist housing. Each point lies on the visible surface of the link, well inside its outline (not on an edge), and the points are spread out over the link. Never place a point on background, on a neighbouring link, on a round or cylindrical joint housing, on an intermediate wrist segment, on the fingers, on a grasped object or on a loose cable.
[IMAGE: the per-link exemplar: a 768-px crop around the exemplar box (padded 15% per side) with its green box, green positive dots and red negative dots]
EXAMPLE (human annotation of a different robot and scene from the same viewpoint): the image above is a {crop W}x{crop H} crop around its {label} box (green). Green dots are its positive points (on the link); red dots are its negative points (on neighbouring parts that must stay out of the mask), placed by other means: you only give positive points.
Its positive points as [y, x] in 0-1000 of that crop: {positives [[y, x], ...]}
Its negative points as [y, x] in 0-1000 of that crop: {negatives [[y, x], ...]}
What its box covers: {description}
TARGET: the anchor frame of the video ({W}x{H}) with the forearm box drawn in green (context only).
[IMAGE: the anchor frame with this link box in green]
TARGET CROP ({crop W}x{crop H}) around the forearm box (green). Place the 3 positive points in this crop.
[IMAGE: the target crop: the link box padded 15% per side, longest side 768 px, box drawn in green]
Return JSON only: {"label": "forearm", "points": [[y, x], ...]} with exactly 3 points, integers [y, x] normalized to 0-1000 of the TARGET CROP. No other field.
\end{promptbox}

\paragraph{Critic.} Claude Opus 5.5 reviews each mask and answers \texttt{accept} or \texttt{refine}, and a refine carries the points of one more SAM3 pass with the same box. For a humanoid, the system prompt is that of the point placer followed by the role below, and a refine keeps points of the first pass by their numbers and adds new ones.
\begin{promptbox}
YOUR ROLE NOW: the critic. SAM has made a mask for this link from the round-1 points. Judge it.
- The mask is right when it covers the visible part of the link as the link card says. It holds no intermediate
  link, neighbouring link, torso, held object or background.
- Check it part by part: (1) it lies on this arm's own chain, connected to the neighbouring links, not on another
  arm or machine; (2) it holds no joint housing at either end (a round end cap, a disc, or a housing that bulges
  wider than the shell); (3) it does not spill onto the torso, a held or nearby object or the
  background.
- A code hint that a negative point lies inside the mask, or that the mask leaks out of the box, means the mask
  covers a part that was meant to stay out. Look at that part, and refine unless it really belongs to the link.
- Small inaccuracies along the outline are fine. If the mask is fine, accept it; do not refine to polish the outline.
- Refine only when the mask is clearly wrong: it misses a large part of the link, or it covers one of those parts.
A refine gives the complete final point set: the round-1 points you keep, by their numbers (P1, P2, ... positive;
N1, N2, ... negative), plus new points. Keep the good points; a point that is wrong is dropped and, if needed,
replaced by a new one. Points are never moved. In total 1-3 positive and 0-4 negative points.
SAM then runs again with the same box and only the final set.
\end{promptbox}
\begin{promptbox}
[IMAGE: the per-link exemplar crop (the same image as in 3H)]
EXAMPLE (human annotation of a different robot and scene, the same viewpoint): the image above is a {crop W}x{crop H} crop around its {arm} forearm box (green). Green dots are its positive points (on the link); red dots are its negative points (on parts that must stay out of the mask).
Its positive points as [y, x] in 0-1000 of that crop: {positives [[y, x], ...]}
Its negative points as [y, x] in 0-1000 of that crop: {negatives [[y, x], ...]}
What its box covers ("gripper" means the palm): {description}
TARGET: the anchor frame ({W}x{H}) with the forearm box (green) and the round-1 mask (filled magenta), for context:
[IMAGE: the anchor frame with the link box (green) and the round-1 mask (filled magenta)]
TARGET CROP ({crop W}x{crop H}) around the forearm box (green), clean:
[IMAGE: the target crop, clean, box in green (the same crop as in 3H)]
The same TARGET CROP with the round-1 mask (filled magenta) and the round-1 points, numbered (P = positive, green; N = negative, red):
[IMAGE: the same crop with the round-1 mask (magenta), the box and the numbered round-1 points (P green, N red)]
The round-1 points as [y, x] in 0-1000 of the TARGET CROP:
P1: [{y}, {x}]
P2: [{y}, {x}]
N1: [{y}, {x}]
Code checks on this mask: nothing flagged.
Answer JSON only, one of:
{"verdict": "accept"} when the mask is right (the expected answer for a mask that is fine);
{"verdict": "refine", "keep": ["P1", "N2"], "positive": [[y, x]], "negative": [[y, x]]} when it is clearly wrong. "keep" lists the round-1 points you keep; "positive" and "negative" hold only the new points (empty lists when there are none). New coordinates are integers [y, x] normalized to 0-1000 of the TARGET CROP. No other field.
\end{promptbox}

For a single arm, the critic receives the message below, and a refine gives the complete new set of points.
\begin{promptbox}
You review ONE link mask of a single-arm robot (serial manipulator): the forearm. An interactive segmentation model (SAM) made the mask from the green box and the click points shown (green = positive, red = negative).
THE LINK: forearm = the long link between the elbow and the wrist, toward the gripper. What it covers (the human annotation convention): the long link between the elbow housing and the first wrist housing; housings excluded.
The mask is right when it covers the visible link as the convention says: the whole visible link, and nothing of a neighbouring link, a joint housing, an intermediate wrist segment, the fingers, a grasped object or background. Small inaccuracies along the outline are fine.
[IMAGE: the per-link exemplar crop with its green box, green positive dots and red negative dots (crop grown to hold all its points)]
EXAMPLE (human annotation of a different robot and scene from the same viewpoint): the image above is a {crop W}x{crop H} crop around its {label} box (green). Green dots are its positive points (on the link); red dots are its negative points (on neighbouring parts that must stay out of the mask).
Its positive points as [y, x] in 0-1000 of that crop: {positives [[y, x], ...]}
Its negative points as [y, x] in 0-1000 of that crop: {negatives [[y, x], ...]}
What its box covers: {description}
TARGET: the anchor frame of the video ({W}x{H}) with the forearm box (green) and the current mask (filled magenta), for context.
[IMAGE: the anchor frame with the link box (green) and the current mask (filled magenta)]
TARGET CROP ({crop W}x{crop H}) around the forearm box (green), clean:
[IMAGE: the target crop, clean, box in green (box padded 15% per side; gripper palm 50%; longest side 768 px)]
The same TARGET CROP with the current mask (filled magenta) and the current points (green positive, red negative):
[IMAGE: the same crop with the current mask (magenta) and the current points (green positive, red negative)]
Code checks on this mask: nothing flagged.
Answer JSON only, one of:
{"label": "forearm", "verdict": "accept"} when the mask is right;
{"label": "forearm", "verdict": "refine", "positive": [[y, x], ...], "negative": [[y, x], ...]} otherwise.
A refine REPLACES every current point: give the complete new set, 1-3 positive points on the link and 0-5 negative points on the parts the mask wrongly covers or spills into (for example a joint housing or a neighbouring link it swallowed). SAM then runs again with the same box and only your points. Positive points lie on the visible surface of the link, well inside its outline and spread out, never on background, a neighbouring link, a joint housing, an intermediate wrist segment, the fingers, a grasped object or a loose cable. Negative points go on exactly such parts, where the mask covers them or could spill into them.
Coordinates are integers [y, x] normalized to 0-1000 of the TARGET CROP (the clean crop and the mask crop share them). No other field.
\end{promptbox}

Gemini 3.8 Flash also picks the anchor frame and the viewing direction that selects the examples, and in first-person views it places and reviews the points of the thumb and the index finger with the same layout of messages.

\paragraph{EVIDENCE\_REPAIR.} Gemini 4 Argon receives the prompt below, followed by five frames of the video. It picks one of four actions and names the object and the earliest shown frame where the object is clearly visible.
\begin{promptbox}
A robot manipulation video (generated, so objects may look odd). An automatic contact check needs two masks: the object the robot manipulates, and the robot's end-effector (gripper or hand). It failed on this video:
{failed}

Task prompt: {task}
The {n} frames are in time order: frames {frames} of {total}.

Pick ONE action:
- "box_object": the manipulated object is clearly visible in at least one of these frames, so a box around it can replace the missing mask.
- "mask_robot": the robot's gripper or hand is visible and the object mask is not the problem; re-mask the robot.
- "both": both masks need replacing.
- "abstain": the manipulated object never appears, or no robot-object interaction can be judged from the video.

Answer with JSON only:
{"action": "...", "object": "<manipulated object, at most 5 words, or null>", "object_frame": <one of {frames}: the earliest shown frame where the object is clearly visible, or null>, "reason": "<at most 15 words>"}
\end{promptbox}

\texttt{\{failed\}} is one of the two lines below, by the state of the two masks.
\begin{promptbox}
the object mask is missing: a text detector prompted with "{target}" found nothing on any frame.the object ("{target}") was masked, but no frame gave a distance between the robot end-effector mask and the object mask (on humanoids the hand mask is often empty).
\end{promptbox}

Gemini 3.8 Flash boxes the object on the named frame with the prompt below, and SAM3 tracks the box in both directions. For \texttt{mask\_robot}, SAM3 segments the whole robot from the text ``robot''. \mContact{} then reruns on these masks.
\begin{promptbox}
One frame of a robot manipulation video. Box the object the robot manipulates: "{name}".
Task prompt: {task}
If several look alike, box the one the task is about (the one the robot reaches for). Box the whole object, tightly, and nothing of the robot.
Answer with JSON only: {"box_2d": [ymin, xmin, ymax, xmax]} with integers 0-1000 relative to the image, or {"box_2d": null} if the object is not in this frame.
\end{promptbox}

Flash then checks the tracked mask against the motion on four frames, with the box of the mask on each frame given as text.
\begin{promptbox}
{n} frames of a robot manipulation video, in time order: frames {frames} of {total}. The video may be generated, so objects can look odd.
Task prompt: {task}
An automatic tracker marks one object through the video. Where it is marked, as box_2d [ymin, xmin, ymax, xmax] with integers 0-1000 relative to the image:
image 1 (frame {t1}): {box_2d or "no mask in this frame"}
...

Is the marked object the object the robot actually picks up or moves in this video?
- "moved_object_is_masked": true if, in the first image where something is marked, the marked region is on the object the robot picks up or moves. False if the robot moves a different object, or the marked region is mostly the robot. If nothing moves, say whether it is the object the task is about.
- "nothing_moves": true if the robot moves no object at all in these frames.
- "mask_jumps": true if the marked region is on different things in different frames.
Answer with JSON only: {"moved_object_is_masked": true|false, "nothing_moves": true|false, "mask_jumps": true|false, "reason": "<at most 15 words>"}
\end{promptbox}

Flash also boxes the moved object on its earliest unobstructed view, from the boxed frame and two later frames.
\begin{promptbox}
{n} frames of a robot manipulation video, in time order: frames {frames} of {total} (image 1 = frame {first}). The video may be generated, so objects can look odd.
Object: "{name}". Task prompt: {task}
The name and its position words may be inaccurate; what counts is which object the robot actually moves.

Box the one object the robot picks up or moves in this video.
- Use the later frames to see which object the robot ends up holding or moving, or which one has left its place. If several similar objects are present, do not just take the one nearest the gripper in the first frame.
- Box that object on the earliest frame where it is fully visible and not covered by the robot, even if you identified it from a later frame. If it only appears in later frames (generated objects can appear or duplicate), box it on the frame where it is clearest.
- Box the object only, tightly. Never include any part of any robot (gripper, fingers, hand, arm).
- Set "none_moves" to true only if no object moves in any frame; then box the object the task is about on the earliest frame where it is visible.
Answer with JSON only: {"frame": <one of {frames}>, "box_2d": [ymin, xmin, ymax, xmax], "none_moves": <true|false>, "reason": "<at most 15 words>"} with box_2d integers 0-1000 relative to that image, or {"frame": null, "box_2d": null, "none_moves": null, "reason": "..."} if the object is in none of the frames.
\end{promptbox}

\paragraph{EDGE\_CASE\_INSPECTION.} Gemini 4 Argon receives the system prompt below. The user message gives the geometry of both masks as text, followed by up to 14 frames, each shown plain and with the object mask outlined in green and the robot mask filled in red, and \cref{sec:supp_agent_run} prints one user message in full. The answer names the causes with their frame numbers and the actions. The code applies \texttt{re\_reference}, \texttt{robot\_absent} and \texttt{object\_owns\_overlap} to the reference and the masks, and \texttt{no\_repair} keeps the result of the video.
\begin{promptbox}
You are the controller of an object-deformation check on robot manipulation videos. Many videos are generated, so robots and objects may look odd, and an object may really change.

HOW THE CHECK WORKS
It compares the manipulated object in later frames with one REFERENCE frame, using two masks on every frame: the OBJECT mask and the ROBOT mask (the gripper, hand or arm near the object). It needs:
1. a reference frame that shows the whole object, not hidden, not covered by the robot mask, on a frame where the robot mask exists;
2. after the first 20% of the video, frames where the object mask is on the object, so that the visible object (object mask minus robot mask) can be laid over the reference.
Both masks mark visible pixels only. When both are right they do not overlap, so a large "% of object under robot" means one of the two masks is wrong on that frame; the images show which one.

On this video the check found no usable reference frame, or could not pair any frame after the 20% mark with it. You get what failed, per-frame mask numbers, and some frames. Each image shows one frame twice: left as it is, right with the OBJECT mask outlined in green and the ROBOT mask filled in red. Decide what is wrong and choose the repairs.

CAUSES AND THEIR ACTIONS
A. PARTIAL REFERENCE. The reference frame shows only part of the object (partly hidden, cut by the image border, or only partly masked), or a clearly different side of it than the rest of the video, and a later frame shows the same object whole. Usual sign: the object's mask area grows a lot afterwards (about 3.5x or more is a strong sign, less can still be this cause) although the object did not come closer to the camera.
   Action "re_reference": the EARLIEST shown frame where the whole object is visible and free of the robot ("frame"), and the last shown frame before it where it is still partial ("partial_until"). The reference must be as early as possible, preferably before the 20% mark: everything before the reference is not checked, and the frames after the 20% mark are the ones compared with it. Go past the 20% mark only when no earlier shown frame shows the whole object.
   A reliable sign of "a different side": both masks are right, the object is in view after the 20% mark, yet the outline fit IoU of those frames against the current reference stays below 0.75. The later frames then show the object differently from the reference: choose the earliest shown frame that shows the whole object as those later frames show it.
   Not this cause: the object is already whole in the reference and later only comes closer to the camera (the same side, just larger). Not this cause either: the only problem of the reference is the robot mask (missing, or lying on the object); that is cause B or C, and the reference stays.
B. ROBOT MASK MISSING. On frames where a robot is visible, the robot mask is empty, or it sits somewhere other than on the robot part that handles the object.
   Action "remask_robot": the earliest shown frame where the robot part that handles the object is visible but not (or wrongly) masked ("frame"), and "replace": true when the existing red mask is not on the robot at all (it is dropped), false when it is on the robot but incomplete or missing on some frames (the new mask is added to it). A helper model boxes the robot on that frame and the mask is propagated through the video.
   If instead no robot is in view on the frames without a robot mask, action "robot_absent" with the frame range ("from", "to"): those frames then count as having a valid, empty robot mask.
C. ROBOT MASK ON THE OBJECT. The red mask covers the object (often from the moment it is grasped) while the green outline is on the object: the object's pixels were given to the robot.
   Action "object_owns_overlap" with the frame range ("from", "to"): the overlapping pixels are removed from the robot mask.
D. OBJECT MASK LOST OR WRONG. The green outline disappears, shrinks to a fragment, or sits on something that is not the object, while the object is still visible in the frame.
   Action "remask_object": the earliest shown frame where the object is visible but not correctly masked ("frame"). A helper model boxes the object there and the new mask replaces the old one from that frame on.
E. NOTHING TO REPAIR. Both masks are right on the shown frames and the reference already shows the whole object as the later frames show it; or the object really is hidden, out of view or gone for the later part of the video.
   Action "no_repair": the video keeps the result it has.
   Not E: the object is in view after the 20% mark with a correct mask, but looks different from the reference there (another side, another pose, a larger or changed shape). Whatever the reason for the different look, that is cause A: give the earliest shown frame that shows the whole object as the later frames show it. Whether the object really changed is decided by the check afterwards, not by you.

RULES
- Several causes can hold at once. Give every action that is needed, each action at most once, and the fewest that make requirements 1 and 2 hold. "no_repair" stands alone.
- Name only frames that are shown as images ("from" and "to" may be any frame).
- Keep the current reference when it already shows the whole object; never move it later for a slightly better view.
- Judge from the images first; the numbers tell you where to look.
- Do not judge whether the object deforms. That is the check's job, run after your repairs. If the object looks different later, still give the repairs that make the reference and the masks right.

Answer with JSON only:
{"object": "<the manipulated object, at most 5 words>",
 "causes": [{"cause": "<A | B | C | D | E>", "evidence": "<at most 25 words, with frame numbers>"}],
 "actions": [{"action": "<re_reference | remask_robot | robot_absent | object_owns_overlap | remask_object | no_repair>", "frame": <int or null>, "partial_until": <int or null>, "replace": <true, false or null>, "from": <int or null>, "to": <int or null>}],
 "confidence": "<high | medium | low>"}
\end{promptbox}

For \texttt{remask\_object}, Gemini 3.8 Flash boxes the object on the named frame with the first prompt below, and for \texttt{remask\_robot} it boxes every visible robot part with the second. SAM3 tracks each box through the video.
\begin{promptbox}
You see one frame of a video of a robot doing a manipulation task. Find the object the robot manipulates: "{name}". The video may be generated, so the object can look different from its name; it is the thing the robot holds, moves or is about to grasp.

Give one tight box around the visible part of that object. Do not include the robot's gripper, hand or arm, other objects, or the surface under it.

Coordinates are integers 0-1000 relative to the full frame: x from the left edge, y from the top edge; the box is [x1, y1, x2, y2] with x1 < x2 and y1 < y2. When the object is not visible in this frame, answer "visible": false and "box_xyxy": null.

Return only this JSON:
{"visible": true or false, "box_xyxy": [x1, y1, x2, y2] or null, "reason": "<one sentence>"}
\end{promptbox}
\begin{promptbox}
You see one frame of a video of a robot doing a manipulation task. Box every visible part of the robot: its arms, hands, fingers, grippers and body. Give each separately visible piece its own tight box (for example, a left hand and a right hand are two boxes; an arm with its hand is one box when they are connected in the image). Do not box people or human hands, the objects the robot holds or touches, the table or the background.

Coordinates are integers 0-1000 relative to the full frame: x from the left edge, y from the top edge; a box is [x1, y1, x2, y2] with x1 < x2 and y1 < y2. When no part of the robot is visible, give an empty list.

Return only this JSON:
{"robot_parts": [{"part": "<short name, for example right robot hand>", "box_xyxy": [x1, y1, x2, y2]}], "reason": "<one sentence>"}
\end{promptbox}

\subsection{Audit agent: one complete execution}
\label{sec:supp_agent_run}

This section follows the apple of \cref{fig:metric_vis}c through EDGE\_CASE\_INSPECTION (\cref{sec:supp_agent}), from the logged run. The video is a HunyuanVideo~1.5 generation of NVIDIA GR1 task 43 with 121 frames, in which the right hand lifts a red apple from the pink plate and carries it to the teal plate. The harness sends a video to the agent when the pairing selects no frame after the 20\% mark, as here.

\begin{table*}[t]
\centering
\caption{\textbf{Masks of the apple before and after the repair.} Pixels of the object mask and of the robot mask, the share of the object under the robot mask, the visible object pixels and the silhouette-fit IoU against the reference, which is frame 0 before the repair and frame 16 after it. The object mask is the same before and after.}
\label{tab:supp_apple}
\footnotesize
\setlength{\tabcolsep}{2pt}
\begin{tabular*}{\textwidth}{@{\extracolsep{\fill}}r r rr rr rr cc@{}}
\toprule
 & & \multicolumn{2}{c}{Robot pixels} & \multicolumn{2}{c}{Object under robot (\%)} & \multicolumn{2}{c}{Visible object pixels} & \multicolumn{2}{c}{Silhouette-fit IoU} \\
Frame & Object pixels & Before & After & Before & After & Before & After & Before & After \\
\midrule
16 & 2,265 & 11,364 & 11,364 & 0.7 & 0.7 & 2,249 & 2,249 & 0.89 & 1.00 \\
17 & 2,714 & 11,341 & 11,341 & 1.5 & 1.5 & 2,672 & 2,672 & 0.87 & 0.88 \\
19 & 3,487 & 14,187 & 10,851 & 95.7 & 0.0 & 151 & 3,487 & 0.57 & 0.87 \\
24 & 4,481 & 16,804 & 12,332 & 99.8 & 0.0 & 9 & 4,481 & 0.22 & 0.91 \\
29 & 4,850 & 17,021 & 12,188 & 99.6 & 0.0 & 17 & 4,850 & 0.29 & 0.92 \\
42 & 4,491 & 16,430 & 11,950 & 99.8 & 0.0 & 11 & 4,491 & 0.29 & 0.90 \\
70 & 4,156 & 16,571 & 12,456 & 99.0 & 0.0 & 41 & 4,156 & 0.42 & 0.85 \\
120 & 2,011 & 13,427 & 11,427 & 99.5 & 0.0 & 11 & 2,011 & 0.50 & 0.87 \\
\bottomrule
\end{tabular*}
\end{table*}

\paragraph{Evidence.} The agent receives the system prompt of \cref{sec:supp_agent_prompts}, the text below and 13 frames (0, 5, 10, 14, 17, 19, 24, 29, 42, 60, 78, 96 and 120), each shown plain and with the object mask outlined in green and the robot mask filled in red. The text gives the reference frame, a summary, and one row for each of 33 frames: the object pixels, the robot pixels, the share of the object under the robot mask, the visible object pixels, the silhouette-fit IoU against the reference, named outline fit IoU in the prompt, and whether the frame is paired. A frame is paired when the fit chosen for it places at least 80\% of its visible pixels on the reference (\cref{sec:supp_objdef}). The IoU column is that of the silhouette fit, and its threshold of 0.75 applies when that fit is chosen; frame 10 pairs by the box mapping, with 851 of its 853 visible pixels inside the mapped box. The counts in the text are over all 121 frames: the 24 paired frames are frames 0 to 17 and frames 86, 93, 96, 101, 106 and 109, and the 95\% rule removes frames 18 to 120 from the selection, those six included.
\begin{promptbox}
Video: 121 frames (0-120); the 20% mark is frame 24.
Task prompt: The robot closes its right hand around the red apple on the pink plate, lifts it clear, and carries it over to the teal plate. It lowers the apple onto the teal plate and opens its fingers, leaving the apple on the plate with the hand still nearby. Keep the camera fixed; preserve robot geometry, physically valid joint motion, and normal gripper opening and closing.

Object the masks were made for: "red apple".
Robot mask parts: left_palm, right_palm, left_index_finger, left_thumb, right_index_finger, right_thumb.
WHAT FAILED: Reference = frame 0 (object 1547 px, 0% of it under the robot mask; more than 5% rejects a reference). No frame after the 20% mark could be paired with it. Of 121 frames: 24 paired (all before the 20% mark), 103 dropped because more than 95% of the object is under the robot mask, 0 dropped because less than 15% of the object is visible.
SUMMARY
- object first moves (point tracks): frame 12
- object mask area: frame 0 1547 px; largest 4850 px at frame 29; largest / first frame with both masks = 3.14x
- frames with an empty object mask: none
- frames with no robot mask: none
- frames with more than 95% of the object under the robot mask: 18-120 (103 frames)
- frames with 5-95% of the object under the robot mask: none
PER-FRAME NUMBERS (* = shown as an image; outline fit IoU against the current reference, 0.75 needed)
frame | object px | robot px | % of object under robot | visible object px | outline fit IoU | paired
0* | 1547 | 13248 | 0% | 1547 | 1.00 | yes
5* | 1586 | 10719 | 0% | 1586 | 0.89 | yes
10* | 857 | 11103 | 0% | 853 | 0.72 | yes
14* | 1061 | 11317 | 2% | 1043 | 0.90 | yes
15 | 1846 | 11188 | 1% | 1822 | 0.88 | yes
17* | 2714 | 11341 | 2% | 2672 | 0.87 | yes
19* | 3487 | 14187 | 96% | 151 | 0.57 | no
20 | 3635 | 14675 | 97% | 121 | 0.43 | no
24* | 4481 | 16804 | 100% | 9 | 0.22 | no
25 | 4684 | 16870 | 99% | 39 | 0.35 | no
29* | 4850 | 17021 | 100% | 17 | 0.29 | no
30 | 4720 | 17156 | 100% | 0 | - | no
35 | 4714 | 16980 | 100% | 10 | 0.30 | no
40 | 4582 | 16793 | 100% | 6 | 0.17 | no
42* | 4491 | 16430 | 100% | 11 | 0.29 | no
45 | 4108 | 15046 | 99% | 46 | 0.37 | no
50 | 3571 | 13624 | 99% | 19 | 0.33 | no
55 | 3607 | 13477 | 99% | 54 | 0.32 | no
60* | 3808 | 14295 | 100% | 7 | 0.43 | no
65 | 4010 | 15471 | 99% | 57 | 0.41 | no
70 | 4156 | 16571 | 99% | 41 | 0.42 | no
75 | 3806 | 17809 | 100% | 19 | 0.44 | no
78* | 3735 | 17217 | 99% | 45 | 0.72 | no
80 | 3264 | 16027 | 100% | 6 | 0.33 | no
85 | 2412 | 13159 | 99% | 30 | 0.32 | no
90 | 2106 | 11988 | 100% | 9 | 0.38 | no
95 | 2065 | 12100 | 100% | 0 | - | no
96* | 2086 | 12116 | 100% | 1 | 1.00 | yes
100 | 2002 | 12300 | 100% | 4 | 0.33 | no
105 | 1826 | 12071 | 100% | 0 | - | no
110 | 1825 | 12329 | 99% | 24 | 0.56 | no
115 | 1969 | 13063 | 100% | 9 | 0.47 | no
120* | 2011 | 13427 | 99% | 11 | 0.50 | no
The 13 images follow in time order: frames 0, 5, 10, 14, 17, 19, 24, 29, 42, 60, 78, 96, 120.
\end{promptbox}

\paragraph{Diagnosis and action.} Argon names two causes and two actions with high confidence. The apple is partly occluded by the hand in frames 0 to 14 and fully visible at frame 17, so it chooses \texttt{re\_reference} with frame 17. The robot mask covers the apple from frame 18 to 120, so it chooses \texttt{object\_owns\_overlap} for frames 18 to 120.
\begin{promptbox}
{"object": "red apple", "causes": [{"cause": "A", "evidence": "The apple is partly occluded by the robot hand in frames 0-14 and becomes fully visible at frame 17."}, {"cause": "C", "evidence": "The red robot mask covers the apple inside the green object outline from frame 18 to 120 (96-100% overlap)."}], "actions": [{"action": "re_reference", "frame": 17, "partial_until": 14, "replace": null, "from": null, "to": null}, {"action": "object_owns_overlap", "frame": null, "partial_until": null, "replace": null, "from": 18, "to": 120}], "confidence": "high"}
\end{promptbox}

\paragraph{Repair.} The code applies both actions. \texttt{object\_owns\_overlap} returns the apple's pixels from the robot mask on frames 18 to 120, 339,330 pixels on 103 frames, and leaves the object mask unchanged. At frame 24, the robot mask covers 4,472 of the apple's 4,481 pixels (99.8\%) before the repair and none after it. Over frames 18 to 120, the visible apple pixels rise from 2,724 to 342,054. The reference moves to frame 16, the earliest frame from 15 to 17 with at least 80\% of the object area of frame 17 and at most 5\% of it under the robot mask. \Cref{tab:supp_apple} gives the masks before and after the repair.

\paragraph{Rerun.} The occlusion test, the MoSE3 pass and the pairing rerun on the repaired masks. All 105 frames from 16 to 120 pair with the new reference, and the pairing selects 10 of them: 27, 29, 69, 70, 73, 74, 97, 116, 117 and 119. AnomalyDINO scores the 10 pairs between 0.51 and 0.59, and the highest, 0.59 at frame 70, is the score shown in \cref{fig:metric_vis}c.

\subsection{Learned judges}

\paragraph{PhyJudge-9B.} Each call has the system prompt below and one user message with the video at 4 frames per second followed by the text, where \texttt{\{prompt\}} is the generation prompt of the video. The \mContact{} column uses the inertia call, \mSolid{} the impenetrability call, \mPersist{} the persistence call and \mObjDef{} the material call.

\begin{promptbox}
You are a strict video evaluation model.
\end{promptbox}
\begin{promptbox}
Evaluate physical realism for one physical law: inertia.

Criterion:
Do stationary objects remain still unless acted upon? Do moving objects maintain their motion unless stopped by friction, collision, or an obstacle?

Caption, for context only:
"{prompt}"

Sub-questions to consider in your mind before scoring:
q1: Does any object spontaneously start moving without a visible cause?
q2: Does any moving object suddenly stop or reverse direction with no visible contact or force?

Judge the video itself. Do not penalize prompt mismatch unless it affects whether this physical law can be evaluated.

Score 1-5:
5=clearly correct
4=mostly correct with minor issues
3=partially correct or ambiguous
2=mostly incorrect
1=severely incorrect

Then output ONLY a JSON object with exactly one key: inertia.

Example:
{"inertia": 3}
\end{promptbox}
\begin{promptbox}
Evaluate physical realism for one physical law: impenetrability.

Criterion:
Do objects maintain impenetrability (*@\textemdash{}@*) no passing through each other?

Caption, for context only:
"{prompt}"

Sub-questions to consider in your mind before scoring:
q1: Does any object pass through another solid object as if it were not there?
q2: Does any part of an object clip into a wall, floor, or another object's interior?

Judge the video itself. Do not penalize prompt mismatch unless it affects whether this physical law can be evaluated.

Score 1-5:
5=clearly correct
4=mostly correct with minor issues
3=partially correct or ambiguous
2=mostly incorrect
1=severely incorrect

Then output ONLY a JSON object with exactly one key: impenetrability.

Example:
{"impenetrability": 3}
\end{promptbox}
\begin{promptbox}
Evaluate Object Persistence.

Caption, for context only:
"{prompt}"

The video was generated using a text+image-to-video (ti2v) model, conditioned on the first frame and the text prompt above.

Sub-questions to consider in your mind before scoring:
q1: Do objects maintain consistent existence throughout the video?
q2: Do objects keep a stable shape, size, color, and texture?
q3: Do objects avoid disappearing, appearing, or transforming unexpectedly?
q4: Do objects preserve identity through motion and brief occlusion?

Score 1-5:
5=fully consistent
4=mostly consistent with minor flicker
3=noticeable issues
2=major inconsistencies
1=severe disappearance or identity changes

Then output ONLY a JSON object with exactly one key: persistence.

Example:
{"persistence": 4}
\end{promptbox}
\begin{promptbox}
Evaluate physical realism for one physical law: material.

Criterion:
Does each material respond according to its properties? (glass shatters, rubber bounces, metal is rigid, cloth deforms softly, etc.)

Caption, for context only:
"{prompt}"

Sub-questions to consider in your mind before scoring:
q1: Does any rigid material (glass, metal, stone) bend or stretch like rubber?
q2: Does any soft material (cloth, rubber, rope) behave as if it were completely rigid and unbending?

Judge the video itself. Do not penalize prompt mismatch unless it affects whether this physical law can be evaluated.

Score 1-5:
5=clearly correct
4=mostly correct with minor issues
3=partially correct or ambiguous
2=mostly incorrect
1=severely incorrect

Then output ONLY a JSON object with exactly one key: material.

Example:
{"material": 3}
\end{promptbox}

\paragraph{WorldModelBench.} The judge sees 16 uniformly sampled frames and no caption, and the score is its yes or no answer. The three prompts give the \mContact{}, \mSolid{} and \mObjDef{} columns.

\begin{promptbox}
Watch the video and determine if it shows any 'violation of newton's law: objects move without any external force.'
            Answer with "Yes" or "No".
\end{promptbox}
\begin{promptbox}
Watch the video and determine if it shows any 'violation of non-physical penetration: objects unnaturally pass through each other.'
            Answer with "Yes" or "No".
\end{promptbox}
\begin{promptbox}
Watch the video and determine if it shows any 'violation of the law of conservation of mass or solid constitutive law: objects deform irregularly.'
            Answer with "Yes" or "No".
\end{promptbox}

\paragraph{RoboJudge-9B.} One call per video, with the system prompt below and a user message holding the video followed by the text. The violation score is 6 minus the returned physical adherence and fills every column.

\begin{promptbox}
You are a strict, calibrated evaluator of the PHYSICAL REALISM of AI-generated embodied / robot-manipulation videos (a robot arm/gripper or a human hand acting on objects). You are shown uniformly-sampled frames of one generated video in temporal order. Judge the physics of the video itself. Be conservative: reserve 5 for clearly flawless physics and 1 for clearly broken physics.
\end{promptbox}
\begin{promptbox}
Task: Judge the PHYSICAL REALISM of this AI-generated robot / embodied-manipulation
video, from the video alone (ignore any task instruction).

Criteria (your reasoning must address each; you may also note other issues):
1. Agent integrity - the arm/gripper/hand stays structurally complete and consistent
   (no melting, fused/extra fingers, warping).
2. Scene & object consistency - background and objects stay temporally stable
   (no flicker, teleport, morphing, appear/disappear).
3. Interaction realism - contacts obey physics (grasps close and bear weight, no
   interpenetration, motion respects gravity/inertia).

Score (integer 1-5): 1 = gross violations throughout; 2 = major violations;
3 = noticeable local inconsistencies; 4 = minor issues only; 5 = no visible violation.

Reason first, then score. Output JSON only:
{"reasoning": "<assess agent integrity, scene & object consistency, and interaction realism, each with concrete visual evidence>", "physical_adherence": <1-5>}
\end{promptbox}

\subsection{Prior evaluators}

\paragraph{RBench.} Qwen3.8-27B with thinking off answers the physical question below on a grid of six frames. \texttt{\{view\}} is the camera view and \texttt{\{description\}} the task prompt. The task-adherence and robot-stability sub-scores use RBench's own prompts unchanged.

\begin{promptbox}
The provided image presents sequential frames, arranged in a grid, from a {view} perspective AI-generated task video about {description}
Does this video comply with common-sense expectations for human-level interactions? 
 
# A. Anomaly Checks:
1) Physical grounding violation: any part of the robot appears floating, or intersecting/penetrating other geometry.
2) Spontaneous object appearance: any object or robotic agent that suddenly appears between frames without a plausible cause. Pay special attention to the top two frames, where such anomalies often occur early in the sequence.
3) Non-contact attachment / false grasp: If the video involves a grasping interaction, carefully check whether the object remains attached to or moves with the robotic gripper/hand without clear physical contact, closure.
If either anomaly is present, treat it as a major violation and assign a low score (1(*@\textendash{}@*)2). 
 
# B. Human-Level Common-Sense Consistency:
Evaluate whether this video complies with common-sense expectations for human-level interactions?
Rate the video on a scale from 1 to 5, where 5 means fully consistent with human common sense and 1 means major violations. 
 
### Special Notes:
1) Be very cautious when assigning scores of 4 or 5: these scores should only be used when clearly correct, and free of errors. Do not give 4 or 5 lightly.
2) Use step-by-step reasoning internally to make your selection.
3) Your output must be a valid JSON object with two fields:
- "reason": a breif justification for the given score
- "score": an integer between 1 and 5
\end{promptbox}

\paragraph{VideoCon-Physics.} The model reads 32 frames and the fixed prompt below, and the score is the probability of yes against no for the next token.

\begin{promptbox}
The following is a conversation between a curious human and AI assistant. The assistant gives helpful, detailed, and polite answers to the user's questions.
Human: <|video|>
Human: Does this video follow the physical laws?
AI: 
\end{promptbox}

\paragraph{PAI and PDI.} Neither queries a language model with a prompt. PAI computes its four scores with no text input, and PDI receives the \mContact{} target name as the text query of its object detector.

\section{Versions of external models and data}
\label{sec:supp_versions}

\paragraph{Models of our metrics.} SAM3 runs the code of the \texttt{sam3}~0.1.4 release with \texttt{sam3.pt}, and RobotSeg \texttt{robotseg.pt} of \path{showlab/RobotSeg} at \texttt{dafb8c0} throughout. \mContact{} runs \path{Qwen/Qwen3-VL-8B-Instruct} at \texttt{0c351dd} in bfloat16 and CoTracker3 with \path{scaled_offline.pth} of \path{facebook/cotracker3}. \mPersist{} uses GroundingDINO SwinB with \path{groundingdino_swinb_cogcoor.pth} and SigLIP \path{google/siglip-so400m-patch14-384} at \texttt{9fdffc5}. \mRobotDef{}, \mObjDef{} and \mSolid{} run MoSE3 at commit \texttt{c39ef00} of its repository with the weights \path{mose3-tracker/MoSE3} at \texttt{5280ca2}, and \mObjDef{} uses the same CoTracker3 checkpoint as \mContact{}, AnomalyDINO from commit \texttt{b9d1c26} and \path{facebook/dinov2-base} at \texttt{f9e44c8}. Gemini 4 Argon and Gemini 3.8 Flash are called as \texttt{gemini-4-argon} and \texttt{gemini-3.8-flash}, and Claude Opus 5.5, the planner and critic of \mRobotDef{}, as \texttt{claude-opus-5-5}.

\paragraph{Baselines.} RBench runs ReVidgen at \texttt{b03df27} with \path{Qwen/Qwen3.8-27B} at \texttt{1d4bf0f}. PhyJudge-9B is \path{NU-World-Model-Embodied-AI/phyjudge-9B} at \texttt{a1a8d98} on \path{Qwen/Qwen3.5-9B} at \texttt{c202236}. The WorldModelBench judge is \path{Efficient-Large-Model/vila-ewm-qwen2-1.5b} at \texttt{827be9e}, VideoCon-Physics is \path{videophysics/videocon_physics} at \texttt{2b908df}, PAI is \path{SHI-Labs/physical-ai-bench} at \texttt{2f3b687} and PDI is \path{AnteaWu/PDI-Bench} at \texttt{2433323}.

\paragraph{Video generators.} ABot-PhysWorld runs the code of \path{amap-cvlab/ABot-PhysWorld} at \texttt{7d47080} with the checkpoint \path{abotpw_i2v_480p.safetensors} of \path{amap_cvlab/Abot-PhysWorld} on ModelScope, on top of \path{Wan-AI/Wan2.1-I2V-14B-480P} at \texttt{6b73f84}. Cosmos~2.5 is \path{nvidia/Cosmos-Predict2.5-2B} at \texttt{0d37c74} and Cosmos3~Nano is \path{nvidia/Cosmos3-Nano} at \texttt{e59a53c}, both run with the diffusers pipelines at commit \texttt{efe46a5}. HunyuanVideo~1.5 runs its 480p image-to-video model, \path{hunyuanvideo-community/HunyuanVideo-1.5-Diffusers-480p_i2v} at \texttt{5a700ee}, and Wan2.2~5B its TI2V-5B model, \path{Wan-AI/Wan2.2-TI2V-5B-Diffusers} at \texttt{b8fff73}, both with diffusers 0.39.0.

\paragraph{Source datasets.} \Cref{tab:supp_sources} gives the repository, the revision and the licence of each source dataset.

\begin{table*}[t]
\centering
\caption{\textbf{Source datasets of RP-Bench.} Repository, revision and licence of each source dataset. All repositories are on Hugging Face except TAVP. $^\dagger$Revision of the stack-block repository; each of the eleven task repositories is pinned to its own revision.}
\label{tab:supp_sources}
\footnotesize
\begin{tabular}{@{}llll@{}}
\toprule
Dataset & Repository & Revision & Licence \\
\midrule
NVIDIA GR1 & \texttt{nvidia/PhysicalAI-Robotics-GR00T-GR1} & \texttt{b114c39} & CC BY 4.0 \\
MolmoAct household & \texttt{allenai/MolmoAct2-MolmoAct-Dataset-Household} & \texttt{13ac3af} & Apache 2.0 \\
RoboCOIN & \texttt{RoboCOIN/Airbot\_MMK2\_stack\_block} and 11 task repositories & \texttt{ee1901c}$^\dagger$ & Apache 2.0 \\
Robo360 & \texttt{liuyubian/Robo360} & \texttt{78c0046} & not stated \\
RoboMIND & \texttt{x-humanoid-robomind/RoboMIND} & \texttt{e7ffe31} & Apache 2.0 \\
ActionNet & \texttt{FourierIntelligence/ActionNet} & \texttt{eb3b5bb} & CC BY-NC-SA 4.0 \\
SO-101 & \texttt{armnet/armnetbench\_v01\_lerobot\_so101} & \texttt{1db69c8} & Apache 2.0 \\
PiperX & \texttt{vikrantsp/piperx\_task1} & \texttt{5869999} & Apache 2.0 \\
UR5e & \texttt{aimsjunhee/UR5e\_pick\_apple\_banana\_place\_v1} & \texttt{f01a4fb} & not stated \\
TAVP & \texttt{HCPLab-SYSU/TAVP} (GitHub) & \texttt{fcb05cd} & not stated \\
Piper & \texttt{coport-uni/PiPER\_pick\_black\_colored\_marker\_to\_box} & \texttt{0a412ab} & Apache 2.0 \\
Piper & \texttt{Smokepord/piper\_pick\_place\_xy\_random\_50\_v1} & \texttt{7b50366} & CC BY 4.0 \\
\bottomrule
\end{tabular}
\end{table*}

\section{Code and data release}
\label{sec:supp_code}
The supplementary material includes RP-Bench. The archive \texttt{videos-150MB.zip} holds all 891 videos of the evaluation set, the 761 generated videos and the 130 real demonstrations. The file \texttt{annotations\_891.csv} has one row per video, with its source dataset, its generator and, for the generated videos, the human labels of the five violation types, the per-link labels of \mRobotDef{} and the agent-mismatch and unsure marks. We will publicly release the code of \method{} upon acceptance.

\end{document}